\PassOptionsToPackage{unicode}{hyperref}
\PassOptionsToPackage{hyphens}{url}
\documentclass[
]{article}
\usepackage{xcolor}
\usepackage{amsmath,amssymb}
\usepackage{iftex}
\ifPDFTeX
  \usepackage[T1]{fontenc}
  \usepackage[utf8]{inputenc}
  \usepackage{textcomp} 
\else 
  \usepackage{unicode-math} 
  \defaultfontfeatures{Scale=MatchLowercase}
  \defaultfontfeatures[\rmfamily]{Ligatures=TeX,Scale=1}
\fi
\usepackage{lmodern}
\ifPDFTeX\else
\fi
\IfFileExists{upquote.sty}{\usepackage{upquote}}{}
\IfFileExists{microtype.sty}{
  \usepackage[]{microtype}
  \UseMicrotypeSet[protrusion]{basicmath} 
}{}
\makeatletter
\@ifundefined{KOMAClassName}{
  \IfFileExists{parskip.sty}{%
    \usepackage{parskip}
  }{
    \setlength{\parindent}{0pt}
    \setlength{\parskip}{6pt plus 2pt minus 1pt}}
}{
  \KOMAoptions{parskip=half}}
\makeatother
\usepackage{longtable,booktabs,array}
\usepackage{calc} 
\usepackage{etoolbox}
\makeatletter
\patchcmd\longtable{\par}{\if@noskipsec\mbox{}\fi\par}{}{}
\makeatother
\IfFileExists{footnotehyper.sty}{\usepackage{footnotehyper}}{\usepackage{footnote}}
\makesavenoteenv{longtable}
\usepackage{graphicx}
\makeatletter
\newsavebox\pandoc@box
\newcommand*\pandocbounded[1]{
  \sbox\pandoc@box{#1}%
  \Gscale@div\@tempa{\textheight}{\dimexpr\ht\pandoc@box+\dp\pandoc@box\relax}%
  \Gscale@div\@tempb{\linewidth}{\wd\pandoc@box}%
  \ifdim\@tempb\p@<\@tempa\p@\let\@tempa\@tempb\fi
  \ifdim\@tempa\p@<\p@\scalebox{\@tempa}{\usebox\pandoc@box}%
  \else\usebox{\pandoc@box}%
  \fi%
}
\def\fps@figure{htbp}
\makeatother
\providecommand{\tightlist}{%
  \setlength{\itemsep}{0pt}\setlength{\parskip}{0pt}}
\usepackage{iclr2027_conference}

\usepackage{newunicodechar}
\newunicodechar{χ}{\ensuremath{\chi}}
\newunicodechar{ρ}{\ensuremath{\rho}}
\newunicodechar{δ}{\ensuremath{\delta}}
\newunicodechar{²}{\textsuperscript{2}}
\newunicodechar{→}{\ensuremath{\rightarrow}}
\newunicodechar{≈}{\ensuremath{\approx}}
\newunicodechar{≲}{\ensuremath{\lesssim}}
\newunicodechar{≳}{\ensuremath{\gtrsim}}
\newunicodechar{≥}{\ensuremath{\geq}}
\newunicodechar{≤}{\ensuremath{\leq}}
\newunicodechar{√}{\ensuremath{\surd}}
\newunicodechar{·}{\ensuremath{\cdot}}
\newunicodechar{−}{\ensuremath{-}}
\newunicodechar{×}{\ensuremath{\times}}
\newunicodechar{∧}{\ensuremath{\wedge}}
\newunicodechar{η}{\ensuremath{\eta}}
\newunicodechar{μ}{\ensuremath{\mu}}
\newunicodechar{ψ}{\ensuremath{\psi}}
\newunicodechar{π}{\ensuremath{\pi}}
\newunicodechar{∎}{\ensuremath{\blacksquare}}
\iclrfinalcopy

\let\iclrOrigMaketitle\maketitle
\renewcommand{\maketitle}{\iclrOrigMaketitle\lhead{Preprint.}}
\usepackage{bookmark}
\IfFileExists{xurl.sty}{\usepackage{xurl}}{} 
\makeatletter
\@ifundefined{xmpquote}{}{}
\makeatother
\hypersetup{
  pdftitle={The Steering Budget: Examples beat Knobs},
  pdfauthor={Raj Kumar Rajendran},
  hidelinks,
  pdfcreator={LaTeX via pandoc}}

\title{The Steering Budget: Examples beat Knobs}
\author{Raj Kumar Rajendran}
\date{}

\begin{document}
\maketitle
\begin{abstract}
Generative models are steered with knobs --- prompts, guidance scales,
property tags. Turn one as hard as you like and, past a point, it stops
moving the property you care about. We find that ceiling is not a
shortcoming of the model but a \emph{budget}, set by the training data
before the model is trained: a property's movable range splits in two
--- the part a knob can reach, and a second, significant part that only
examples --- concrete instances of what you want more of --- can reach.
Usually that second part is the larger, so examples win, and by a lot
--- but not always, and the same budget says in advance where the range
is small enough that a strong knob is enough instead.

Reaching that second part takes a different move: instead of turning a
knob, you show the model examples, composed from what it already learned
rather than added to its training. A cheap audit of the training data
measures the budget; we give a recipe for building the example set that
reaches all of it.

This buys two things a knob can't. \emph{Reach:} it moves a property
across the whole budget, not just the part a knob reaches.
\emph{Expressiveness:} it steers toward targets you can only specify by
example --- including ones you can't put into words. We turn these into
a handful of falsifiable claims and verify them in two unrelated
domains, image and crystal-structure generation --- marking where a knob
is enough, and where only examples will do.
\end{abstract}

\section{Introduction --- Steering with knobs, and steering by
example}\label{introduction-steering-with-knobs-and-steering-by-example}

Suppose you have a generative model and you want it to lean a certain
way --- images with the feel of nature photography, crystal structures
with a wider band gap. You reach for the controls: a prompt, a guidance
scale, a property tag. Push them to their limit and they still carry the
output only so far --- the property you are steering settles well short
of where you wanted it, and leaning harder does nothing.

There is a more powerful move, and it is not a better knob. It is a
change in \emph{what you hand the model}. Instead of working the
controls, you \emph{show} it: a set of examples, and the instruction
``more like these.'' The examples are drawn from what the model already
learned --- you are not adding to its training or retraining it, only
choosing which of its own material it works from. We find that this is
the more powerful and more expressive way to steer --- and, more
surprisingly, that the gap between the two is not incidental but
\emph{budgeted}: how far each can move a property is fixed, before the
model is trained, by the data it learned from.

That budget is the spine of the paper. Examples play two different roles
in it. The \emph{training data} sets how big each part of the budget is
--- settled before the model exists. The \emph{examples you show at
steering time} are how you reach the part a knob cannot. The rest of
this section previews the split, names the two operations, shows why the
part a knob misses is the part people most often want --- and ends on
the targets an expert can recognize but cannot put into words.

\subsection{Telling and showing}\label{telling-and-showing}

So far we have two handles: knobs and examples. To say precisely how far
each reaches, we need names for the two \emph{acts}, not just the two
things you hold. Turning a knob is \emph{telling}. Handing over examples
--- ``more like these'' --- is \emph{showing}. These two acts have a
precise structural difference, and it is the crux of the paper.

Fixing an input parameter before generation --- an image's class, a
crystal's element-set --- picks one \emph{bin} to generate into: a group
of outputs that all share that value. Turning a knob instead varies some
\emph{other} input parameter, continuous or discrete, hoping to bias the
target property, without itself picking a bin. Both moves are
\emph{telling}: whichever bins are in play --- one, if an input was
fixed to it, or all of them in their natural mix, if no input was fixed
--- telling can only reweight the outputs \emph{within} them, never move
weight \emph{between} them. Confining a knob to one bin doesn't worsen
this limit, and leaving it unconfined doesn't escape it either: either
way, telling is reweighting whatever bins it operates over, one or all
of them, and never crossing from one into another. That is the whole of
what limits telling's reach, regardless of which or how many bins are in
play.

\emph{Showing} --- handing over a set of examples and asking for more
like them --- works differently: it reweights the mix \emph{across} bins
instead of within them, reaching values no single bin contains.

\begin{quote}
\textbf{The telling/showing split.} \emph{Telling} reweights whichever
bins are in play --- one or all of them --- but can never move weight
\emph{between} bins. \emph{Showing} reweights the mix across bins
directly. That crossing is the one capability telling structurally
lacks, no matter how hard the knob is pushed.
\end{quote}

That contrast --- telling confined to whichever bins it operates within,
showing moving weight across them --- is why showing reaches what
telling cannot. The rest of the paper makes it precise, shows exactly
how far each posture reaches, and tests the prediction in two unrelated
domains.

\subsection{A budget on what steering can
reach}\label{a-budget-on-what-steering-can-reach}

Behind that difference is something you can measure. \emph{Telling} is
the familiar operation: decide what you want and lean on the controls
--- the guidance dial, a descriptive token, an adapter --- nudging the
distribution one request at a time, within whichever bins are in play.
\emph{Showing} is the other: hand over a set of examples and let their
mix of bins shape a whole batch, never saying what they have in common.

Any property you care about, like brightness or a crystal's band gap,
varies two ways across the bins we just defined: \emph{within} bins
(golden retrievers differ in brightness) and \emph{between} bins (a
beach averages brighter than a wine cellar). Those two pieces add up to
all the variation there is --- and \emph{telling}, confined to whichever
bins are in play, reaches only the within-bin part, while
\emph{showing}, choosing which bins the batch is drawn from, reaches the
between-bin part. Each piece is a \emph{budget} --- a ceiling on how far
that operation can move the property --- fixed by the training data
before the model is built. \textbf{Preliminaries} makes this precise.

One question this raises: if the budget is set by the data, doesn't that
write the model out of the story? No --- a stronger model produces a
denser, more novel \emph{set} of outputs, but that is a richer set, not
a wider \emph{spread} of any one property; we make the distinction
precise later.

So the two reach genuinely different ground. Much of what people
actually want --- outputs that meet two conditions at once, real variety
across a batch, a preference you cannot put into words --- lives in the
part \emph{telling} cannot reach. That is the ground \emph{showing}
covers.

\subsection{Steering toward what you can't
name}\label{steering-toward-what-you-cant-name}

Reach --- moving a property further than a knob can --- is the
measurable payoff, and most of the paper is about it. There is a second
payoff, and it matters most exactly where naming the target is hardest.
A materials scientist scrolling a list of candidate structures stops on
one: \emph{that one looks promising.} Pressed for a reason, they
hesitate --- the judgment is fast and reliable, the justification slow
and partial. A film editor who keeps one take in ten is in the same
position: the cut hangs together, but the rule that chose those takes
resists being written down. This is the ordinary condition of expertise,
not an edge case; Polanyi's phrase is \emph{we know more than we can
tell.}

Every knob asks the expert to tell anyway --- to name the target in the
model's vocabulary before it can be pursued. Showing does not. The
specification lives in the examples themselves, so you can steer toward
a taste you can demonstrate but not articulate: keep the ten candidates
you would have kept, hand them over, and ask for more like them. And
because showing composes by \emph{union} --- the examples may disagree,
and ``more like these'' finds what they share without anyone naming it
--- it reaches tastes, trade-offs, and combinations no single named
target captures. This expressiveness is what the budget's between-bin
part pays for; we return to it, and to its uses, in
\textbf{Application}.

\subsection{Why this matters even if you never use
exemplars}\label{why-this-matters-even-if-you-never-use-exemplars}

You might expect the \emph{telling}/\emph{showing} split to matter only
when you have exemplars in hand. It reaches further than that: it sits
underneath some of the most familiar frustrations in generative
modeling.

Here is why. Almost every way we train or tune a model --- fitting it to
data, optimizing it against a reward, nudging it toward preferences ---
works \textbf{one output at a time}, pushing each toward ``better.''
Tuning output by output can sharpen each one, but it cannot build
\emph{variety} --- because variety is not a property of any single
output. It is a property of the whole \emph{collection}, and the
collection is exactly the ground \emph{showing} governs and
\emph{telling} never reaches.

So the harder you tune for quality, the less variety comes back. Anyone
who has fine-tuned a model and watched it turn repetitive has met this.
The variety did not evaporate for no reason: output-by-output tuning can
only narrow it, never produce it; the variety lives between bins, and
only \emph{showing} reaches it.

The field already has names for the symptoms, usually treated one by
one: a model collapsing onto a handful of outputs (\emph{mode
collapse}); a tuned model that answers well but has lost its range (the
\emph{alignment tax}); a model that games its score by overproducing one
safe thing (\emph{reward hacking}); a model that can satisfy two
requirements on their own but not together (\emph{compositional
generalization}). One cause sits under all four: each asks
output-by-output tuning to deliver something it structurally cannot ---
variety, coverage, a joint, all properties of the \emph{collection}
rather than of any single output --- and that tuning can only
concentrate, never supply it. Reaching them takes a move across the
collection, which is what \emph{showing} names. Alongside each
phenomenon's familiar drivers, Theory makes that tie explicit. The lost
variety, it turns out, is part of the budget itself --- used up by
tuning, and put back only by acting on the collection.

\subsection{Evidence across two
domains}\label{evidence-across-two-domains}

We do not just argue this; we test it, in two fields that share no
model, data, or code --- a diffusion model trained on ImageNet, where
the bins and the steering are easy to see, and a generator of crystal
structures, where an expert routinely flags a promising candidate
without being able to say why. This section previews what we find;
\textbf{Evidence} has the numbers.

\textbf{Examples are the more powerful control in general --- and the
budget says exactly when.} On crystals, choosing which chemistries a
batch is drawn from moves a target tens to hundreds of times more than
the model's own ``high band gap'' dial --- and still several times more
when we replace that dial with the strongest within-chemistry knob we
can build. On images, choosing which classes moves the semantic targets
a guidance dial can barely touch. The advantage is not universal, and
the framework predicts when knobs are powerful enough. In the image
domain, we show that knobs are sufficient to move brightness, and can be
made to improve \emph{aesthetics} too --- but only by drifting across
bins.

\textbf{Examples express what a knob cannot.} Ask for a batch that holds
two opposing kinds at once --- clean vehicles and clean food, or
crystals spanning a materials trade-off --- and a knob gives you one
kind, or a blurred average of the two; only composing examples puts
genuine instances of both in the batch.

\textbf{And the budget is real, not just bookkeeping.} The split into
the two parts is exact by construction. Properties aligned with the
model's training goals carry through to the trained model; some do not
--- and we show you can predict which will.

Most of the checks are cheap to verify. The budget itself is a variance
calculation on training data --- it runs on a laptop. Only the last
step, confirming that a real model behaves the way the budget predicts,
has to generate samples; even that fit on a single machine with a GPU in
a few hours.

\subsection{What the paper
contributes}\label{what-the-paper-contributes}

The rest of the paper turns the tell/show idea into something you can
compute, test, and use. The central claim is this: \emph{the steerable
mean-shift of a labeled property splits into a telling part and a
showing part, both computable from training data before the model exists
--- and this reveals which targets telling cannot reach.} Five
contributions follow, in the order they arrive:

\begin{itemize}
\tightlist
\item
  \textbf{The distinction itself, made precise} --- \emph{telling} and
  \emph{showing} as two operations, and the within-bin/between-bin
  mechanism that fixes what each can reach.
\item
  \textbf{A budget you can compute in advance} --- the exact accounting
  that splits a property's variation into a within-bin part (what
  \emph{telling} reaches) and a between-bin part (what \emph{showing}
  reaches), with a ceiling on each, from training data alone.
\item
  \textbf{A unifying lens on four familiar problems} --- mode collapse,
  the alignment tax, reward hacking, and the failure to compose
  requirements, read as a single mismatch: \emph{telling} sent to do a
  \emph{showing} job.
\item
  \textbf{Evidence in two unrelated fields} --- images and crystals ---
  that composing \emph{examples} is, in general, the stronger lever: it
  moves a target several-fold further than the strongest knob we could
  build wherever the budget is mostly between-bin, and it is the only
  way to cover two opposed properties in one batch (which neither
  prompting nor composing \emph{conditions} can do). The budget also
  calls the cases where examples are \emph{not} the move --- the
  low-level scalars (image brightness, aesthetic) whose range sits
  mostly \emph{inside} the bins, where a strong knob wins; a goal that
  is merely a batch average, which a single well-chosen category already
  meets --- and the one place nothing helps: a property the model
  reproduces too poorly for any recipe to reach.
\item
  \textbf{A recipe --- even when you can't name the target.} When
  \emph{telling} falls short, we show how to choose which categories to
  draw from: if you can \emph{score} the target, find the categories
  that rank high on it in the training data. If you \emph{can't} --- a
  preference you can only point to --- use a handful of samples you like
  to choose categories for you by picking the categories they fall into.
\end{itemize}

\textbf{What's new, and what isn't.} Steering by example is not itself
new --- example- and reference-conditioned generation take similar
approaches. What's new here is the articulation of the budget: a
quantitative account, read from the training data, of how far
\emph{telling} and \emph{showing} can each reach. (Related Work places
it among these.)

\subsection{When a knob is enough}\label{when-a-knob-is-enough}

Examples are the stronger lever, but not everywhere --- and the same
budget that says when they win says when they are not worth the trouble.
There are two such cases, each with a cheap check:

\begin{itemize}
\tightlist
\item
  \textbf{The knob already reaches.} When little of a property's
  variation sits between bins --- its between-bin share is small --- and
  a knob drives it directly, a knob is enough. \emph{Check:} measure
  that between-bin share from the data; if it is small, the knob has the
  room. (Image brightness is the one such case we found.)
\item
  \textbf{The model doesn't reproduce the property.} The budget is read
  off the data, so it transfers only to properties the model makes
  faithfully; for one it does not, no recipe reaches it. \emph{Check:}
  audit a sample of the model's own outputs against the data.
\end{itemize}

In short: measure the budget first, and let it tell you whether to reach
for examples at all.

\subsection{Roadmap}\label{roadmap}

\textbf{Preliminaries} sets up the shared language in two pages --- the
bins, the two operations, the two budgets --- and sends you on.
\textbf{Theory} derives the two bounds, shows how the choice of
exemplars scales what showing can reach, states the four falsifiable
claims, and closes by connecting them back to the phenomena above.
\textbf{Evidence} tests the claims in both fields. \textbf{Application}
turns it into a recipe: measure your budget, measure how far
\emph{telling} already gets you, and --- if a gap remains --- read its
shape to choose how to close it.

\section{Preliminaries --- the cheap key and the expensive
target}\label{preliminaries-the-cheap-key-and-the-expensive-target}

The introduction named two moves --- \emph{telling} and \emph{showing}
--- and a budget that splits between them. This section makes all three
precise: the bins, the two moves, and the two budgets. Everything after
is built from them.

A generative model hands you a stream of outputs, and you want to steer
it --- more of some kinds, fewer of others. Some of what makes an output
desirable is cheap to read: the class of an image, the elements in a
crystal. What you usually \emph{care} about is expensive: whether an
image \emph{looks good} takes a second trained model, or a person, to
judge; a crystal's \emph{band gap} --- the number that decides whether
it is a useful semiconductor --- takes a physics simulation that can run
for hours.

These two reads --- the cheap one and the expensive one --- are most of
the notation in this section.

\begin{itemize}
\tightlist
\item
  The cheap read is a \textbf{label} \(P\), from a function \(\pi\) (the
  classifier, the formula-reader): \(P = \pi(D)\) for an output \(D\).
\item
  The expensive property is \(L\), from a function \(\psi\) (the
  aesthetic scorer, the band-gap simulation): \(L = \psi(D)\).
\end{itemize}

The one idea to carry: \textbf{\(P\) is a key, not a measurement.} It
\emph{sorts} outputs into groups: every golden-retriever picture into
one, every table-salt crystal into another. Call each group a
\textbf{bin}. The whole paper is about steering the expensive \(L\) by
choosing how we work with the cheap bins \(P\) gives us.

\subsection{Bins}\label{bins}

Run \(\pi\) over a pile of outputs and it drops each into a bin --- one
bin per distinct label. For the image model there is a ``golden
retriever'' bin, a ``ladybug'' bin, one per class. For the crystal model
\(\pi\) reads (elements, atom count), so there is a
\texttt{(Si,\ O;\ 6)} bin, a \texttt{(Mg,\ O;\ 2)} bin, and so on. With
\(B\) distinct labels there are \(B\) bins, and \(P_b\) is the label of
bin \(b\). (A real request carries more than the key --- a guidance
scale, a style token --- the \emph{knobs} of the introduction. \(P\) is
the part this paper turns on, so we fold those into ``the controls'' and
set them aside for now; they return below as \emph{telling}.)

\subsection{What the model produces}\label{what-the-model-produces}

What is the model trained to make? More of whatever it learned from ---
and it is \textbf{random}: ask it once and you get one output; ask
again, same request, and you get a different one. A request doesn't pin
down an output, it pins down a \emph{spread} of outputs. What that
spread looks like is fixed by the data the model trained on. So name
that data: \textbf{\(\mathcal{J}\) is the distribution of your training
set.} Train on ImageNet and \(\mathcal{J}\) is ImageNet's distribution;
train on a different collection and \(\mathcal{J}\) is different. The
framework is computed relative to the data you used: your budget for a
property is set by \emph{your} training set.

We never need \(\mathcal{J}\) itself --- a distribution over whole
images, or whole crystals. We need only its \textbf{shadow} on
\((P, L)\): pass each output through the cheap label \(P\) and the
property \(L\), and keep a few numbers per bin. That shadow is all the
machinery ever touches, and it is concrete --- just statistics of your
data.

\subsection{The working picture}\label{the-working-picture}

Now take the expensive property \(L\) --- the band gap --- and look at
how it varies across the bins. It varies two ways:

\begin{itemize}
\tightlist
\item
  \emph{within} a bin, the \texttt{(Si,\ O;\ 6)} crystals do not all
  share one band gap;
\item
  \emph{between} bins, the averages differ --- silicon oxides sit near 1
  eV, magnesium oxides up near 7 eV.
\end{itemize}

So picture \textbf{\(B\) columns, one per bin, each a cloud of \(L\)
values} --- every cloud has some spread, and the columns sit at
different heights. One output's band gap is one dot in one column.

\begin{figure}
\centering
\includegraphics[width=0.6\linewidth,height=\textheight,keepaspectratio,alt={Each bin is a column of L values: every column has some spread (the within-bin variation) and sits at a height (the bin's average). Some bins are tight and high, others wide and low. (Schematic --- illustrative numbers, not measured data.)}]{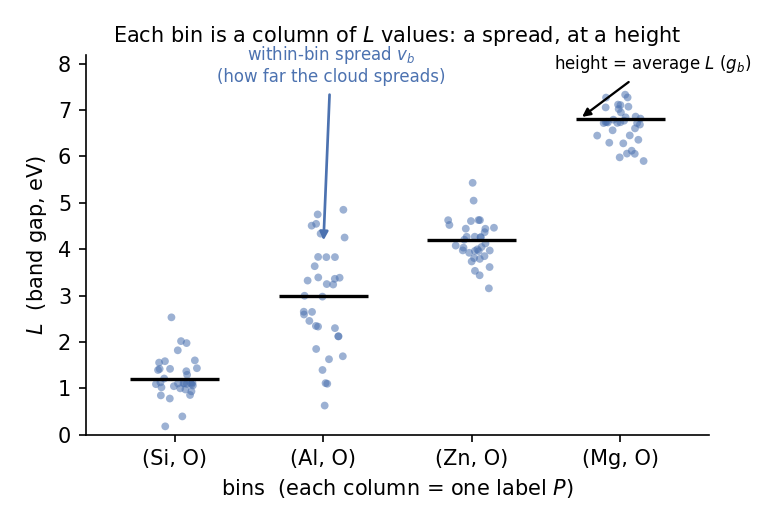}
\caption{Each bin is a column of L values: every column has some spread
(the within-bin variation) and sits at a height (the bin's average).
Some bins are tight and high, others wide and low. (Schematic ---
illustrative numbers, not measured data.)}
\end{figure}

\emph{Each column is a bin; the dots are the \(L\) values in it. A
column's spread is the within-bin variation; its height is the average.}

This is the picture to keep for the whole paper. (The image model looks
the same: each class is a column, brightness varies within a class, and
some classes --- beaches --- average brighter than others --- coal
mines.)

\subsection{The audit}\label{the-audit}

The working picture reduces to three numbers per bin, and producing them
is one fixed procedure --- an \textbf{audit}: take a set of outputs,
sort each into its bin via \(P\) --- cheap, so you bin them all --- then
score the expensive \(L\) on a \emph{handful} of outputs from each bin
(few, because \(\psi\) is dear, but enough to pin down the bin's average
and spread), and record, per bin,

\begin{itemize}
\tightlist
\item
  its \textbf{share} \(w_b\) --- the fraction of outputs that land in
  it;
\item
  its \textbf{height} \(g_b\) --- the average \(L\) in it;
\item
  its \textbf{spread} \(v_b\) --- how much \(L\) varies inside it.
\end{itemize}

Those three numbers per bin are the audit's output --- its
\textbf{shadow} on \((P, L)\) --- and the entire input to everything
that follows. Because \(\pi\) is cheap but \(\psi\) dear, the shares
\(w_b\) come out \emph{exact} --- you bin everything --- while the
heights and spreads are \emph{estimates} from the scored handful: a few
per bin are enough for the targets steering cares about, more when \(L\)
varies mostly within bins. \textbf{Application} gives the rule for
choosing the count, and the correction for sampling so few. You can run
the very same procedure on two different sets of outputs: your training
data (\textbf{audit \(\mathcal{J}\)}) or a batch of the model's own
samples (\textbf{audit \(\mathrm{G}\)}).

\textbf{From here on those two terms mean precisely this} --- the
binning and the three numbers --- and the rest of the paper uses them
without re-explaining.

\subsection{Two ways to steer}\label{two-ways-to-steer}

Say you want brighter images. Here are the two moves the introduction
named, now made precise --- and the rest of the paper turns on the
difference between them.

\begin{itemize}
\tightlist
\item
  \textbf{Telling} (\emph{per-prompt steering}). Work \emph{within} one
  bin --- say ``golden retriever'' --- and lean on the controls: turn up
  the guidance dial, add the token ``bright'', attach an adapter. You
  shift \emph{that bin's} outputs toward its brighter end. In the
  picture: you stay in one column and pull its dots toward an end.
\item
  \textbf{Showing} (\emph{exemplar steering}). Instead of describing the
  target, you hand the model a batch of examples and ask for more like
  them. Sorted through the cheap label \(P\), that batch \emph{is} a mix
  of bins --- a recipe ``70\% beach scenes, 30\% snow,'' written
  \(\mu_P\). It is the mix, not the individual examples, that the move
  turns on, so from here we describe showing by its recipe \(\mu_P\)
  over bins. In the picture: you choose \emph{which columns} the batch
  comes from.
\end{itemize}

You hand over examples, but what steers is the mix of bins they fall
into --- that is the one shift in vocabulary to keep. \emph{Telling}
works one column; \emph{showing} weights different columns. They reach
different parts of the picture. We make all of this precise next.

\subsection{Two budgets}\label{two-budgets}

The total spread of \(L\) --- its variance across the data --- splits
cleanly into the two parts the picture shows, and each move draws on
one:

\begin{figure}
\centering
\includegraphics[width=0.85\linewidth,height=\textheight,keepaspectratio,alt={The variance budget: the total spread of L splits into the within-column part (the within-bin budget, telling) and the between-column part (the between-bin budget, showing). (Schematic --- illustrative numbers, not measured data; the measured crystal split, e.g.~band gap E/(E\{+\}T)=0.98, is in Evidence §4.)}]{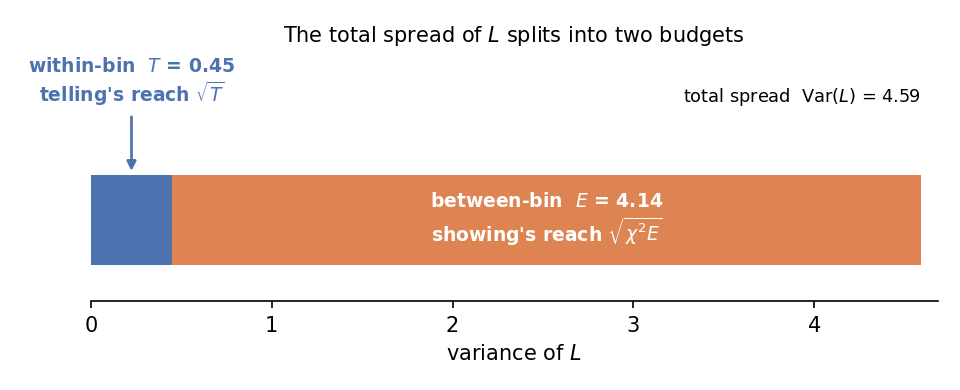}
\caption{The variance budget: the total spread of L splits into the
within-column part (the within-bin budget, telling) and the
between-column part (the between-bin budget, showing).
(\textbf{Schematic --- illustrative numbers, not measured data;} the
measured crystal split, e.g.~band gap \(E/(E{+}T)=0.98\), is in Evidence
§4.)}
\end{figure}

\begin{itemize}
\tightlist
\item
  \textbf{Telling} draws on the spread \emph{within} a column --- the
  \textbf{within-bin budget} \(T\).
\item
  \textbf{Showing} draws on the spread \emph{between} columns --- the
  \textbf{between-bin budget} \(E\).
\end{itemize}

They sum to the whole spread, exactly:

\[\mathrm{Var}_{\mathcal{J}}(L) \;=\; \underbrace{\textstyle\sum_b w_b\, v_b}_{T\ \text{(within)}} \;+\; \underbrace{\textstyle\sum_b w_b\,(g_b - \bar g)^2}_{E\ \text{(between)}}, \qquad \bar g = \textstyle\sum_b w_b\, g_b.\]

A budget is a \emph{variance} --- the room available, nothing more. It
says how far a property \emph{could} be moved, not what any particular
move does with it.

The two moves each draw on one budget. \emph{Telling} reweights
\emph{within} a bin, so it draws on \(T\); \emph{showing} reweights
\emph{across} bins with the recipe \(\mu_P\), so it draws on \(E\).
Either way, the move shifts the average \(L\) of what you generate, and
that shift is at most \(\sqrt{\chi^2\cdot\text{budget}}\) --- where
\(\chi^2\) measures how hard the move pushes from the original mix.

The budgets are fixed by the three numbers, \emph{before the model is
built}. So the deciding difference between the moves is not how hard you
push but \textbf{which} budget each can touch at all --- \(T\) for
telling, \(E\) for showing. Telling cannot cross bins.

We have written the goal as the batch's average \(L\), but it can be any
property of the batch --- its spread, or whether it \emph{covers} two
kinds at once. \textbf{Theory} derives the bounds, unpacks \(\chi^2\),
and shows which move each kind of goal needs.

\subsection{Data audit, model audit}\label{data-audit-model-audit}

Everything so far has audited the training data \(\mathcal{J}\) --- the
budget is read off the data, before the model is even built. But what
you ultimately care about is what the \emph{model} produces: an audit of
its own outputs, \(\mathrm{G}\). So why should the data audit tell you
anything about the model's?

Because a generative model is trained to reproduce its data. For a
property it reproduces well, the model's outputs fall into the bins with
the same shares, heights, and spreads as the data --- the model audit
matches the data audit --- so the cheap data audit (no model, no
sampling) stands in for it.

The catch is that the model is not trained on your property \(L\). It is
trained to reproduce the data as a whole --- to predict the next token,
to denoise toward real samples --- and \(L\) is just something you
happen to care about. So the data audit stands in faithfully only when
\(L\) is something the model reproduces well. A crystal generator's
\textbf{density} --- read straight off the shape it draws --- comes out
matching the data; its \textbf{stability} --- an energy that a small
error in the structure throws off --- does not, and there the model
audit drifts from the data's.

When you are not sure a property carries, don't guess --- check it:
sample the model, audit those outputs, and compare them to the data's
bin by bin --- the two numbers to read are the \textbf{drift} \(\delta\)
(did the bin means move?) and the \textbf{spread ratio} \(\rho\) (did
the within-bin variety carry?). \textbf{Application} gives that check
step by step, \textbf{Evidence} runs it across easy and hard properties
in both domains, and \textbf{Theory}'s \emph{Scope} marks where the data
audit holds and where it breaks. (A fine-tuned model reproduces a
\emph{reshaped} version of the data --- \textbf{Theory} handles that
too.)

Every symbol above is also collected in a one-page table, grouped by
kind, in the \textbf{Appendix} (§A0, \emph{Notation}) --- for reference
when a symbol turns up later without its introduction nearby.

\subsection{Where to go from here}\label{where-to-go-from-here}

You now have the working picture --- \(B\) bins of \(L\), the two moves,
the two budgets. The rest of the paper builds on it in order:

\begin{itemize}
\tightlist
\item
  \textbf{Theory} derives where the two square-root bounds come from,
  shows how the choice of recipe \(\mu_P\) scales showing's reach (the
  \(\chi^2\) factor), and states the falsifiable claims that follow ---
  then closes with a short note on what this implies for familiar
  training phenomena (mode collapse, the alignment tax, and the like).
  (It also notes \emph{why} variance is the right currency --- a
  methodological choice: cruder measures like max-minus-min are noisy.)
\item
  \textbf{Evidence} tests those claims in two unrelated domains.
\item
  \textbf{Application} turns the result into a recipe: how to measure
  your own budget, tell whether telling already suffices, and pick the
  construction that closes the gap if it does not.
\end{itemize}

A reader who wants the practice before the derivations can jump straight
to \textbf{Application} and refer back.

\section{Theory --- How far telling and showing
reach}\label{theory-how-far-telling-and-showing-reach}

\textbf{Preliminaries} set up the working picture and the two ways to
steer. To recall: a generator \(\mathrm{G}\) produces outputs \(D\); a
cheap key \(P = \pi(D)\) sorts them into \(B\) bins; the expensive
target \(L = \psi(D)\) fills each bin with a spread of values; and you
steer \(L\) two ways --- \textbf{telling} stays inside one bin and
nudges its outputs, while \textbf{showing} changes \emph{which} bins the
batch is drawn from. Preliminaries also named the two budgets --- \(T\)
(within-bin) and \(E\) (between-bin) --- and their reaches. This section
derives them.

§1 shows how the total spread of \(L\) splits into the two budgets. §2
derives how far each operation can move \(L\), how the recipe \(\mu_P\)
scales showing's reach through a factor \(\chi^2\), and --- given a goal
--- which operation it calls for and, when that is showing, the
\emph{shape} (curvature) of the recipe it needs. §3 states the
falsifiable claims that follow, including how the data-side budget
transfers to a real model.

§4 then reads off what the budget implies for some familiar training
phenomena, and §5 marks where the framework applies and where it stops.
The detailed machinery, the case-by-case reach of each operation, and
the full statement of the boundaries are in the \textbf{Appendix}. A
practitioner can skip to \textbf{Application}.

Recall the per-bin quantities: bin \(b\) has share \(w_b\), mean
\(g_b = \mathbb{E}_{\mathcal{J}}[L \mid P_b]\), and within-bin variance
\(v_b = \mathrm{Var}_{\mathcal{J}}[L \mid P_b]\).

\begin{center}\rule{0.5\linewidth}{0.5pt}\end{center}

\subsection{1. The budget: one variance, split in
two}\label{the-budget-one-variance-split-in-two}

Take the property \(L\) and ask where its variation across the data
lives. Some sits \emph{within} bins --- outputs that share a key still
differ in \(L\) --- and some sits \emph{between} bins --- different bins
have different average \(L\). Those are the only two places it can live,
and across the \(B\) bins they add up exactly:

\[\underbrace{\mathrm{Var}_{\mathcal{J}}(L)}_{\text{total}} \;=\; \underbrace{\sum_b w_b\, v_b}_{T\ \text{(within-bin)}} \;+\; \underbrace{\sum_b w_b\,(g_b - \bar g)^2}_{E\ \text{(between-bin)}}, \qquad \bar g = \sum_b w_b\, g_b.\]

\(T\) is the within-bin spread averaged over bins --- the room
\textbf{telling} works in, since telling stays inside one bin. \(E\) is
the spread of the bin means --- the room \textbf{showing} works in,
since showing moves between bins. This identity is the \textbf{law of
total variance} (Casella \& Berger, 2002), exact for any
\((\pi, L, \mathcal{J})\) --- no overlap, no remainder. Its two terms
are the room the two operations work in. The pieces here are elementary
--- the total-variance split and the Cauchy--Schwarz reweighting bound
below are textbook; the contribution is not the mathematics but
composing them into a cheap, pre-model, three-numbers-per-bin audit that
forecasts the telling/showing crossover and, under a testable condition,
carries to a trained model. And how the total divides is the
practitioner's to set through \(\pi\) --- finer bins move variance into
\(E\), coarser into \(T\) (at the extremes, one-output bins give
\(T \to 0\); a single bin gives \(E \to 0\)) --- fixing, before any
model exists, which operation has room to work.

\textbf{Note --- the choice of variance.} Why measure spread by variance
rather than the range \(\max L - \min L\)? The range carries no
within/between split, so it could not be divided into the two budgets at
all; it is fixed by a single extreme output; and it is routinely
exceeded by the model's own outputs. Variance is the stable data-side
quantity --- a methodological choice, argued in full in the Appendix.

\begin{center}\rule{0.5\linewidth}{0.5pt}\end{center}

\subsection{2. The two reaches}\label{the-two-reaches}

A budget is \emph{room}; the \textbf{reach} is how far an operation can
actually move the average of \(L\), given that room. Both operations
move it the same way --- by \textbf{reweighting}: shifting how much of
the batch comes from where. Reweighting obeys a single law --- depart
from a set's own proportions and the most you can move its average is
set by how \emph{far} you departed:
\[|\Delta\,\mathbb{E}[L]| \le \sqrt{\chi^2 \cdot \sigma^2},\] where
\(\sigma^2\) is the variance of the set you reweight and the
\textbf{Pearson \(\chi^2\)-divergence} measures the size of the
departure --- \(0\) when you change nothing, growing as you pile weight
onto rare values. This is Cauchy--Schwarz (derived just below). One
standard deviation, \(\sqrt{\sigma^2}\), is only the \emph{gentle} case
\(\chi^2 = 1\); push harder and you move further, at a price that grows
as \(\sqrt{\chi^2}\). The two operations differ in \textbf{which} set
they are allowed to reweight.

\textbf{Telling} stays inside the bin you named. A prompt or a guidance
knob can shift weight among the outputs \emph{within} a bin, but it
cannot move an output into a different bin --- it cannot change which
class, or which chemistry, the sample is. So telling reweights within
bins, and the variance it acts on is the within-bin budget \(T\):
\[|\Delta\,\mathbb{E}[L]| \le \sqrt{\chi^2_{\text{tell}}\cdot T}.\]
Pushing the controls harder raises \(\chi^2_{\text{tell}}\), but telling
is still confined to the room a bin already has. In practice the
controls push \emph{gently}: turning the guidance dial to its limit
moves a bin's average only a little, so \(\chi^2_{\text{tell}}\) stays
small and telling's realized reach sits near the unit-effort scale
\(\sqrt{T}\) (Evidence measures it).

\textbf{Showing} changes the bin mix instead. Its design knob is the
recipe \(\mu_P\), a distribution over the bins (Preliminaries: ``70\%
beaches, 30\% snow''). Draw bins in the data's own proportions ---
\(\mu_P = \mathcal{J}_P\) --- and you reproduce the data, shifting
nothing; to move the average you must \emph{depart} from
\(\mathcal{J}_P\), drawing more from the high-\(L\) bins and fewer from
the low. The shift you get is
\[\Delta\,\mathbb{E}[L] = \sum_b \big(\mu_P(b) - \mathcal{J}_P(b)\big)\, g_b,\]
a sum across bins of \emph{how much you over- or under-weight each bin}
times \emph{that bin's average} --- an inner product of two lists. And
an inner product is at most the product of the two magnitudes
(\textbf{Cauchy--Schwarz}): one is how far \(\mu_P\) departs from
\(\mathcal{J}_P\), the \textbf{Pearson \(\chi^2\)-divergence}
\[\chi^2(\mu_P\|\mathcal{J}_P) = \sum_b \frac{(\mu_P(b) - \mathcal{J}_P(b))^2}{\mathcal{J}_P(b)},\]
the other the spread of the bin means, \(\sqrt{E}\). So showing
reweights \emph{across} bins and reaches the between-bin budget \(E\):
\[|\Delta\,\mathbb{E}[L]| \le \sqrt{\chi^2(\mu_P\|\mathcal{J}_P)\cdot E},\]
largest when the recipe piles its extra weight on the bins with the most
extreme means. The \(\chi^2\) depends only on \((\mu_P, \mathcal{J}_P)\)
--- never on the label.

The two reaches have the \emph{same form} ---
\(\sqrt{\chi^2\cdot\text{budget}}\) --- so the asymmetry between telling
and showing lives in two other places. First, and decisively, it is
\textbf{structural}: telling is locked to the within-bin budget \(T\)
and showing to the between-bin budget \(E\), a division fixed by the key
\(\pi\) before any model exists. Telling cannot reach \(E\) \emph{at
all}, because it cannot change bin membership --- that is the heart of
the matter, and it holds no matter how hard the controls are pushed.
Second, it is \textbf{empirical}: pushed to its limit a knob departs
only slightly from its bin's own proportions (\(\chi^2\) near \(1\)),
while a recipe departs sharply from the data's mix (\(\chi^2\) of order
\(100\)) --- so showing realizes a far larger \(\chi^2\), and moves the
mean far more (Evidence measures both).

These two halves are easy to conflate, so we separate them. The
\textbf{structural} half is true \emph{by definition}: we call an
operation \emph{telling} precisely when it holds the key \(\pi\) fixed
--- reweighting within a bin without changing bin membership --- so
``telling only reaches \(T\)'' by definition. (Note: In one experiment
in Evidence, we see that a knob pushed hard enough to rival showing
turns out to \emph{leave} its bin. Since it no longer is confined to a
bin, we call that showing). The \textbf{empirical} content is the
separate, non-definitional part: that real, deployed knobs realize only
\(\chi^2\approx1\), two orders below a recipe's \(\chi^2\approx100\) ---
and that the one knob strong enough to close the gap does so only by
ceasing to hold \(\pi\) fixed, i.e.~by becoming showing.

Both reaches are \textbf{ceilings} --- computable from the bins before
any model is built, and bounding the most an operation \emph{could} move
the mean, not predicting what it will. They come from Cauchy--Schwarz,
tight only when all the weight sits on the most extreme members; a real
recipe lands below, and the gap widens as \(\chi^2\) grows (Claim 2).
Whether a model even delivers the in-support reach is Claim 1.

\subsubsection{Which operation a goal
needs}\label{which-operation-a-goal-needs}

The reaches give the \emph{room}; which operation claims it is the
\textbf{reach question}. A shift small enough to find \emph{inside} your
bin --- under \(\sqrt{T}\) --- needs only \textbf{telling}: lean on the
knobs. A larger one is out of any single bin's within-reach, so you must
\textbf{show} --- act on the mix of bins \(\mu_P\). Telling steers
\emph{within} the bin you are in; it never chooses the bin.

Some goals you cannot even write down --- a quality you only recognize
by eye; there you \emph{show} directly, letting a set of examples' own
mix of bins be the recipe (the \textbf{lift}; Application). For a goal
you \emph{can} write, one question remains once you are showing: the
\emph{shape} of the recipe.

\subsubsection{Curvature --- the shape of the showing
recipe}\label{curvature-the-shape-of-the-showing-recipe}

The reach question decided \emph{whether} to show; curvature decides
\emph{how}. Run a recipe \(\mu\) over the bins and you get a
\textbf{batch} of outputs; the goal is some number you read off that
batch --- its average band gap, say, or whether it holds both a wide-gap
and a stable structure. So the goal is a function \(F(\mu)\) of the
recipe, maximized over the simplex
\(\Delta = \{\mu \ge 0,\ \sum_b \mu(b) =
1\}\). Its curvature places the maximum:

\begin{itemize}
\tightlist
\item
  \textbf{\(F\) convex (linear included) → a vertex of \(\Delta\):
  concentrate \(\mu\) on one bin.} Say you want the batch's average band
  gap as high as possible: \(F(\mu) = \sum_b \mu(b)\, g_b\) is linear in
  \(\mu\), so it is largest with all the weight on the single
  highest-gap chemistry. A rate --- the fraction of the batch past a bar
  --- behaves the same way. That one bin is what the \emph{audit} finds,
  so concentrating on it is still showing (reach \(E\)), not telling.
\item
  \textbf{\(F\) strictly concave → the interior: spread \(\mu\) across
  bins.} Now you want a batch that \emph{covers} two kinds at once --- a
  wide-gap structure and a stable one. No single chemistry is both; each
  sits in one corner, so a mix of a wide-gap bin and a stable bin holds
  both, and the blend beats any lone bin. Diversity behaves the same
  way.
\end{itemize}

Both are \emph{showing}; naming the goal (telling) reaches neither. So
the split --- \(E\) against \(T\) --- sets \emph{how far} steering can
move the mean; curvature sets \emph{whether the showing recipe
concentrates on one bin or spreads across many.}

\textbf{More than one property.} Take two properties \(A\) and \(B\),
and mark each bin as a point on a plane --- its mean of \(A\) across,
its mean of \(B\) up. A recipe's batch lands at the blend of those
points, so its reachable averages fill the \textbf{convex hull} of the
bins. The goal is usually to span the \textbf{Pareto front} --- the
trade-off edge of non-dominated points --- which is a coverage goal, so
it needs a mixture of \emph{specialist} bins (showing): a single bin is
just one point on the edge. The one escape is within-bin spread --- a
single bin's draws fan out into an ellipsoid, and if that reaches the
front's corners it can cover the edge alone; but the wider the front,
and the more properties you stack, the less one ellipsoid reaches, so
specialist bins (showing) become the only way. (The crossover, and how
it tightens with the number of properties, is in the Appendix.)

\begin{figure}
\centering
\includegraphics[width=0.72\linewidth,height=\textheight,keepaspectratio,alt={Two properties. Each bin is a point in the (mean-A, mean-B) plane; a recipe's batch average is a blend of those points, so its reachable averages fill the convex hull (shaded). Spanning the Pareto front --- the non-dominated upper-right edge --- takes a mixture of specialist bins (showing); a single bin is one point, and its within-bin spread (the ellipse) reaches the corners only when the front is narrow. (Schematic --- illustrative, not measured data.)}]{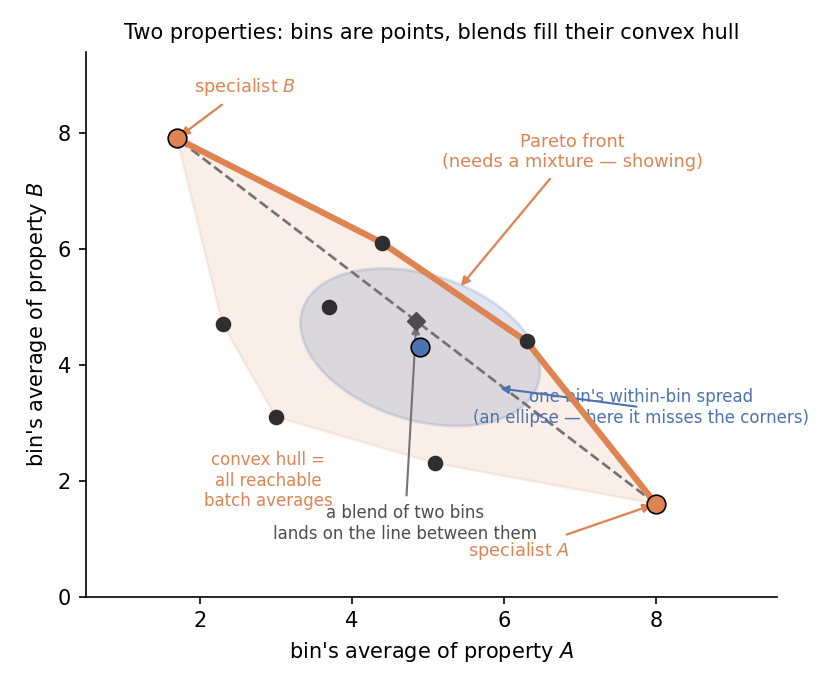}
\caption{\textbf{Two properties.} Each bin is a point in the
(mean-\(A\), mean-\(B\)) plane; a recipe's batch average is a blend of
those points, so its reachable averages fill the \textbf{convex hull}
(shaded). Spanning the \textbf{Pareto front} --- the non-dominated
upper-right edge --- takes a mixture of \emph{specialist} bins
(showing); a single bin is one point, and its within-bin spread (the
ellipse) reaches the corners only when the front is narrow. (Schematic
--- illustrative, not measured data.)}
\end{figure}

\begin{center}\rule{0.5\linewidth}{0.5pt}\end{center}

\subsection{3. What follows: four falsifiable
claims}\label{what-follows-four-falsifiable-claims}

The framework is not vacuous; a few sharp, falsifiable claims fall out,
and \textbf{Evidence} tests each across two unrelated domains.

\begin{enumerate}
\def\labelenumi{\arabic{enumi}.}
\tightlist
\item
  \textbf{The split is exact, and it carries to the model.} \(T + E =
  \mathrm{Var}_{\mathcal{J}}(L)\) to machine precision for any
  \((L, \pi, \mathcal{J})\), and the split moves with \(\pi\) as
  described in §1. It is not only a fact about the data: for a target
  the model reproduces well, the model's own outputs carry the same
  split, so the data-side budget describes the model too. But the model
  is never trained on your target \(L\) --- only to reproduce the data
  as a whole --- so a target carries faithfully only insofar as the
  model reproduces \emph{it}, and that varies: geometry rides the
  training objective closely, a sharp energy loosely. Auditing a model
  therefore means asking \emph{how well it reproduces the particular
  target you chose}. This is the framework's one \textbf{empirical}
  input, which \textbf{Evidence} tests across easy-to-hard targets, with
  the diagnostics --- drift \(\delta\), spread ratio \(\rho\) --- read
  per target.
\item
  \textbf{The two reaches touch orthogonal budgets} --- telling is
  confined to the within-bin budget \(T\), showing reaches the
  between-bin budget \(E\); each bounded by
  \(\sqrt{\chi^2\cdot\text{budget}}\) and computable before a model
  exists.
\item
  \textbf{A goal's curvature sets the shape of the showing recipe} (§2).
  Convex/linear goals --- batch averages, rates, conjunctions --- peak
  at one bin, so showing \textbf{concentrates} \(\mu_P\) there (the bin
  the audit finds); concave goals --- coverage, diversity --- peak at a
  blend, so showing \textbf{spreads} \(\mu_P\) across bins, and no
  single bin reaches them. Both are showing; \emph{naming} the goal
  (telling) reaches neither. For several properties at once the
  reachable averages are the convex hull of the bins, and covering a
  trade-off frontier is such a spread goal. (The Intro's
  within-bin/between-bin split, made quantitative.)
\item
  \textbf{Training reshapes the budget --- predictably.} Plain
  likelihood training aims only to reproduce the data --- exact for the
  crystal LM, a variational bound for the diffusion image model --- and
  its objective is minimized when the model's conditional \emph{is} the
  data's, so at the optimum the budget is the data's, unchanged.
  Fine-tuning deliberately departs from that: filtering or RLHF
  reweights \(\mathcal{J}\) into a reshaped \(\mathcal{J}'\), and the
  new per-bin numbers are computable from the old ones and the recipe
  --- so recompute the budget from \(\mathcal{J}'\) and you predict the
  tuned model's reaches without ever sampling it. One boundary: a
  fine-tune can pull so far from \(\mathcal{J}\) that nothing data-side
  is left to recompute from --- there, audit the model's own outputs
  instead. (The machinery and the reshaped \(\mathcal{J}'\):
  \textbf{Appendix}.)
\end{enumerate}

\begin{center}\rule{0.5\linewidth}{0.5pt}\end{center}

\subsection{4. One mechanism, four
phenomena}\label{one-mechanism-four-phenomena}

The budget also explains a familiar frustration: the harder we tune a
model to score well, the more variety it loses --- and the same
mechanism appears to run under several named problems at once.

Take an image generator and fine-tune it for looks --- the standard
RLHF/DPO move. Underneath, the move is precise: it multiplies each
output's share of the distribution by a factor set by that output's
score, \[\mathcal{J}'(D) \;\propto\; \mathcal{J}(D)\, e^{R(D)/\beta},\]
where \(R(D)\) is the score an output earns --- here, its aesthetics ---
and \(\beta\) is how hard you push (smaller \(\beta\), harder push).
Outputs that score well get scaled up; the rest get scaled down. (This
is the same reshaped \(\mathcal{J}'\) that Claim 4 routes fine-tunes
through.)

Watch what that does to our columns of bins. \emph{Within} ``golden
retriever,'' the best-looking retriever swells and the rest fade --- the
column narrows. \emph{Across} classes, the photogenic ones (lakes,
sunsets) grow while the drab ones (parking lots) shrink --- the lineup
of columns narrows too. Push harder (\(\beta \to 0\)) and it ends in one
place: a single image, the highest-scoring one, everything else gone.
Both kinds of spread have collapsed to nothing --- the within-bin \(T\)
\emph{and} the between-bin \(E\).

And here is the crux. A reward scores \textbf{one output at a time}, and
variety is not a property of any one output --- it lives in the
\emph{collection}. So a per-output objective has nothing in it that can
even \emph{see} spread; it can only concentrate. To produce that spread
you must act on the collection itself --- choose the mix of bins. That
is exactly what \emph{showing} does: its recipe \(\mu_P\) is the only
lever on the between-bin budget \(E\) from §2 --- the part a per-output
objective can never reach. Tuning collapses that spread; only showing
realizes it.

One exception proves the rule. Objectives that score a whole \emph{set}
--- contrastive losses, batch-diversity penalties --- genuinely
\emph{can} reward spread. But standard RLHF and DPO do not; and even a
contrastive batch is a \emph{shuffled random} sample pushed apart, not a
\emph{designed} mix of bins --- it sees a batch without choosing the
collection, which is what showing does.

So one trade sits under four familiar problems --- each buys per-output
quality by collapsing collection spread a per-output method cannot
recover:

\begin{itemize}
\tightlist
\item
  \textbf{Mode collapse} --- collection spread driven to the floor; the
  model parrots a handful of outputs.
\item
  \textbf{The alignment tax} --- the diversity a tuned model loses
  \emph{is} the spread it traded away for polish.
\item
  \textbf{Reward hacking} --- hand it a reward to chase and it sprints
  to the single highest-scoring output.
\item
  \textbf{Compositional generalization} --- asking for two things at
  once (\emph{A and B}) needs outputs that already have both, and
  concentration only narrows. Showing can find the joint by drawing from
  the bins that hold it; per-output tuning never reaches it.
\end{itemize}

Read this way, the four are one mechanism wearing four names:
collection-level spread that per-output training collapses and only
showing restores. Each still has its own familiar driver --- the
architecture, the data, a badly-chosen reward --- and those hold; what
the framework adds is \emph{a} common cause beneath all four --- the
between-bin spread they narrow and cannot recover, which only showing
reaches; an interpretation the framework invites.

\begin{center}\rule{0.5\linewidth}{0.5pt}\end{center}

\subsection{5. Scope --- where this
applies}\label{scope-where-this-applies}

We want to be precise about where the framework applies and where it
does not. It applies to the \textbf{mean shift of a labeled property,
achieved by selection, in a model that matches its data}
(\(\mathcal{J}\), or the reshaped \(\mathcal{J}'\) after a standard
fine-tune).

There are degenerate extremes where it does not. The bound is a
\emph{forecast}: a real model can fall short of it (contraction) or, off
the data's support, exceed it (expansion). And it bounds the \emph{size
of a mean shift} --- not the literal extremes of a property, and not the
\emph{value} of a set-level goal like coverage or diversity. For those
goals the framework still says \emph{which} operation reaches them --- a
mixture, showing (§2) --- just not how large the metric gets. The full
list of edges --- off-support (\(\chi^2 \to \infty\)), heavy-tailed bin
means, the steered-batch variance, multi-property targets, leaving the
selection regime --- sits in the \textbf{Appendix}, each with a cheap
check.

\begin{center}\rule{0.5\linewidth}{0.5pt}\end{center}

\subsection{Pointer}\label{pointer}

For practical use: \textbf{Application}. For empirical verification of
the claims: \textbf{Evidence}. For the shared setup and notation:
\textbf{Preliminaries}. For the detailed machinery (data-to-model
bridge), the case-by-case reach of each operation, and the full boundary
statement: the \textbf{Appendix}.

\section{Evidence --- The same claims, in two unrelated
domains}\label{evidence-the-same-claims-in-two-unrelated-domains}

\textbf{Theory} made its predictions on paper. This section asks whether
real generative models, on real data, obey them.

The theory hands us a \emph{budget}: for any target, the total variation
splits into a piece that \emph{telling} can reach and a piece that only
\emph{showing} can. What we don't yet know is whether any of it survives
a real model --- whether the split leaks, whether a model reaches past
its budget, whether fine-tuning breaks the accounting. To keep any one
result from being a quirk of a single setup, we test in two domains that
share no model, no data, no labels, and no scoring code. A prediction
that holds in both is hard to explain away; it reads as a property of
the framework.

Vocabulary from \textbf{Preliminaries}: a cheap \emph{key} sorts each
output into a \emph{bin}; the \emph{within-bin budget} \(T\) is what
\emph{telling} reaches, the \emph{between-bin budget} \(E\) what
\emph{showing} reaches.

\begin{center}\rule{0.5\linewidth}{0.5pt}\end{center}

\subsection{The two test beds}\label{the-two-test-beds}

\textbf{Images --- the legible track.} A pretrained DiT-XL/2-256 (a
diffusion model that turns a class label into an image), used as-is. The
key is the class a frozen ResNet-50 reads off each image, so each of
ImageNet's 1000 classes is a bin. We steer seven plain measurements:
\emph{brightness}, a learned \emph{aesthetic} score, compressed
\emph{file size per megapixel} (a stand-in for visual busyness), and
four CLIP \emph{content scores} --- how well an image matches ``a photo
of an animal / a vehicle / food / nature.'' You can inspect any bin's
images directly.

\textbf{Crystals --- the consequence track.} A pretrained
encoder--decoder that generates a crystal structure to meet requested
properties, trained on data derived from Alexandria, a large database of
computed materials. The key is chemistry --- the element set and atom
count, read off the output. We steer band gap, energy above hull
(distance from the stability frontier; zero is stable), density,
dielectric constant, and one collection-level score --- \emph{Combined},
the fraction of generated structures that are at once novel, unique, and
at least metastable.

\textbf{Cost.} The image study runs in a few hours on a single GPU. The
data-side claims (Claim 1, the Claim 2 bounds) are cheaper still: they
need only the audit's \emph{shadow} --- per-bin share, mean, and
variance, 132 KB in all --- and reproduce with \texttt{numpy} alone, no
model and no ImageNet. The companion repo ships it;
\texttt{verify\_dataside.py} checks them in under a second.

\begin{center}\rule{0.5\linewidth}{0.5pt}\end{center}

\subsection{What we set out to
establish}\label{what-we-set-out-to-establish}

Four claims, each tested in both domains:

{\def\LTcaptype{none} 
\begin{longtable}[]{@{}
  >{\centering\arraybackslash}p{(\linewidth - 2\tabcolsep) * \real{0.0411}}
  >{\raggedright\arraybackslash}p{(\linewidth - 2\tabcolsep) * \real{0.9589}}@{}}
\toprule\noalign{}
\begin{minipage}[b]{\linewidth}\centering
\#
\end{minipage} & \begin{minipage}[b]{\linewidth}\raggedright
Claim
\end{minipage} \\
\midrule\noalign{}
\endhead
\bottomrule\noalign{}
\endlastfoot
1 & The budget splits exactly, the split moves as you change the bins,
and it carries to the model \\
2 & Telling and showing reach different parts of the budget --- and we
measure how much further showing reaches than a knob (bins vs knobs) \\
3 & An average over a batch is reached by concentrating on one bin; a
spread goal only by a mix of bins --- both are showing, while naming
(telling) reaches neither \\
4 & Training changes the budget, but in a way you can predict \\
\end{longtable}
}

The rest takes them one at a time, images first. The headline ---
\emph{how much further} showing reaches than the best knob, and the
between-bin share that says when --- lives inside Claim 2, where we pit
bins against the strongest knob each domain allows, property by
property.

\begin{center}\rule{0.5\linewidth}{0.5pt}\end{center}

\subsection{Claim 1 --- The budget splits exactly, moves with the bins,
and carries to the
model}\label{claim-1-the-budget-splits-exactly-moves-with-the-bins-and-carries-to-the-model}

Three checks: the split is real, it moves with the bins, and it survives
the model.

\subsubsection{1. The split is exact}\label{the-split-is-exact}

Total spread = spread inside the bins + spread between them, exactly.

\emph{Images}, brightness, bins the 1000 classes. Total variance 0.0181;
within-class (one photo in sun, another in shade) averages 0.0151;
between-class (snowfields brighter than coal) 0.0030. They sum to
0.0181.

\emph{Crystals}, energy-above-hull, bins the chemistry. Over 225,000
structures the total is 584 (meV/atom); within-chemistry spread is 22,
between chemistries 562. They sum to 584.

This is the \textbf{law of total variance}, an identity --- so the exact
match is not evidence for anything; it just confirms we measured the
three numbers right. Claim 1's empirical weight is in the next two parts
--- does the split \emph{move} as predicted, and does it \emph{survive
the model} --- and in Claims 2--4.

\subsubsection{2. Coarser bins shift the
split}\label{coarser-bins-shift-the-split}

Make the bins coarser and variation moves from between to inside, as
predicted.

\emph{Images}: merge the 1000 classes at random into 50 --- within rises
0.0151 → 0.0179, between falls 0.0030 → 0.0002, total unchanged.
\emph{Crystals}: drop the atom count, keep only the element set ---
within 22 → 120, between 562 → 464, total unchanged.

\begin{figure}
\centering
\includegraphics[width=0.5\linewidth,height=\textheight,keepaspectratio,alt={Claim 1: the brightness budget, fine bins versus coarse --- the split is exact in both, and coarsening shifts it toward within-bin.}]{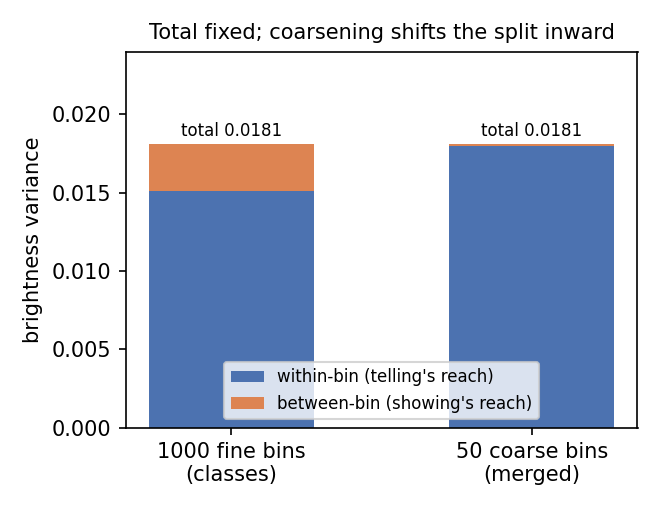}
\caption{Claim 1: the brightness budget, fine bins versus coarse --- the
split is exact in both, and coarsening shifts it toward within-bin.}
\end{figure}

That the split \emph{moves} with the bins raises a fair worry: is the
budget then just an artifact of an arbitrary key? No --- it tracks
\emph{meaningful} structure, not the bin count. Recomputing the image
split under several keys, the between-bin share's ordering across
targets is stable from 1000 classes down to a structured 50-bin grouping
(class-aligned targets stay high, pixel-level targets stay low); only a
\emph{random} 50-bin key collapses the between-bin share to near zero
for every target. A structured and a random key of the \emph{same}
50-bin size differ roughly eighteen-fold --- so what the budget reads is
the key's structure, not its resolution. (Same on crystals: the
carry-over ordering below holds across element-set and atom-count keys.)

\subsubsection{3. Does the split carry through the
model?}\label{does-the-split-carry-through-the-model}

The split is a fact about the data; it matters only if the model
reproduces it. A model can miss two ways: a bin lands in the wrong place
(its average is off --- the \emph{drift} δ), or it holds the wrong
amount of spread inside (the \emph{spread ratio} ρ). We checked both.

\textbf{Do the bin averages land where the data says?} Only when the
property is close to what the model was trained to produce. The model
isn't trained on these properties --- the crystal model learns to get a
structure's tokens right, DiT to get pixels right --- so a property
comes out right only if getting the tokens (or pixels) right forces it
to. Per-bin agreement (correlation between the model's per-chemistry
averages and the data's), with 95\% confidence intervals from
bootstrapping over chemistries and three generation seeds:

\begin{itemize}
\tightlist
\item
  density \textbf{0.985} {[}0.91, 1.00{]} --- fixed by the structure's
  shape, which the tokens fix, so it carries tightly;
\item
  bonding plausibility \textbf{0.879} {[}0.04, 0.97{]} --- the
  intermediate rung, and its wide interval says so: the point estimate
  lands between density and stability as predicted, but few ionic
  chemistries enter its fit, so bonding alone is pinned down only
  loosely;
\item
  stability \textbf{0.488} {[}0.17, 0.71{]} --- a few stray atoms flip a
  structure off the frontier, so the tokens pin it down only loosely;
  its agreement is both lower and far less certain.
\end{itemize}

\begin{figure}
\centering
\includegraphics[width=0.6\linewidth,height=\textheight,keepaspectratio,alt={Claim 1: how well the model's per-chemistry averages match the data's --- 0.99 for density, 0.88 for bonding, 0.49 for stability, with 95\% CIs (1 is exact, 0 is none).}]{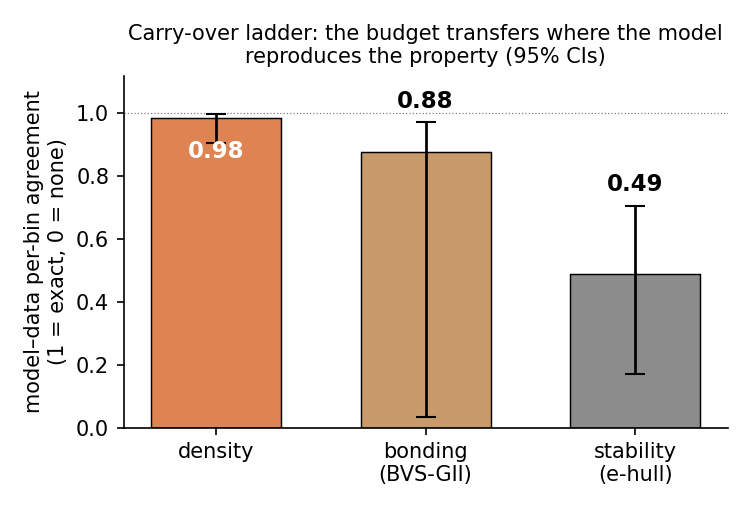}
\caption{Claim 1: how well the model's per-chemistry averages match the
data's --- 0.99 for density, 0.88 for bonding, 0.49 for stability, with
95\% CIs (1 is exact, 0 is none).}
\end{figure}

And we did not only \emph{measure} this ordering --- we
\textbf{predicted it before running the audit.} From a small independent
pilot, using only cheap pre-measurement signals (how often the model's
outputs even land in the data's bins, plus the early drift δ and spread
ratio ρ), we committed in advance --- timestamped in the repo --- to the
ordering density \textgreater{} bonding \textgreater{} stability,
density above \textasciitilde0.9 and stability below \textasciitilde0.4.
The converged audit confirmed the ordering exactly (0.99 / 0.88 / 0.49);
the magnitude calls were close, stability landing at 0.49 --- a touch
above its \textasciitilde0.4 band, the one number that overshot. The
ordering rests on the point estimates, which are cleanly ranked; the two
ends --- density high, stability low --- are firm, while the middle rung
(bonding) carries a wide interval and is the least certain of the three.
So carry-over is a \emph{prediction the data makes ahead of time}, not a
pattern read off afterward --- which is exactly what lets you trust the
budget on a property \emph{before} you have measured it on the model.

So where the agreement is high and tight --- density, and brightness on
images --- the budget you read off the data is the one the model
produces; steer by it. Where it is low and uncertain --- stability ---
the data audit is an unreliable guide, so measure the model directly
instead. And stability's low score is the property genuinely failing to
carry, not a noisy yardstick: re-scoring it with a cleaner ensemble
label leaves the agreement essentially unchanged (0.42 vs 0.49), and
restricting to the chemistries whose labels are most reliable does not
raise it.

\textbf{Does each bin hold the right variety?} This is the room
\emph{telling} works in. On images the model produces 0.96--1.14× the
data's within-bin spread (spread ratio ρ ≈ 1), across every target ---
so \emph{telling}'s part of the budget is real on the model, not just on
paper. And the bin averages track too: for class-aligned targets ---
nature, animal, food --- the model's per-class means correlate with the
data's at r² 0.85--0.94 (the higher-spread brightness and aesthetic
targets sit lower, 0.48--0.59, and need more samples to pin down).

The two checks are independent: one crystal model hit the right
chemistries but filled them with unstable structures (high stability
drift). Check both before trusting the budget on a new model.

\begin{center}\rule{0.5\linewidth}{0.5pt}\end{center}

\subsection{Claim 2 --- Telling and showing reach different parts of the
budget}\label{claim-2-telling-and-showing-reach-different-parts-of-the-budget}

Telling reweights \emph{within} a bin; showing reweights \emph{across}
them. Two things follow --- one structural, one we measured.

\textbf{Images.} The experiment: take one target --- the animal score
--- and try to move its batch average two ways, generating a batch each
way and measuring how far the average moved from the model's default. To
\emph{tell}, fix a class and turn classifier-free guidance up as hard as
it goes (staying inside that one bin). To \emph{show}, draw the batch
from the classes whose animal-averages are highest. The figure plots how
far each got:

\begin{figure}
\centering
\includegraphics[width=0.8\linewidth,height=\textheight,keepaspectratio,alt={Claim 2: how far telling and showing can move a target's average from where it starts --- telling within one bin's width, showing across the whole span between bins.}]{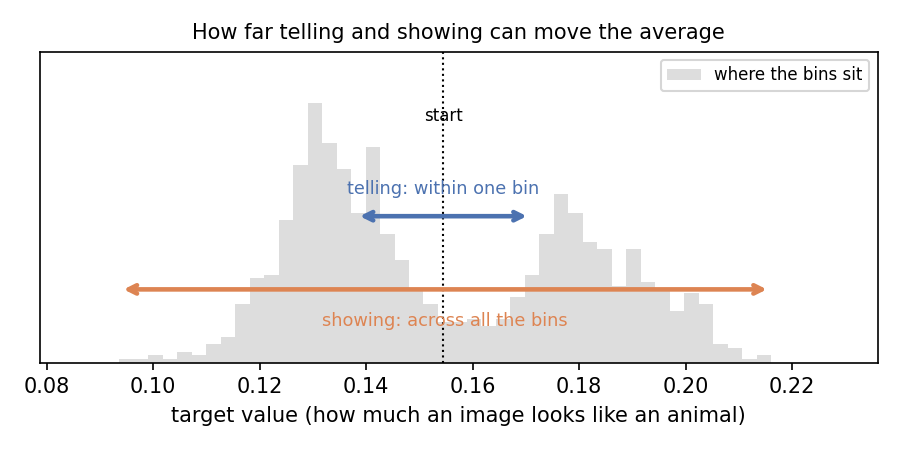}
\caption{Claim 2: how far telling and showing can move a target's
average from where it starts --- telling within one bin's width, showing
across the whole span between bins.}
\end{figure}

\textbf{Structurally}, telling cannot leave the bin it is in, so it can
only reach the within-bin budget \(T\); showing moves mass across bins,
so it reaches the between-bin budget \(E\). Same model, same target ---
different budget.

\textbf{And the \emph{effort} each spends differs by two orders of
magnitude.} We measured the χ² each operation actually puts in: telling
pushed to its limit (classifier-free guidance swept up to CFG = 15)
moves the within-bin distribution by a χ² of only about \textbf{1 to
1.6}, while the exemplar recipes below reach χ² of order \textbf{100}.
At every guidance scale the realized shift stayed under telling's own
within-bin ceiling √(χ²·T) --- the exact counterpart of showing's
√(χ²·E) --- so √T is not a wall telling hits, just the χ² ≈ 1 scale it
can afford. \textbf{In one line: both obey the same √(χ²·budget) law;
telling is confined to the small budget T and can spend only χ² ≈ 1,
showing reaches the large budget E and spends χ² ≈ 100 --- and that is
the whole of the asymmetry.}

Showing's reach also grows the harder you concentrate the recipe.
Drawing from the top 400, then 100, 25, and 8 of the 1000 classes, the
realized shift climbed --- animal score 0.031 → 0.051 (95\% CI {[}0.029,
0.034{]} to {[}0.049, 0.053{]}), brightness 0.036 → 0.118 ({[}0.014,
0.059{]} to {[}0.085, 0.151{]}) --- and every recipe stayed under the
budget's ceiling, √(χ²·E), with χ² measuring how far the recipe departs
from the data's own mix of bins. The ceiling rises faster than the
reach, so here it is a true but \emph{loose} cap: the top bins, weighted
bluntly, don't sit far enough apart to use all the room.

\begin{figure}
\centering
\includegraphics[width=0.85\linewidth,height=\textheight,keepaspectratio,alt={Claim 2: as the recipe concentrates (χ² rising), showing's realized shift grows and stays under the budget's ceiling √(χ²·E). The ceiling climbs faster, so the cap is loose at high concentration; for brightness the model also falls short of the shift the budget allows.}]{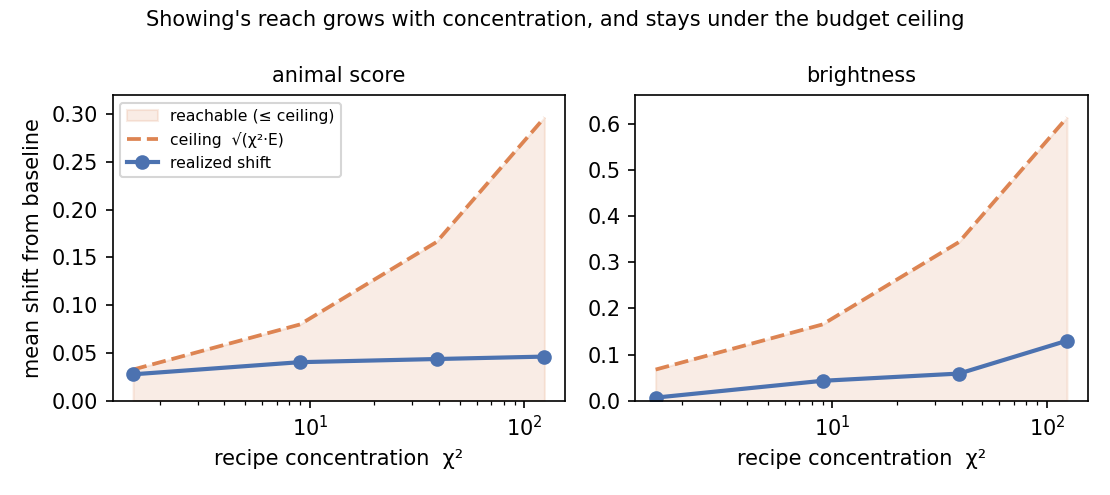}
\caption{Claim 2: as the recipe concentrates (χ² rising), showing's
realized shift grows and stays under the budget's ceiling √(χ²·E). The
ceiling climbs faster, so the cap is loose at high concentration; for
brightness the model also falls short of the shift the budget allows.}
\end{figure}

But the ceiling is loose only because the blunt top-\emph{k} recipe
spends its χ² inefficiently. There is a recipe that spends it optimally
--- the \textbf{min-χ²} direction, which puts exactly the bin weights
the Cauchy--Schwarz bound is tight for --- and there the ceiling stops
being a cap and becomes a \emph{forecast}. Steering one target with that
recipe at a small shift, the realized move (0.0162) landed on the
predicted ceiling (0.0195), inside the bootstrap CI {[}0.0114,
0.0209{]}, while the blunt top-\emph{k} recipe at the same χ² reached
only a fraction of its ceiling. So the bound has two faces: a loose cap
when you concentrate crudely, a tight prediction when you steer
efficiently. \emph{(At larger shifts the min-χ² weights hit the
non-negativity floor and the realized move falls back below the ceiling;
the tight regime is the small-shift, in-simplex one.)}

\begin{figure}
\centering
\includegraphics[width=0.7\linewidth,height=\textheight,keepaspectratio,alt={Claim 2: the min-χ² recipe turns the ceiling into a forecast --- at a small target shift the realized move meets the predicted ceiling √(χ²·E) (its CI includes the bound), and falls below only once the recipe saturates against the non-negativity floor.}]{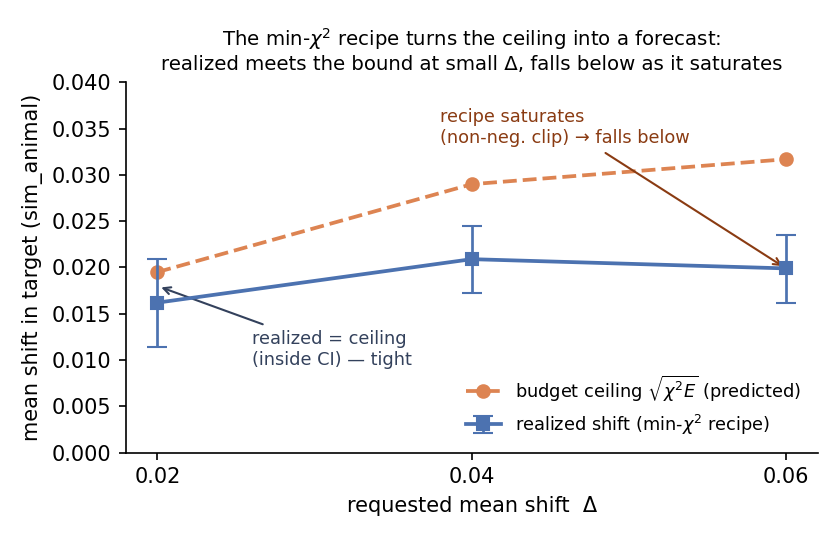}
\caption{Claim 2: the min-χ² recipe turns the ceiling into a forecast
--- at a small target shift the realized move meets the predicted
ceiling √(χ²·E) (its CI includes the bound), and falls below only once
the recipe saturates against the non-negativity floor.}
\end{figure}

\textbf{Crystals.} Telling means pushing within one chemistry --- there
\emph{are} knobs (the property-tag conditioning the model was trained
with, e.g.~a ``high band gap'' or ``stable'' tag, turned up like a
guidance scale), but once the chemistry is fixed they barely move the
result: the chemistry has already set most of what you get, and the tags
only nudge within it. Showing means choosing which chemistries to draw
from, and it dials the \emph{Combined} score smoothly from 0.2 to 0.8
--- the gap between the two reaches, as wide as the framework predicts.

\subsubsection{How much further: bins versus knobs, and where the
crossover
falls}\label{how-much-further-bins-versus-knobs-and-where-the-crossover-falls}

Start with the fair, knob-proof number. At \textbf{matched effort}
telling can spend what it has only inside the bins and reaches at most
\(\sqrt{\chi^2\,T}\); showing spends the same effort across bins and
reaches \(\sqrt{\chi^2\,E}\). The common effort cancels, so their ratio
is \(\sqrt{E/T}\) --- fixed by the data before any knob exists, and
something no knob can argue its way out of. For the three crystal
targets it is \textbf{6.8} (band gap), \textbf{3.2} (density) and
\textbf{1.1} (stability): effort for effort, showing reaches several
times further, and most where the budget is most between-bin. That is
the ceiling. What a \emph{real} knob realizes is a separate question ---
and it is the one a skeptic presses: maybe the knobs we tried were weak.
So we pushed the knob as hard as each domain allows.

\textbf{Crystals.} The deployed property tag is a \emph{weak} knob: fix
the chemistry and it realizes almost no effort (\(\chi^2\approx1\)), so
the realized gap is even wider than the structural ceiling ---

{\def\LTcaptype{none} 
\begin{longtable}[]{@{}lcccc@{}}
\toprule\noalign{}
target & \(E/(E{+}T)\) & knob shift & bin shift & realized bin / knob \\
\midrule\noalign{}
\endhead
\bottomrule\noalign{}
\endlastfoot
band gap & 0.98 & 0.003 & 2.71 & \textbf{≈800×} \\
density & 0.96 & 0.09 & 8.40 & \textbf{≈91×} \\
stability (metastable rate) & 0.57 & 0.009 & 0.385 & \textbf{≈41×} \\
\end{longtable}
}

--- every knob's 95\% CI through zero, every bin's clear of it. (Because
the knob moves essentially nothing, its CI straddling zero, the realized
\emph{ratio} has a near-zero denominator and is unstable --- read it as
``the tag does nothing,'' and trust the knob-free \(\sqrt{E/T}\) column
above as the equal-effort number.) But a tag is not the strongest knob a
chemistry allows. The strongest is \textbf{best-of-twelve}: generate a
dozen candidates in a \emph{fixed} chemistry and keep the best by the
property scorer --- the largest reweighting a per-output selector can
do, at matched effort \(\chi^2\approx5.3\). Even this loses, by
\textbf{26× / 14× / 4.8×} on gap / density / stability. On crystals the
sweep holds from the weak tag all the way up to the strongest knob that
stays inside a bin.

\textbf{Images.} Here the strongest knob is more powerful, and it
changes the story --- in the framework's favour. We built a
\textbf{learned soft-prompt}: a continuous conditioning vector
warm-started at the best class and optimised directly against the target
scorer, the strongest single-conditioning knob we can imagine. Against
it the contest turns \emph{clean} --- the bin/knob ratio now rises
\textbf{monotonically}:

{\def\LTcaptype{none} 
\begin{longtable}[]{@{}
  >{\raggedright\arraybackslash}p{(\linewidth - 6\tabcolsep) * \real{0.2500}}
  >{\centering\arraybackslash}p{(\linewidth - 6\tabcolsep) * \real{0.2500}}
  >{\centering\arraybackslash}p{(\linewidth - 6\tabcolsep) * \real{0.2500}}
  >{\centering\arraybackslash}p{(\linewidth - 6\tabcolsep) * \real{0.2500}}@{}}
\toprule\noalign{}
\begin{minipage}[b]{\linewidth}\raggedright
target
\end{minipage} & \begin{minipage}[b]{\linewidth}\centering
\(E/(E{+}T)\)
\end{minipage} & \begin{minipage}[b]{\linewidth}\centering
\(\sqrt{E/T}\) (equal-effort)
\end{minipage} & \begin{minipage}[b]{\linewidth}\centering
bin / strongest knob (realized)
\end{minipage} \\
\midrule\noalign{}
\endhead
\bottomrule\noalign{}
\endlastfoot
brightness & 0.17 & 0.45 & \textbf{0.29} \\
aesthetic score & 0.27 & 0.61 & \textbf{0.41} \\
vehicle content & 0.57 & 1.15 & 3.0 \\
animal content & 0.72 & 1.60 & 2.9 \\
\end{longtable}
}

With a strong enough knob, \textbf{two targets flip}: on brightness and
aesthetic --- the two lowest-\(E/(E{+}T)\) scalars --- the learned knob
\emph{beats} bins, while on the two high-share semantic targets it still
loses by about three. But it wins only by \textbf{leaving its bin}: on
those two targets the winning images land almost entirely outside the
class they started from. So the experiment lets us \emph{construct} a
case where a knob wins, but only by breaking the bin barrier that
defines telling in the first place.

One caveat keeps this honest. DiT's guidance conditions on the ImageNet
class --- which \emph{is} our key \(\pi\) --- so on images ``telling
stays inside its bin'' is close to true by construction, and the image
contest is best read as illustration. The crystal knob is the cleaner
test: the property tag (band-gap, stability) is a \emph{different}
handle from the chemistry key, so a tag that moves nothing while
chemistry selection moves the property is genuine evidence, not a
tautology of the setup. The two domains together --- one where the knob
and key coincide, one where they are distinct --- bound the claim from
both sides.

That is the whole result, stated cleanly. It is \textbf{not} that
examples always beat knobs. \emph{\(\sqrt{E/T}\) is the fair comparison;
best-of-twelve shows no within-bin knob escapes it; the soft-prompt
shows the only escape is to stop being a knob --- and the budget tells
you, in advance, the one regime (low \(E/(E{+}T)\)) where that escape is
worth making.} The between-bin share is a number you read off the data
before generating anything, and it names its own exceptions: brightness
and aesthetic, its two lowest values.

\subsubsection{Choosing the key}\label{choosing-the-key}

The whole budget is defined against a key \(\pi\) --- the partition into
bins --- so a fair question is how to pick one without knowing the
answer first. The same between-bin share answers it. Enumerate a few
cheap candidate keys (a classifier's classes, k-means over a cheap
feature, element-set for crystals), and for each score \(E/(E{+}T)\)
from the data alone --- no generation. The key with the largest share is
the one to use, and the score \emph{predicts} the payoff: across
candidate keys and targets, the data-side \(E/(E{+}T)\) tracks the
realised bin/knob ratio (rank correlation 0.88 on images; on crystals a
key-quality that also credits carry-over recovers the operative
element-set key), with a structureless random partition sitting at the
floor. And if \emph{no} candidate key has a large share, that is itself
the answer: the target is intrinsically knob-shaped --- which is exactly
why aesthetic and brightness, whose variance no key concentrates between
bins, are the two a strong knob wins. (This needs \emph{some} cheap
label to bin by; discovering a key with no labels at all we leave open.)

\begin{center}\rule{0.5\linewidth}{0.5pt}\end{center}

\subsection{Claim 3 --- The goal's kind sets the recipe: one bin for an
average, a mix for a
spread}\label{claim-3-the-goals-kind-sets-the-recipe-one-bin-for-an-average-a-mix-for-a-spread}

A goal is a property of the batch you generate, and the theory splits
goals in two --- and with them the \emph{shape} of the showing recipe.
Some are an \textbf{average over the batch} --- a high average score, a
high fraction clearing a bar --- and the best recipe
\textbf{concentrates} on the single best bin (which the audit must find;
it is not the average bin). Others are about the batch's \textbf{spread}
--- the batch must \emph{cover} two different kinds of output at once
--- and no single bin can do that, so the recipe \textbf{spreads} across
bins. Both are showing; \emph{naming} the goal (telling) reaches
neither. We test both kinds, in both domains, and each needs the recipe
shape the theory predicts.

\subsubsection{An averaging goal: one bin
wins}\label{an-averaging-goal-one-bin-wins}

Start with a goal that \emph{looks} like it needs a spread of bins but
doesn't. Ask for images that are \emph{both} an animal and a natural
scene --- an otter on a riverbank --- and count the fraction of the
batch that clears both bars at once. That fraction is an
\textbf{average} over the batch, so the theory says the recipe should
\emph{concentrate} --- the single best bin should win, not a broad mix.

No class label is ``animal in nature,'' so \emph{naming} the property
gets you nowhere new: asking for animals, asking for nature, or
composing the two with Composable Diffusion (the strongest way to
compose two conditions) all land near one image in seven. But the audit
points to the bins whose images are \emph{already} both --- otters,
flamingos --- and once you know them, the best move is to draw the whole
batch from the single richest one (128 images per seed, ten seeds):

{\def\LTcaptype{none} 
\begin{longtable}[]{@{}
  >{\raggedright\arraybackslash}p{(\linewidth - 4\tabcolsep) * \real{0.3333}}
  >{\centering\arraybackslash}p{(\linewidth - 4\tabcolsep) * \real{0.3333}}
  >{\centering\arraybackslash}p{(\linewidth - 4\tabcolsep) * \real{0.3333}}@{}}
\toprule\noalign{}
\begin{minipage}[b]{\linewidth}\raggedright
how we steered
\end{minipage} & \begin{minipage}[b]{\linewidth}\centering
strongly both
\end{minipage} & \begin{minipage}[b]{\linewidth}\centering
95\% CI
\end{minipage} \\
\midrule\noalign{}
\endhead
\bottomrule\noalign{}
\endlastfoot
ask for animals (naming) & 14.5\% & {[}12, 17{]} \\
ask for nature (naming) & 12.8\% & {[}11, 15{]} \\
compose both --- Composable Diffusion & 13.5\% & {[}12, 15{]} \\
mix 30 audit-chosen both-high bins & 45.0\% & {[}42, 48{]} \\
mix the top 5 & 61.0\% & {[}58, 65{]} \\
\textbf{the single richest bin (showing, concentrated)} &
\textbf{66.5\%} & {[}64, 69{]} \\
\end{longtable}
}

Naming the property fails (\textasciitilde13\%, \emph{Composable
Diffusion} --- the strongest way to compose two conditions ---
included). Showing works, and the \emph{more concentrated} the recipe
the better: thirty audit-chosen bins reach 45\%, the top five 61\%, the
single richest bin 67\%. That is the averaging prediction exactly ---
for an average, the showing recipe should collapse onto the one best
bin. It is the extreme of showing: the bin is one the audit \emph{chose}
(the richest, not the average bin), and a knob held inside any one bin
reaches only \(\sqrt{T}\). So the conjunction never needed a
\emph{spread} of bins; it is an averaging goal, and a concentrated
recipe --- one bin --- wins.

\begin{figure}
\centering
\includegraphics[width=0.7\linewidth,height=\textheight,keepaspectratio,alt={Claim 3 (averaging goal): how often an image is strongly both animal and natural. Naming the property stays near one in seven; among audit-chosen bins, concentrating the batch scores higher, peaking at the single richest bin.}]{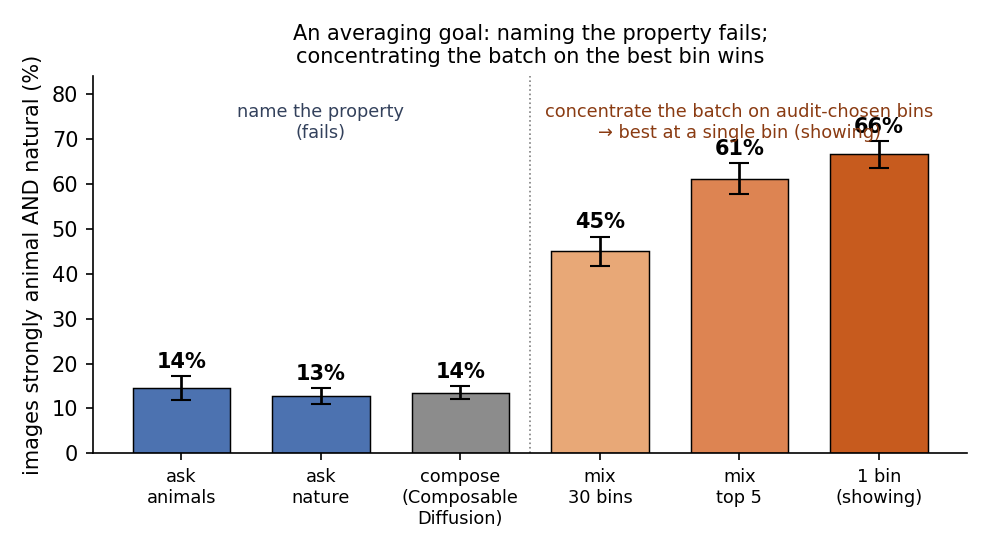}
\caption{Claim 3 (averaging goal): how often an image is strongly both
animal and natural. Naming the property stays near one in seven; among
audit-chosen bins, concentrating the batch scores higher, peaking at the
single richest bin.}
\end{figure}

(One bin wins, but you still need the \textbf{audit} to find
\emph{which} bin --- that is the showing move. Pick a class at random
and crank the knob (pure telling) and you reach only \(\sqrt{T}\), the
within-bin budget; the audit is what points the recipe at the right bin.
And which bins pay off is itself forecastable from the audit alone, by a
data-side number --- the \textbf{lift} --- with rank correlation 0.94 to
the realized rates.)

\textbf{The same averaging goal on crystals} --- a structure that is
\emph{both} wide-gap and stable --- adds one wrinkle. Three recipes at
matched sample size (joint hit = surrogate gap ≥ 3 eV \emph{and}
energy-above-hull ≤ 0.1; six seeds; 95\% CIs):

{\def\LTcaptype{none} 
\begin{longtable}[]{@{}
  >{\raggedright\arraybackslash}p{(\linewidth - 4\tabcolsep) * \real{0.3333}}
  >{\centering\arraybackslash}p{(\linewidth - 4\tabcolsep) * \real{0.3333}}
  >{\centering\arraybackslash}p{(\linewidth - 4\tabcolsep) * \real{0.3333}}@{}}
\toprule\noalign{}
\begin{minipage}[b]{\linewidth}\raggedright
how we steered
\end{minipage} & \begin{minipage}[b]{\linewidth}\centering
joint hit-rate
\end{minipage} & \begin{minipage}[b]{\linewidth}\centering
95\% CI
\end{minipage} \\
\midrule\noalign{}
\endhead
\bottomrule\noalign{}
\endlastfoot
asked naively for both (telling) & 2.7\% & {[}1.7, 4.3{]} \\
conditioned on the wide-gap prior (strong telling) & 15.5\% & {[}12.5,
19.1{]} \\
drew from wide-gap-and-stable chemistries (showing) & 13.2\% & {[}10.5,
16.5{]} \\
\end{longtable}
}

As an averaging goal should, one region wins. Mixing chemistries
(13.2\%) beats a \emph{naive} request about five-fold (2.7\%), but it
does \textbf{not} beat the \emph{strong} single request --- conditioning
on the model's own wide-gap prior, 15.5\% --- the two tie (overlapping
CIs, the gap holding at six seeds). Concentrating on the right region is
enough; spreading does not help, exactly as for an average.

\begin{figure}
\centering
\includegraphics[width=0.55\linewidth,height=\textheight,keepaspectratio,alt={Claim 3 (crystals): the joint of a wide gap and stability --- showing beats a naive request about four-fold, but ties a strong single request, because stability (the property Claim 1 found does not carry over) caps both.}]{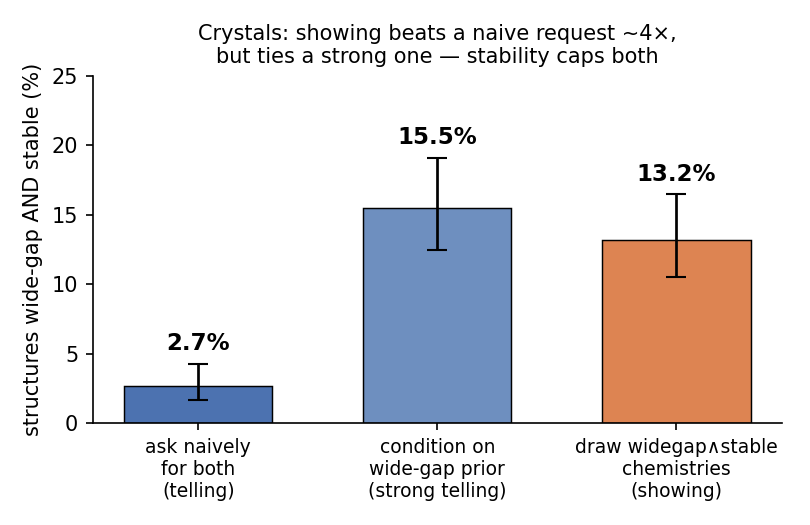}
\caption{Claim 3 (crystals): the joint of a wide gap and stability ---
showing beats a naive request about four-fold, but ties a strong single
request, because stability (the property Claim 1 found does not carry
over) caps both.}
\end{figure}

And the framework says exactly why. Showing does its job on the half it
can reach --- it produces the most wide-gap structures of any recipe.
What caps the joint is the \emph{other} half: stability sits near one in
two for both showing and strong telling. Stability is the property Claim
1 found does not carry over (agreement 0.49) --- the model can't
reliably make stable structures however you steer it --- so it becomes
the binding constraint, and the two recipes converge on it. Showing can
choose the chemistry; it cannot make the model honour a property the
model never learned to reproduce.

\subsubsection{A goal targeting spread: only showing covers
it}\label{a-goal-targeting-spread-only-showing-covers-it}

Now the goal the theory says genuinely needs showing: a batch that
\textbf{covers both corners} of two properties that pull apart --- some
outputs clearly one, some clearly the other. Score coverage as the
smaller of the two corner-fractions, so a recipe scores only if the
batch holds \emph{both}. We committed each prediction before running.

\emph{Images --- a batch with both clean vehicles and clean food} (two
concepts that genuinely exclude each other, unlike animal-and-nature;
128 images per seed, five seeds):

{\def\LTcaptype{none} 
\begin{longtable}[]{@{}
  >{\raggedright\arraybackslash}p{(\linewidth - 4\tabcolsep) * \real{0.3333}}
  >{\centering\arraybackslash}p{(\linewidth - 4\tabcolsep) * \real{0.3333}}
  >{\centering\arraybackslash}p{(\linewidth - 4\tabcolsep) * \real{0.3333}}@{}}
\toprule\noalign{}
\begin{minipage}[b]{\linewidth}\raggedright
how we steered
\end{minipage} & \begin{minipage}[b]{\linewidth}\centering
coverage
\end{minipage} & \begin{minipage}[b]{\linewidth}\centering
95\% CI
\end{minipage} \\
\midrule\noalign{}
\endhead
\bottomrule\noalign{}
\endlastfoot
ask for one (telling, one bin) & 0.00 & {[}0.00, 0.00{]} \\
compose both --- Composable Diffusion & 0.00 & {[}0.00, 0.00{]} \\
\textbf{mix a vehicle bin and a food bin (showing)} & \textbf{0.30} &
{[}0.27, 0.34{]} \\
\end{longtable}
}

\emph{Crystals --- a batch spanning the band-gap / dielectric trade-off}
(genuinely anti-correlated, \(-0.54\); pre-registered, three seeds):

{\def\LTcaptype{none} 
\begin{longtable}[]{@{}
  >{\raggedright\arraybackslash}p{(\linewidth - 4\tabcolsep) * \real{0.3333}}
  >{\centering\arraybackslash}p{(\linewidth - 4\tabcolsep) * \real{0.3333}}
  >{\centering\arraybackslash}p{(\linewidth - 4\tabcolsep) * \real{0.3333}}@{}}
\toprule\noalign{}
\begin{minipage}[b]{\linewidth}\raggedright
how we steered
\end{minipage} & \begin{minipage}[b]{\linewidth}\centering
coverage
\end{minipage} & \begin{minipage}[b]{\linewidth}\centering
95\% CI
\end{minipage} \\
\midrule\noalign{}
\endhead
\bottomrule\noalign{}
\endlastfoot
ask for one (telling, one chemistry) & 0.00 & {[}0.00, 0.00{]} \\
condition on the wide-gap prior & 0.02 & {[}0.01, 0.03{]} \\
\textbf{mix corner-A and corner-B chemistries (showing)} & \textbf{0.40}
& {[}0.35, 0.44{]} \\
\end{longtable}
}

\begin{figure}
\centering
\includegraphics[width=0.98\linewidth,height=\textheight,keepaspectratio,alt={Claim 3 (images): cover a batch with both clean vehicles and clean food. Telling sits in one corner; composing the two conditions collapses to the muddy middle (car-food chimeras, neither a clean car nor a clean meal); only showing --- a mix of a vehicle bin and a food bin --- puts genuine examples of both in the batch.}]{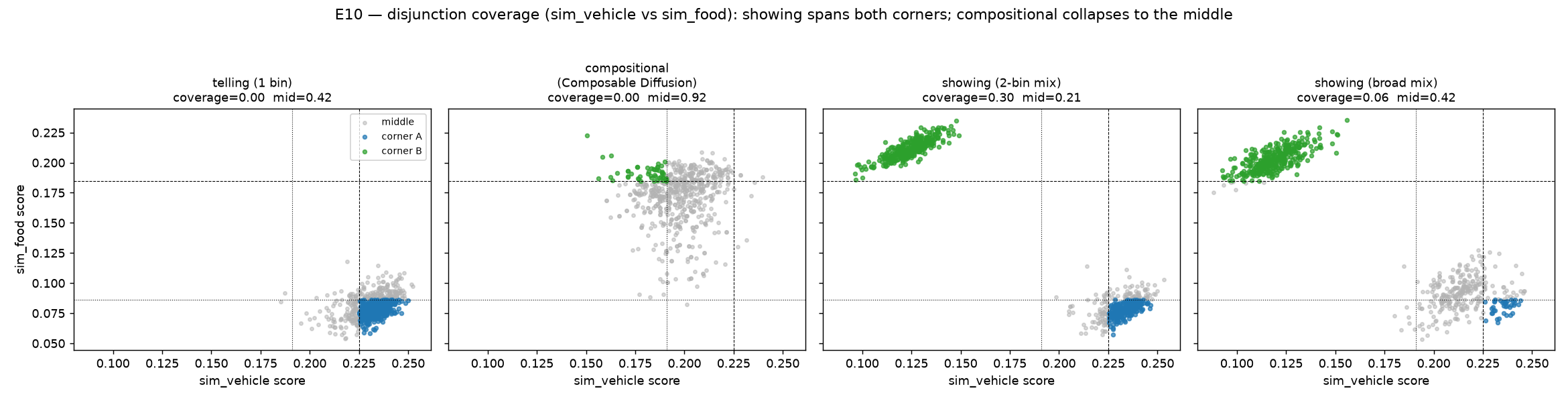}
\caption{Claim 3 (images): cover a batch with both clean vehicles and
clean food. Telling sits in one corner; composing the two conditions
collapses to the muddy middle (car-food chimeras, neither a clean car
nor a clean meal); only showing --- a mix of a vehicle bin and a food
bin --- puts genuine examples of both in the batch.}
\end{figure}

\begin{figure}
\centering
\includegraphics[width=0.8\linewidth,height=\textheight,keepaspectratio,alt={Claim 3 (crystals): cover the band-gap / dielectric trade-off. Telling lands in one corner, conditioning collapses, only showing spans both corners.}]{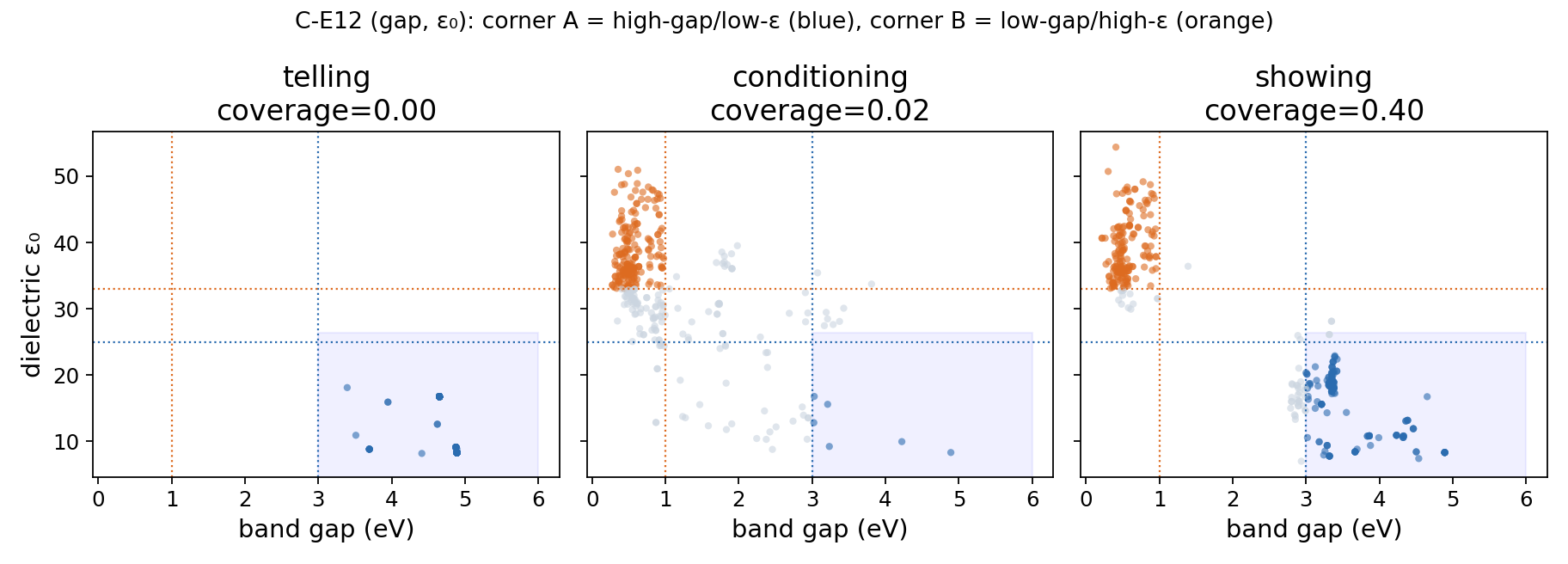}
\caption{Claim 3 (crystals): cover the band-gap / dielectric trade-off.
Telling lands in one corner, conditioning collapses, only showing spans
both corners.}
\end{figure}

Both domains land the same way, and the two scatters tell it at a
glance:

\begin{itemize}
\tightlist
\item
  \textbf{Telling sits in one corner.} A single bin is all one kind, so
  coverage is zero by construction.
\item
  \textbf{Composing the two conditions collapses to the muddy middle.}
  Asked for ``vehicle and food,'' Composable Diffusion makes car-food
  chimeras --- neither a clean car nor a clean meal (92\% of the batch
  in the middle, coverage zero); the crystal conditioning does the same.
  \emph{This is the measured finding:} the practitioner's go-to tool for
  ``both at once'' returns the \emph{average}, not the spread.
\item
  \textbf{Showing covers both corners.} A mix of a clean-corner-A bin
  and a clean-corner-B bin puts genuine examples of each into the batch
  --- coverage 0.30 on images, 0.40 on crystals --- the only recipe
  above zero, by a wide, CI-separated margin. (Precise two-corner mixing
  is the right recipe: the two cleanest bins cover better than a broad
  pool, which leaks back toward the middle.)
\end{itemize}

\textbf{And the gap widens with the number of corners.} Two corners is
the smallest spread goal. Push to three --- a batch that must hold clean
vehicles, clean food, \emph{and} a third clearly distinct kind at once
--- and the structural gap grows. Count how many corners a recipe covers
with \emph{real} mass (each corner's fraction CI-separated from zero):
telling covers exactly \textbf{one}, its own; composing conditions also
covers one (its extra corners are 1--8\% leakage, below any honest
floor); and showing covers \textbf{all \(N\)} --- two of two at \(N=2\),
three of three at \(N=3\). The gap, showing minus telling, is \(N-1\),
and it widens as you ask for more corners --- the disjunctive structure
of showing made quantitative: each added corner is one more bin a mix
can include and a single named target cannot. \emph{(The three
properties are mutually uncorrelated, so the corners are genuinely
distinct, and the clean \(N=3\) test is in images. The richest crystal
triple we could form was only about two-dimensional, so it tests
coverage but not the \(N\)-scaling cleanly; the magnitude claim rests on
the image result.)}

The two crystal goals split exactly by \emph{kind}.
Wide-gap-\textbf{and}-stable is an averaging goal --- you want a high
fraction of the batch to be both --- so the recipe should
\emph{concentrate} on one well-chosen region, and spreading across many
adds nothing; the concentrated recipe tied the strong knob. (It tied
\emph{low}, not high, because stability is the property that does not
carry to the model --- Claim 1 --- so every recipe runs into the same
stability ceiling.) The band-gap/dielectric trade-off is a spread goal
--- the batch must cover both extremes --- so only a mix of chemistries
reaches it, and there showing wins outright, the same signature as
images.

\subsubsection{The signature}\label{the-signature}

What kind of goal you have sets the \emph{shape} of the showing recipe.
An \textbf{average over the batch wants one bin}: the audit finds the
best bin and the recipe concentrates on it --- showing at its extreme
--- capped only when the property does not carry (crystal stability,
Claim 1). A goal targeting the batch's \textbf{spread wants a mix}:
covering two corners is something no single bin can do, and composing
conditions does not do it either --- it returns the average. Both are
showing --- concentrate for an average, spread for a spread;
\emph{naming} (telling) reaches neither, falling short of the best bin
for an average and collapsing to the middle for a spread.

\begin{center}\rule{0.5\linewidth}{0.5pt}\end{center}

\subsection{Claim 4 --- Training changes the budget, but in a way you
can
predict}\label{claim-4-training-changes-the-budget-but-in-a-way-you-can-predict}

Train a model only to reproduce the data and the budget is unchanged.
But fine-tuning (filtering, RLHF) deliberately changes which outputs the
model favours, so it changes the per-bin numbers the budget is built
from --- and the claim is that it changes them predictably: recompute
the budget from the data and the training recipe, and you have the
trained model's budget without ever sampling it.

Plain pretraining changes nothing, so the model's per-bin averages
should match the data's --- and they do, falling on the diagonal: 0.99
agreement for crystal density, and 0.85--0.94 for DiT's content targets,
with the fit tightening as you draw more samples (at the small sample
sizes we first used it was noise-limited; the high-spread targets need
the most samples and stay loosest).

\begin{figure}
\centering
\includegraphics[width=0.45\linewidth,height=\textheight,keepaspectratio,alt={Claim 4: predicted (data) versus observed (model) per-bin average under plain pretraining --- the points lie on the diagonal.}]{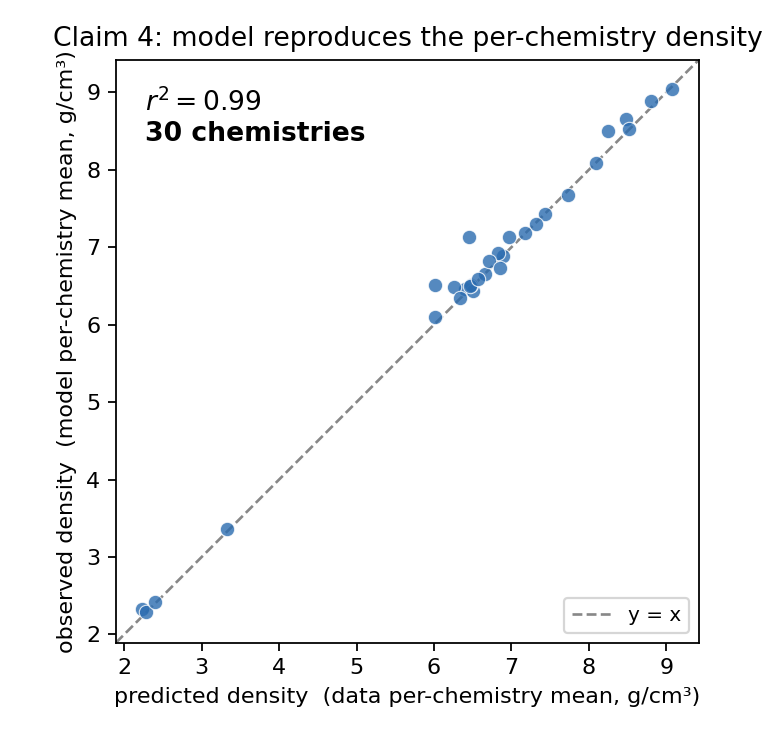}
\caption{Claim 4: predicted (data) versus observed (model) per-bin
average under plain pretraining --- the points lie on the diagonal.}
\end{figure}

The two fine-tunes we tested each change the data in a known way; we
predict the new per-bin averages from that change, then sample the model
and check.

\emph{Filtering.} Keep only outputs above a bar --- say, crystals above
a density cutoff --- and the budget recomputes on the survivors.
Predicted versus observed per-bin averages agreed to within 0.03 of a
bin width on crystals, 0.002 on images.

\emph{Guidance.} Turning up an image model's guidance scale should shift
the target's average by a growing amount. Swept over 1, 3, 7, 12 on one
class, the average climbed monotonically as predicted --- animal score
0.18 → 0.20, brightness 0.52 → 0.57.

One case breaks it. Fine-tune a model so hard it abandons its training
data and there is nothing left to recompute from. A crystal model
fine-tuned aggressively for novelty scored real, held-out structures at
−15.1 per token (95\% CI {[}−15.2, −15.0{]}) --- below the −9.7 of
random tokens, every held-out structure without exception. For a model
like that, drop the data-side prediction and audit the model's own
samples (\textbf{Application} shows how).

\begin{figure}
\centering
\includegraphics[width=0.65\linewidth,height=\textheight,keepaspectratio,alt={Claim 4: how likely the aggressively fine-tuned model thinks real, held-out structures are (per-token score) --- below the level for random tokens. It has left its training data behind.}]{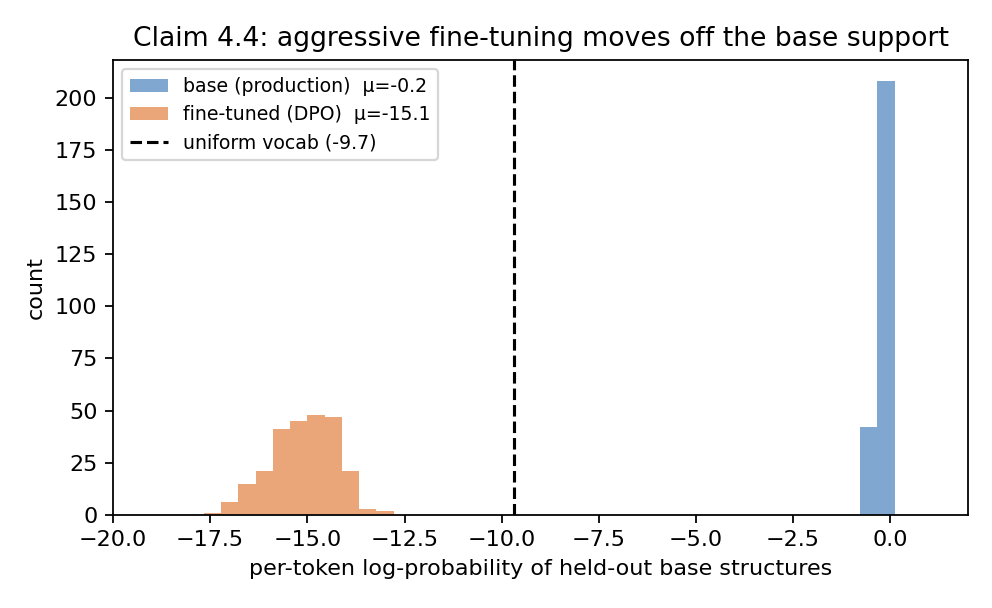}
\caption{Claim 4: how likely the aggressively fine-tuned model thinks
real, held-out structures are (per-token score) --- below the level for
random tokens. It has left its training data behind.}
\end{figure}

(Two more cases --- mixing in supervised fine-tuning data, and
distilling from a teacher --- we leave for later.)

\begin{center}\rule{0.5\linewidth}{0.5pt}\end{center}

\subsection{Robustness check --- does the audit survive losing the
training
data?}\label{robustness-check-does-the-audit-survive-losing-the-training-data}

Every claim above assumes access to \(\mathcal{J}\), the training set
itself. Preliminaries and Application already describe the fallback for
when it's gone --- a third-party checkpoint, an undocumented fine-tune,
a training set that no longer exists: pick a reference recipe
\(\mu_P^{\text{ref}}\) over the key's bins, sample many outputs from
\(\mathrm{G}\) under it, and audit that sample exactly as if it were
\(\mathcal{J}\). Until now that fallback was \emph{asserted}, not
measured --- the recipe was specified, its accuracy never tested. We
tested it directly, in both domains, holding the same ground truth used
everywhere else in this section fixed and asking how close the
reference-based audit comes to it.

\textbf{Crystals.} Ground truth (from Claim 1's carry-over check,
restricted to its 30 chemistry bins): \(T_{\mathcal{J}} = 0.092\),
\(E_{\mathcal{J}} = 3.107\), \(E/(E{+}T) = 0.971\) --- the property is
almost entirely between-bin, exactly the regime where showing dominates.
We generated fresh structures from the released model under four
reference recipes and audited each pool, binning by the chemistry
actually realized:

\begin{center}\small
\resizebox{\linewidth}{!}{%
\begin{tabular}{@{}p{5.4cm}rp{2.7cm}rp{2.7cm}rc@{}}
\toprule
$\mu_P^{\text{ref}}$ & bins & $T$ (95\% CI) & rel.\ err & $E$ (95\% CI) & rel.\ err & $E/(E{+}T)$ \\
\midrule
\textbf{B} --- uniform over the 30 training bins & 89 & 0.19 [0.10, 0.30] & 105\% & 3.13 [1.34, 4.82] & 0.8\% & 0.943 \\
\textbf{S} --- a real, skewed pool (rare-earth-heavy) & 433 & 0.036 [0.022, 0.053] & 60\% & 3.04 [2.43, 3.72] & 2.2\% & 0.988 \\
\textbf{O} --- B's pool, reweighted by the true training marginal & 89 & 0.18 & 92\% & 3.28 & 5.5\% & 0.949 \\
\textbf{U} --- a pool with no support overlap with training & 143 & 60 [2.2, 190] & 659\% & 10.6 [5.8, 55.7] & 242\% & \textbf{0.150} \\
\bottomrule
\end{tabular}%
}
\end{center}

\(E\) --- the number the telling-vs-showing call turns on --- comes back
within a few percent of truth under B, S, and O alike; even a real,
deliberately skewed guess (S) barely moves it. \(T\) is noisy under all
three (B and O's CIs comfortably contain the truth; S's is biased low
but modest), but \(T_{\mathcal{J}}\) is small to begin with, so the
noise is on a small number. \textbf{U is the one that matters}: a
reference pool built from a chemistry region with essentially no overlap
with what the model trained on collapses \(E\)'s recovery outright and
flips the qualitative call --- \(E/(E{+}T)\) falls from \(0.97\) to
\(0.15\), which would tell a practitioner \emph{telling} is competitive
when it plainly is not.

A second question, since bin count differed sharply across conditions
(89 vs.~433 realized bins): is \(T\)'s blow-up under B a fragmentation
artifact of binning by realized chemistry rather than the intended one?
We re-aggregated the same generated structures by their \emph{nominal}
(requested) bin instead --- no new generation --- for B and S (U is
wrong-support by design, so re-binning it tests nothing new; O reuses
B's pool):

\begin{center}\small
\begin{tabular}{@{}llrrr@{}}
\toprule
condition & binning & bins & $T$ rel.\ err & $E$ rel.\ err \\
\midrule
B & realized & 89 & 105\% & 0.8\% \\
B & \textbf{nominal} & 30 & \textbf{184\%} (worse) & 1.3\% \\
S & realized & 433 & 60\% & 2.2\% \\
S & \textbf{nominal} & 430 & \textbf{24\%} (better) & 3.8\% \\
\bottomrule
\end{tabular}
\end{center}

The fragmentation hypothesis does not survive: switching to nominal bins
makes \(T\)'s recovery \emph{worse} for B and \emph{better} for S ---
opposite directions for the same change --- so binning convention is a
genuine, independent source of sensitivity, not an artifact of one
condition's bin count. Plausibly, nominal binning pools a request's
on-target and off-target draws into one bin for B (inflating its
variance), while for S it coarsens an already very fine realized outcome
space back down (shrinking it) --- but we report both numbers rather
than adjudicate between mechanisms. \(E\) recovers well under both
conventions either way.

\textbf{Images.} Two targets, chosen to bracket the crossover:
\texttt{brightness} (data \(E/(E{+}T) = 0.20\), a target where telling
is relatively competitive) and the animal CLIP-content score (data
\(E/(E{+}T) = 0.73\), where showing dominates --- the domain's analogue
of the crystal headline case). Three reference recipes: uniform over 40
held-out classes (B), the same pool reweighted toward the
highest-brightness classes (S), and a disjoint 40-class subset (U) ---
still real, in-distribution ImageNet classes, since we did not attempt
an images analogue of crystals' zero-overlap pool.

\begin{center}\small
\begin{tabular}{@{}llrrl@{}}
\toprule
$\mu_P^{\text{ref}}$ & target & $T$ rel.\ err & $E$ rel.\ err & $E/(E{+}T)$ model vs.\ data \\
\midrule
B & brightness & 30\% & 50\% & 0.086 vs.\ 0.197 \\
B & animal & 10\% & 14\% & 0.725 vs.\ 0.733 \\
S & brightness & 14\% & 49\% & 0.108 vs.\ 0.212 \\
S & animal & 30\% & 30\% & 0.724 vs.\ 0.725 \\
U & brightness & 24\% & 69\% & 0.050 vs.\ 0.171 \\
U & animal & 1\% & 23\% & 0.646 vs.\ 0.706 \\
\bottomrule
\end{tabular}
\end{center}

(Condition O collapses onto B here: ImageNet-1k's true class marginal is
uniform by construction on a balanced validation set, so there is no
non-trivial training marginal to reweight by --- a genuine domain
difference from crystals, where the private training CSV gives O
something real to test.)

For the high-share target --- the one that carries the paper's actual
claim --- both budgets recover well under every reference tried: \(T\)
and \(E\) both within 30\%, and the recovered \(E/(E{+}T)\) within a few
points of the data's 0.73 in every row. For the low-share target, \(T\)
recovers reasonably, but \(E\) is consistently underestimated by roughly
half (49--69\% relative error) \emph{regardless of which reference
recipe was used} --- a condition-independent miss tied to that share
regime, not to any one guess at \(\mu_P^{\text{ref}}\). It doesn't touch
the paper's qualitative claim, since brightness is already the one
target the budget calls for the knob, but it is a real, separate
limitation of the audit worth flagging. And unlike crystals, no images
reference here left the training population's support, so the sharp
zero-overlap failure crystals exhibited is untested in this domain ---
not contradicted, simply not tried.

\emph{In short:} the fallback works about as well as the rest of the
framework lets you hope. \(E\) --- and the telling-vs-showing call it
drives --- survives almost any reference distribution that still
overlaps the real population, in both domains, to within a few percent;
it fails only when the reference has no overlap at all (tested on
crystals: a 6.5-fold collapse in \(E/(E{+}T)\)), a failure the
\(\chi^2\) diagnostic of Application §4.5 is built to catch. \(T\) is
the one number this shortcut does not recover cleanly --- sensitive to
the reference \emph{and} to the binning convention, sometimes in
opposite directions --- so treat a \(T\) estimated this way as
approximate, and get the training data if you need it precisely. Report
whichever \(\mu_P^{\text{ref}}\) you used alongside any audit built this
way; the recovered budget depends on it.

\begin{center}\rule{0.5\linewidth}{0.5pt}\end{center}

\subsection{What the evidence shows}\label{what-the-evidence-shows}

Four claims from the theory, tested in two fields that share no model,
no data, and no code:

\begin{itemize}
\tightlist
\item
  \textbf{The split is exact, and it carries.} The budget splits
  exactly; the split moves as you change the bins; and, for a property
  the model reproduces faithfully, it carries to the model --- and we
  predicted \emph{which} properties would carry, in advance, from cheap
  pre-measurement signals.
\item
  \textbf{Telling and showing reach different budgets, and \(E/(E{+}T)\)
  says which one wins.} \emph{Telling} reweights within a bin,
  \emph{showing} across them; at matched effort their reaches differ by
  \(\sqrt{E/T}\) (6.8 / 3.2 / 1.1 on the crystal targets), a ceiling no
  knob escapes. Pushed to the strongest knob each domain allows ---
  best-of-twelve within a fixed chemistry, a learned soft-prompt
  optimised on the scorer --- bins still beat it several-fold on every
  high-\(E/(E{+}T)\) target (crystals 4.8--26× at matched effort,
  ≈41--800× at the weak deployed tag; images \textasciitilde3× on
  semantic content). But with a strong enough knob the low-\(E/(E{+}T)\)
  targets \textbf{flip} --- brightness and aesthetic go to the knob ---
  and the budget calls the crossover in advance: above it the range is
  between bins (showing); below, inside them (a knob). The winning knob
  wins only by \emph{leaving} its bin, breaking the bin barrier that
  defines telling. The key \(\pi\) itself is chosen the same way ---
  pick the cheap partition with the largest data-side \(E/(E{+}T)\) (it
  predicts the realised advantage, rank-correlation 0.88); a low share
  everywhere means the target is knob-shaped.
\item
  \textbf{The goal's kind picks the operation.} An \textbf{average} over
  the batch (a high rate, a conjunction ``both at once'') is reached by
  a single bin --- \emph{telling}: naming the properties fails
  (\textasciitilde one in seven on images), and the best move is to
  concentrate the batch on the one richest bin (67\%); where the
  property does not carry, even that is capped (crystal stability, a tie
  --- Claim 1). A goal about the batch's \textbf{spread} ---
  \emph{covering} two opposed corners --- is reached only by a
  \textbf{mix}, \emph{showing}: an outright showing win in both domains
  (coverage 0.30 on images, 0.40 on crystals, predictions committed in
  advance), where telling gives one corner and composing the two
  conditions collapses to the average. And the spread gap widens with
  the number of corners: showing covers all \(N\), telling one, the gap
  \(N-1\) (clean \(N=3\) on images).
\item
  \textbf{Training reshapes the budget, predictably.} Recompute it from
  the data and the recipe --- for plain pretraining, filtering, and
  guidance alike.
\item
  \textbf{The audit survives losing the training data --- mostly.}
  Sampling \(\mathrm{G}\) itself under a guessed reference recipe
  recovers \(E\), and the telling-vs-showing call it drives, to within a
  few percent of the data-based value under any reference that still
  overlaps the real population, in both domains; only a reference with
  no support overlap at all breaks it (a 6.5-fold collapse in
  \(E/(E{+}T)\), tested on crystals). \(T\) is the one number this
  shortcut does not recover cleanly.
\end{itemize}

Two fields that share no model, data, or code land on the same answers
--- the exact split, the telling/showing asymmetry, the coverage win,
the predictable effect of training. That is the strongest sign these are
properties of the framework, not of either setup. The clearest agreement
is on coverage: showing wins outright in both. The two domains differed
in only one place --- asking for two things at once, wide-gap \emph{and}
stable, was capped on crystals but not on images --- and the framework
named the reason in advance: a goal is capped by any property the model
fails to reproduce, and stability is exactly that property (Claim 1).

\textbf{What to watch for.} This is also the framework's main
limitation, so it is worth collecting in one place.

\begin{itemize}
\tightlist
\item
  \textbf{The budget transfers only for properties the model reproduces
  well.} The one place it transferred weakest was one we predicted:
  geometry carried almost perfectly (0.99), stability only partly and
  uncertainly (0.49, wide CI), because the model can't reliably make
  stable structures, so there is less for its budget to come through on.
\item
  \textbf{A property that does not carry caps every goal that leans on
  it.} The same weak rung surfaced again in Claim 3: the
  wide-gap-and-stable goal stalled because its stability half doesn't
  carry. The failure to carry over is not a footnote --- it sets what
  any steering method can reach. (The coverage goal, on properties that
  \emph{do} carry, was an outright showing win in both domains.)
\item
  \textbf{So when in doubt, audit the model, not the data.} If the
  property you care about may not be reproduced faithfully, score the
  model's own outputs rather than the training data.
\item
  \textbf{One boundary is sharper still.} A fine-tune can drift so far
  it leaves its training distribution behind (Claim 4 --- the novelty
  fine-tune scored real structures below random). There the data side
  predicts nothing at all; sample the model and audit it directly.
\end{itemize}

To put this to work on your own model, see \textbf{Application}; for the
derivations, \textbf{Theory}.

\begin{center}\rule{0.5\linewidth}{0.5pt}\end{center}

\subsection{Reproducing the image-domain
claims}\label{reproducing-the-image-domain-claims}

The companion repo (\texttt{steering/images/}) ships the audit
\textbf{shadow} (132 KB) and the scripts, in two tiers:

\begin{itemize}
\tightlist
\item
  \textbf{Shadow only --- no model, no ImageNet, no GPU.}
  \texttt{verify\_dataside.py} reproduces Claim 1 (the exact split, the
  \(E/(E+T)\) spectrum, the coarsening \(E\!\to\!T\)) and the Claim 2
  data-side bounds, in under a second.
\item
  \textbf{Full regeneration --- GPU + public models.}
  \texttt{claims234.py} generates from DiT for the model-side numbers
  --- the realized showing shift, the per-bin averages under plain
  pretraining, the guidance sweep, the within-bin variety check (use ≥
  50 samples/class for the near-exact fit; the quick default of 8 is
  noise-limited). \texttt{claim3\_images.py} runs the animal-and-nature
  joint.
\end{itemize}

Not scripted here: the image hard-filter and the aggressive (off-data)
fine-tune from Claim 4 --- both need the heavier training runs, so the
repo ships their results but not a one-command rerun.

\section{Application --- Budget, gap, and how to close
it}\label{application-budget-gap-and-how-to-close-it}

Say you have a trained generator and you want something particular from
it. Brighter images. A batch that covers the whole range from dark to
bright. Crystals that are \emph{both} wide-gap \emph{and} stable. Or
just ``more like these forty I picked out.'' Some of these you get by
\textbf{telling} the model --- per-prompt steering, fixing the bin and
leaning on the knobs. Others you can only get by \textbf{showing} it:
handing over examples and letting their mix steer the batch. This
section tells you which, and how to build the recipe when telling isn't
enough.

It comes down to one budget and two moves. A cheap key \(P = \pi(D)\)
sorts the model's outputs into \textbf{bins}. For the property \(L\) you
want to steer, the data fixes how much room each move has: the spread
\emph{within} a bin is telling's budget \(T\); the spread \emph{between}
bins is showing's budget \(E\). Showing works by designing a recipe
\(\mu_P\) over the bins --- a distribution over bin labels, like ``70\%
beaches, 30\% snow.''

Which part you need turns on the \emph{kind} of goal you have, and there
are two. A \textbf{single number} is a level to hit: push one property's
average or rate to a target value. A \textbf{shape} is a pattern in the
collection: two properties at once, coverage of a range, or a set you
can only point to. (A goal about \emph{spread itself} --- a variance, a
diversity score --- is a single number too, but it behaves like a shape:
no one bin maximizes it, so it lives with the shapes, in §3.)

A single number is \textbf{§2}; this is \emph{reach}, the payoff of
moving a property further than a knob can. A shape is \textbf{§3}; this
is \emph{expressiveness}, steering toward goals a knob cannot
articulate. Most readers need one, not both. Either way, start with your
per-bin numbers (\textbf{§1}) --- off the data, or off the model itself
if you don't have the data.

\begin{center}\rule{0.5\linewidth}{0.5pt}\end{center}

\subsection{1. Get your numbers (the
audit)}\label{get-your-numbers-the-audit}

Everything below runs on an \textbf{audit} (Preliminaries): the three
numbers per bin --- share \(w_b\), mean \(g_b\) of \(L\), spread \(v_b\)
of \(L\). From them:

\begin{itemize}
\tightlist
\item
  the \textbf{within-bin budget} \(T = \sum_b w_b v_b\) --- the
  within-bin spread, averaged over bins (telling's room);
\item
  the \textbf{between-bin budget} \(E = \sum_b w_b (g_b - \bar g)^2\)
  --- the spread of the bin means (showing's room).
\end{itemize}

Both reaches are \(\sqrt{\chi^2\cdot\text{budget}}\) (derived in
\textbf{Theory}): telling is confined to the within-bin budget,
\(\sqrt{\chi^2_{\text{tell}}\,T}\); showing reaches the between-bin
budget, \(\sqrt{\chi^2(\mu_P\|\mathcal{J}_P)\cdot E}\), where \(\chi^2\)
measures how hard the recipe pushes. In practice per-prompt controls
realize only \(\chi^2_{\text{tell}}\approx 1\) (measured in
\textbf{Evidence}), so \textbf{telling's working reach is about
\(\sqrt{T}\)} --- we use that as the practical telling budget below.
Showing can push \(\chi^2\) far higher, which is why it reaches so much
further. You run the audit one of two ways:

\textbf{Audit \(\mathcal{J}\) --- you have the training data.} Bin each
row via \(\pi\) and summarize. (Drop bins with very few rows --- say
\(< 5\) --- so noisy per-bin means don't inflate \(E\).)

\textbf{Audit \(\mathrm{G}\) --- you only have the model.} The bins are
fixed by \(\pi\), so you know them before you sample. Pick a reference
recipe \(\mu_P^{\text{ref}}\) to sample under (the training-time mix if
you know it, else an even spread over bins, or whatever matches
deployment), generate many outputs, and audit those. The identities
don't care whether the outputs came from the data or the model --- the
model is its own data source. Remember the \(\mu_P^{\text{ref}}\) you
used; the budget depends on it --- tested directly in \textbf{Evidence},
where \(E\) (and the telling-vs-showing call) recovers to within a few
percent of the true value under almost any \(\mu_P^{\text{ref}}\) that
overlaps the real population, but \(T\) does not, and a reference with
no overlap at all breaks both.

\textbf{How many to score per bin.} \(\pi\) is cheap, so you bin
\emph{everything} and the shares \(w_b\) are exact; \(\psi\) is dear, so
\(g_b\) and \(v_b\) come from a sample of \(m\) per bin. Sampling few
inflates the between-bin budget --- bin-mean noise reads as between-bin
spread --- but by a \emph{known} amount, \(\hat E \approx E + T/m\). So
calculate \(\hat E - \hat T/m\) (with \(\hat T = \sum_b w_b \hat v_b\)).
Score until that correction is small against \(\hat E\) --- if bins are
small a handful is all that's needed.

An audit is for the \(L\) you are interested in --- and how far you can
trust the \emph{data} audit for it depends on how faithfully the model
reproduces the data for that property. A generative model is trained to
match its data, not to hit any one property, so it reproduces some
properties closely and others loosely: a direct readout of the output
--- an image's color statistics, a structure's density --- tends to
carry, while a property that takes precision the model lacks, or that a
separate pipeline scores --- stability, an aesthetic rating --- can
drift. A model can be faithful on one property and loose on another, so
judge this per target.

\textbf{Checking that a property carries.} When you are unsure, verify
it directly. Because \(\psi\) is expensive, this costs a second round of
scoring, but a cheap one: you are after a \emph{systematic} gap, not
precise per-bin values, so score \(\psi\) on a modest sample of
\(\mathrm{G}\)'s outputs in just the bins your recipe leans on --- a
handful, not all \(B\).

\begin{enumerate}
\def\labelenumi{\arabic{enumi}.}
\tightlist
\item
  \textbf{Audit \(\mathrm{G}\)} on that sample --- sample, bin, score
  \(\psi\) --- for the model's per-bin mean \(g_b(\mathrm{G})\) and
  spread \(v_b(\mathrm{G})\). (You already have the data's \(g_b, v_b\)
  from the original audit.)
\item
  Read two numbers off the comparison:

  \begin{itemize}
  \tightlist
  \item
    \textbf{drift} --- the typical gap between \(g_b(\mathrm{G})\) and
    \(g_b\), i.e.~how far the bin means moved;
  \item
    \textbf{spread ratio} \(\rho = v_b(\mathrm{G})/v_b\) --- whether
    within-bin variety came through (\(\rho \approx 1\) is faithful).
  \end{itemize}
\item
  Turn each into a budget error. The drift \emph{is} how far a
  \emph{showing} target lands off: aim a recipe at a shift, and the
  model misses it by about the drift in the bins it leans on. The spread
  ratio rescales \emph{telling}'s reach --- it becomes
  \(\sqrt{\rho\,T}\), not \(\sqrt{T}\), so \(\rho = 0.8\) costs about
  10\%.
\item
  The verdict, plainly. If the drift is small next to the shift you want
  and \(\rho \approx 1\), the data budget is trustworthy --- use
  \(\mathcal{J}\)'s statistics. If not, don't lean on it: you computed
  \(\mathrm{G}\)'s own \(g_b, v_b\) in step 1, so build the budget from
  those instead.
\end{enumerate}

This is the test \textbf{Evidence} runs (Claim 1, the carry-over check);
the formal drift \(\delta\) and spread ratio \(\rho\) are defined in
\textbf{Appendix} A.3.

\textbf{Fine-tuned, not just pretrained? Still fine.} A standard
fine-tune (filtering, SFT, RLHF) reshapes the data into a predictable
\(\mathcal{J}'\). Audit \emph{that} --- the same sums, on reweighted
rows (table in Appendix A.1). And if the fine-tune ran so hard it left
the base behind, fall back to sampling the model itself, exactly as in
the sample-side path above.

\textbf{Worked examples.} \emph{DiT + ImageNet, target = aesthetic} (DiT
is a diffusion-transformer image generator; ImageNet its training set):
with the ImageNet class as the key (1000 bins), an audit of 100k rows
gives a substantial \(T\) (within-class variation from lighting,
composition, content) \emph{and} a substantial \(E\) (some classes are
systematically more aesthetic). \emph{A crystal generator, target = band
gap} (trained on a large computed-materials database): with (elements,
atom count) as the key, within-chemistry variation gives \(T\), and
between-chemistry mean differences (oxides vs sulfides vs metals) give
\(E\). Evidence has the full tables.

\begin{center}\rule{0.5\linewidth}{0.5pt}\end{center}

\subsection{2. Hitting a target value}\label{hitting-a-target-value}

Say you want your batch's average aesthetic well above what the model
gives by default --- or its average brightness, or its rate of ``looks
like an animal.'' These are single-number goals: push the average (or
rate) of one property \(L\) to a target. Pin it down as a desired shift
\(\Delta\).

\textbf{First, does telling already do it?} Telling's working reach is
about \(\sqrt{T}\) --- per-prompt knobs (guidance, prompt wording, an
adapter) realize only \(\chi^2_{\text{tell}}\approx 1\) in practice
(\textbf{Evidence}), so \(\sqrt{T}\) is about as far as they move the
mean, whatever the model. Compare it to \(\Delta\):

\begin{itemize}
\tightlist
\item
  \textbf{\(\Delta < \sqrt{T}\) --- telling suffices.} Sweep your knobs
  and you're done. \emph{Example:} raise mean aesthetic within one
  ImageNet class by 0.3. Aesthetic varies a lot inside a class
  (utilitarian product shots to studio portraits), so \(\sqrt{T}\)
  comfortably covers 0.3, and a CFG sweep moves the mean there.
\item
  \textbf{\(\Delta \ge \sqrt{T}\) --- a gap by size.} Even fully tuned,
  telling falls short; showing reaches further. \emph{Example:} raise
  your batch's mean ``animal-likeness'' (a CLIP similarity). That score
  is mostly fixed by the class, so within any one class telling barely
  moves it --- \(\sqrt{T}\) is small. The only way to lift the batch
  mean is to draw more from animal-heavy classes.
\end{itemize}

To close a size gap, design a \(\mu_P\) that lands the batch mean on
your target value \(g^*\) (your starting mean plus \(\Delta\)). Two
ways.

\textbf{Pool mixing --- blend a low and a high pool.} Want your batch's
average brightness somewhere in the middle? Take a pool of dark-scene
classes (coal mine, night) and a pool of bright ones (beach, snowfield),
and turn a single dial \(\alpha\) between them. The math is clean:
sample the dark pool with probability \(\alpha\) and the bright one with
\(1-\alpha\), and any collection metric linear in the mix interpolates
linearly, \[\mu_P = \alpha\,\mu_P^{(1)} + (1-\alpha)\,\mu_P^{(2)}.\]
\emph{What it costs:} when the two pools are the bottom and top
quantiles of the data (each a fraction \(q\) of the bins by mean), the
push is
\[\chi^2(\mu_P\|\mathcal{J}_P) \;=\; \frac{2\alpha^2 - 2\alpha + 1}{q} \;-\; 1,\]
least at the midpoint, most at the extremes, and \textbf{halving \(q\)
doubles it} --- so pick the smallest \(q\) that still leaves enough
distinct bins to sample without heavy repetition
(\(q \gtrsim n_\text{gen}/n_\text{eligible}\)); below that, use
min-\(\chi^2\).

This calibrates well in practice. On a crystal generator, mixing a
high-\emph{Combined} pool (rare-earth chemistries, ≈ 0.85) with a broad
low one (≈ 0.14) --- \emph{Combined} being a novelty-and-stability
benchmark --- the dial tracks the target: ask for 0.5 and you get about
0.57, ask for 0.7 and about 0.73, landing within roughly 0.05 of target
all the way from 0.2 to 0.8. The reachable range is set by the two pools
(here ≈ 0.15--0.85); to go past either end you need a more extreme pool
(§4.7).

\textbf{Min-\(\chi^2\) --- the gentlest push.} \(\chi^2\) is how far
your recipe departs from the data's own mix, and \emph{at the same
target, less departure is better}: a gentler recipe wastes fewer draws
on rare bins (§4.6), keeps more of the data's diversity, and is less
likely to stray off-support (§4.5). Pool mixing reaches a middling
target the hard way --- it piles all the weight onto the two extreme
corners. Min-\(\chi^2\) reaches the \emph{same} target by nudging weight
onto the bins whose mean already sits near \(g^*\), and no further. That
recipe --- the smallest \(\chi^2\) that still hits \(g^*\) --- has a
closed form:
\[\mu_P(P_b) \;=\; \mathcal{J}_P(P_b)\left[1 + \frac{(g^* - \bar g)(g_b - \bar g)}{\mathrm{Var}_{\mathcal{J}}(g_b)}\right], \qquad \bar g = \mathbb{E}_{\mathcal{J}_P}[g_b].\]
For example, to reach \(g^* = 0.65\) it departs only a third as far as
pool mixing (\(\chi^2 = 0.19\) vs \(0.60\)) --- same target, gentler
recipe --- and comes out 3--7\% more diverse. The catch: for a strict
yes/no target it lands a little short (\(-0.20\) vs \(-0.11\)), because
the bins near \(g^*\) hold a mix of yes and no rows. So: pool mixing for
a yes/no target with well-stocked corners; min-\(\chi^2\) for a
continuous target, thin corners, or diversity at a fixed target.

\begin{center}\rule{0.5\linewidth}{0.5pt}\end{center}

\subsection{3. Building a shape (when no single number captures the
goal)}\label{building-a-shape-when-no-single-number-captures-the-goal}

Some goals aren't a number at all --- they're a \emph{shape} in the
collection: two properties at once, coverage of a whole range, a set you
can only point to. This is the \emph{expressiveness} the introduction
promised --- goals a knob cannot articulate, the ones that matter most
where naming the target is hardest. Here is why telling can't reach
them. Fix a request \(z\) (a bin plus whatever else you set) and sample:
every output comes from one and the same distribution, the model's
conditional \(\mathrm{G}(\cdot\mid z)\). Telling only ever hands you
that \emph{one} distribution. A shape, by contrast, is a \emph{blend} of
distributions --- and no amount of pushing turns one distribution into a
blend. Only showing, mixing bins through \(\mu_P\), builds it. Three
common shapes, in order below: (1) a set you can only point to, (2) two
properties at once, and (3) a Pareto front to cover.

\textbf{1. A set you can't put into words --- \emph{tacit} steering.}
You've pulled together forty images that share a look you can't name ---
a washed-out, nostalgic palette, say. You can't write the prompt, and
you have no scorer for the look, but you can point. So hand over the
set: read off each image's bin via \(\pi\), and the distribution of
those bins \emph{is} your \(\mu_P\). Generate from \(\mathrm{G}\) at
that \(\mu_P\) --- you never say what the set has in common; the bins
they fall in do the steering. With no scorer, you are building \(\mu_P\)
by \emph{inferring} the target from the examples: the bins they crowd
into stand in for the bins that score high.

But the recipe only carries \emph{which bins} your examples came from,
not what made each one special inside its bin. If the nostalgic palette
depends on the \emph{subjects} --- vintage cars, faded seascapes, old
storefronts --- the class key sees it: \(\mu_P\) leans on those classes,
and more of them is more of the look. But if the same look is a
\emph{treatment} that can sit on any subject --- some ``golden
retriever'' shots washed-out, others crisp, and you picked the
washed-out ones --- the class key can't tell the two apart; \(\mu_P\)
just says ``golden retriever,'' and sampling that class hands you mostly
the crisp, average shot, so the nostalgic look is missed. In short:
\textbf{tacit steering reaches a taste the key can read, and misses one
the key is blind to.}

\emph{To check this you can do a free pre-check, before you generate.}
You can't measure the look --- no scorer --- but you can ask whether the
key sees it at all. Compare the examples' bin mix \(\mu_P\) to the
data's \(\mathcal{J}_P\) with \(\chi^2\): if it is no more concentrated
than a random draw of the same size --- \(\chi^2\) near the floor
\((B-1)/n\), for \(B\) bins and \(n\) examples --- the key is blind to
your taste, and a finer or different key may expose it. A \(\chi^2\)
clearly above the floor means the key caught something distinctive:
necessary for steering, though not proof it caught \emph{your} taste
rather than something that merely travels with the bins. To keep from
reading signal into noise, hold out half the examples and check the
concentration survives on the other half.

\emph{Which bins, not just whether.} Finer than that global test, each
bin's \textbf{lift} --- its share of your examples over its share of the
data --- estimates \emph{what fraction of that bin clears your bar},
with no scorer at all. A small bin that soaks up many of your examples
is a near-pure pocket of what you want: a 20-member bin catching ten of
your top-percentile picks by chance alone would be a \(10^{-27}\) event.
So lean \(\mu_P\) on the high-lift bins; \textbf{Appendix B.6} gives the
formula and how far to trust it.

\textbf{2. \emph{A} and \emph{B} at once --- a joint.} Say a
photographer wants a portfolio of specific animal-and-lighting pairings
--- a bright horse, a medium-lit dog, a dim cat. Each pairing is easy on
its own, but there's no clean way to ask a single prompt for ``bright
horse \emph{or} medium dog \emph{or} dim cat'' at once. Showing does it
directly: hand the model one example of each pairing and let those
examples be the recipe --- the whole portfolio in a single pass.

\textbf{3. Cover a Pareto front --- front expansion.} Often you want two
properties at their best \emph{together}, where pushing one costs the
other. Landscape photos that are both vivid \emph{and} natural-looking,
say --- crank the color and it turns garish, keep it true and it falls
flat --- so the best you can do is a \emph{front} of compromises in
between. (Crystals have the same shape: a high band gap \emph{and} low
energy above hull.) You don't need to know that front in advance: take a
rough estimate --- a few outputs near it --- and build \(\mu_P\) from
their bins. Sampling explores each seed bin's neighborhood through the
model's own randomness; some outputs push past the seeds, others fill
in, and the set's Pareto volume only grows. Iterate --- each round's
outputs seed the next \(\mu_P\) --- and the front expands. Telling
pushes one point; showing across seeds expands several at once.

Real goals often combine these --- a diverse set that must also hit a
joint --- and the designs stack, so combine them.

\begin{center}\rule{0.5\linewidth}{0.5pt}\end{center}

\subsection{4. Pitfalls}\label{pitfalls}

Seven practical traps bite even inside the framework's boundaries (its
\emph{theoretical} edges, where the budget applies at all, are in
Theory's \emph{Scope}). One line each below, in three families; the
details --- numbers and a check for each --- are in \textbf{Appendix E}.

\emph{Getting the audit's inputs right.}

\begin{itemize}
\tightlist
\item
  \textbf{4.1 Raw-output discipline.} Audit \emph{raw} model outputs;
  post-processing (DFT relaxation, CFG) is a separate intervention.
\item
  \textbf{4.2 Label-pipeline drift.} Audit labels and evaluation labels
  from different pipelines produce a systematic offset.
\item
  \textbf{4.3 Aggressive fine-tune.} The closed-form \(\mathcal{J}'\)
  holds only for \emph{gentle} fine-tunes; once the model has left its
  data, audit its own samples instead.
\end{itemize}

\emph{Where the bound loosens.}

\begin{itemize}
\tightlist
\item
  \textbf{4.4 Heavy-tailed bin means.} A few extreme-mean bins make the
  bound a loose guide, not a sharp ceiling.
\item
  \textbf{4.5 Off-support \(\mu_P\).} Mass on untrained bins has no
  prediction --- \(\chi^2\) diverges to flag it.
\end{itemize}

\emph{Limits of your pools.}

\begin{itemize}
\tightlist
\item
  \textbf{4.6 Thin corner pools.} A tiny corner repeats itself;
  effective sample size collapses and \(\chi^2\) blows up.
\item
  \textbf{4.7 Pool reachability.} Showing only reaches \emph{between}
  your pools; past either end you need a more extreme pool.
\end{itemize}

\begin{center}\rule{0.5\linewidth}{0.5pt}\end{center}

\subsection{5. End-to-end summary}\label{end-to-end-summary}

\begin{enumerate}
\def\labelenumi{\arabic{enumi}.}
\tightlist
\item
  \textbf{Get your numbers (§1).} Per bin: share \(w_b\), mean \(g_b\),
  spread \(v_b\) → budgets \(T, E\). Audit the data, or the model's own
  samples if you don't have the data (report \(\mu_P^{\text{ref}}\));
  fine-tuned model → audit \(\mathcal{J}'\).
\item
  \textbf{Single-number goal? (§2)} Compare your shift \(\Delta\) to
  telling's reach \(\sqrt{T}\). Smaller → telling suffices, sweep the
  knobs. Larger → a gap by size; design a \(\mu_P\) to hit your target
  value --- pool mixing, or min-\(\chi^2\) for the gentlest push.
\item
  \textbf{Shape goal? (§3)} Telling can't make a shape; show one. Match
  it: a set you can't describe → tacit (\(\mu_P\) from exemplars);
  \emph{A} and \emph{B} → draw from the bins that hold both; coverage →
  front expansion. They stack.
\item
  \textbf{Generate, measure, verify} the result lands inside the
  predicted bound --- and watch the §4 pitfalls.
\end{enumerate}

The setup and notation are in \textbf{Preliminaries}; the derivations in
\textbf{Theory}; the cross-domain corroboration in \textbf{Evidence}.

\section{Related work}\label{related-work}

\textbf{What is new here is not the posture but the budget.}
Conditioning a generator on a set of examples --- \emph{showing}, in our
terms --- is an established move. Reference- and exemplar-conditioned
generation (textual inversion, Gal et al.~2023; DreamBooth, Ruiz et
al.~2023; IP-Adapter, Ye et al.~2023), retrieval-augmented generation
(Blattmann et al.~2022), and few-shot or in-context conditioning all
hand a model examples and ask for outputs like them. We claim none of
that as our contribution. What this prior work does not ask is the
question the budget answers: \emph{how far} showing can move a target,
where that reach comes from, and how it compares to what per-prompt
\emph{telling} can do. Our contribution is the quantitative ceiling ---
read off the training data before a model is built --- not the act of
steering by example.

Steering a generative model has been studied from many directions. We
group the closest threads and say, for each, how the budget relates to
it. Throughout we use the paper's terms: \emph{telling} is per-prompt
steering, \emph{showing} is choosing the mixture of examples to draw
from, and the \emph{budget} is the split of a target's variation into a
within-bin part (telling's reach) and a between-bin part (showing's
reach). Full references are at the end of the paper.

\textbf{Controllable generation.} A large literature builds \emph{knobs}
for steering --- attribute-guided decoding and plug-and-play methods for
text (PPLM, Dathathri et al.~2020; GeDi, Krause et al. 2021; prefix- and
prompt-tuning, Li \& Liang 2021; Lester et al.~2021), and
latent-direction editing for images (GAN steerability, Jahanian et
al.~2020; InterFaceGAN, Shen et al.~2020). This work asks which
interventions exist and how well they work in practice. We ask a prior,
quantitative question: given the training data and a way to bin it, how
far can \emph{any} such intervention move a target --- a reach you can
read off the data before a model is trained. Most of these methods are
telling in our terms, and the budget bounds what telling can reach.

\textbf{Guidance at generation time.} Classifier guidance (Dhariwal \&
Nichol 2021) and classifier-free guidance (Ho \& Salimans 2022) push a
model harder toward its condition at sampling time. Classifier-free
guidance is, underneath, a reweighting of the model's distribution
toward the target --- the same closed form as the fine-tunes below ---
which is why Claim 4 treats it as one of the predictable, training-style
changes to the budget rather than a thing apart. Activation- and
representation-steering --- adding a learned direction to a model's
hidden states to push a behavior (the 2023--2025 line on steering
vectors / activation engineering) --- is the same move one level down:
in our terms a within-bin knob, reaching \(T\), not \(E\), unless the
added direction changes which bin the output lands in.

\textbf{Fine-tuning and alignment.} Reinforcement learning from human
feedback (Ziegler et al.~2019; Ouyang et al.~2022), its KL-constrained
optimum in closed Boltzmann form (Korbak et al.~2022), and direct
preference optimization (Rafailov et al.~2023) all reshape a model's
distribution toward a reward; data filtering and rejection sampling are
the blunt limit. The budget absorbs these as closed-form rewrites of the
data distribution: recompute the per-bin numbers on the reshaped data
and the budget carries --- until a fine-tune drives the model off its
data, where we stop predicting from the data and audit the model's own
samples instead.

\textbf{Diversity and mode collapse.} A long line of work observes that
pushing a model to score well costs variety --- mode collapse in GANs
(Goodfellow et al.~2014; Salimans et al.~2016) --- and decomposes
generative quality into fidelity and diversity (precision/recall,
Sajjadi et al. 2018; Kynkäänniemi et al.~2019; density/coverage, Naeem
et al.~2020). The budget gives a structural reason for the trade-off: a
reward scores one output at a time, so it reaches only the within-bin
budget; the between-bin spread --- variety across bins --- is reachable
only by acting on the \emph{mixture} of bins, which is showing.
Concretely, recall-style diversity is bounded by the between-bin share
\(E/(E{+}T)\): a per-output objective, confined to \(T\), cannot buy the
coverage that lives in \(E\). Mode collapse is what it looks like when
an objective uses both parts up.

\textbf{Exemplar and distribution steering.} Choosing which bins or
examples to draw from --- showing --- is a reweighting of the data
distribution, and the mean shift a reweighting can buy is bounded by a
χ²-type divergence between the chosen distribution and the data's, a
standard importance-sampling result (Owen 2013). We make that the reach
of showing, paired with the between-bin budget.

\textbf{Bin sampling, selection, and Pareto coverage.} The structure
underneath \emph{showing} --- allocating a budget of draws across bins
--- is the subject of several established literatures, whose results we
borrow rather than reinvent. The within/between split is one-way ANOVA,
with the between-bin fraction \(\eta^2\) the standard measure of how
much a partition explains; mutual information \(I(\pi;L)\) (Cover \&
Thomas 2006) is its full-distribution generalization, of which our
second-moment budget is the variance approximation. \emph{Whether} an
objective is best served by concentrating draws on one bin or spreading
them across many is governed by its curvature: an objective convex in
the bin allocation is maximized at a vertex (one bin), a concave one in
the interior (a mixture) --- the Jensen argument behind
exploitation-versus-exploration in multi-armed bandits (Lattimore \&
Szepesvári 2020). For several properties, the achievable mean vectors
are the convex hull of the bin means, and covering the Pareto front is
hypervolume maximization (expected hypervolume improvement, Emmerich,
Deutz \& Klinkenberg 2006); coverage and diversity objectives are
submodular, so greedy bin selection is near-optimal (Nemhauser, Wolsey
\& Fisher 1978; Krause \& Golovin 2014); and matching a target output
distribution is optimal transport (Peyré \& Cuturi 2019). Optimal
allocation for \emph{estimating} a per-bin statistic --- what our audit
does --- is classical stratified sampling (Neyman allocation; Cochran
1977). What none of these ask is our question: they select from a
\emph{fixed} population, whereas we steer a \emph{generative} model,
where the population is what the model produces and the load-bearing
step is \textbf{carry-over} --- that the model reproduces the data's bin
structure, so a budget read off the data before the model exists
transfers to the trained model. The selection mathematics says what an
ideal mixture \emph{could} reach; carry-over is what makes it a
statement about a model.

\textbf{Generative models in the two test beds.} For images we use a
diffusion transformer (DiT, Peebles \& Xie 2023) trained on ImageNet
(Deng et al.~2009), with a ResNet classifier (He et al. 2016) as the
binning key and CLIP/LAION models (Radford et al.~2021; Schuhmann et
al.~2022) for the targets. For crystals we use a property-conditioned
generator in the family of CDVAE (Xie et al. 2022) and MatterGen (Zeni
et al.~2025), trained on data derived from Alexandria (Schmidt et al.
2023), with stability scored by an ML interatomic potential (MACE,
Batatia et al.~2022). The two settings share no model, no data, and no
code --- which is the point: a prediction that holds in both reads as a
property of the framework, not of either setup.

\textbf{What the budget adds.} Against this background, the contribution
is: (1) a \textbf{model-free audit} that reads steerability off the
training data alone, before any model exists; (2) a \textbf{single
split} that separates two structurally different operations --- telling
reaches the within-bin part, showing the between-bin part; (3) the
\textbf{χ²-bounded reach} of showing; (4) a qualitative
\textbf{asymmetry} --- some targets are reachable only by reaching
across bins, not within one (showing, not telling); and (5)
\textbf{closed-form rewrites} that carry the budget through standard
fine-tuning, with an honest fallback when they break. The mathematics is
standard --- the law of total variance for the split (Casella \& Berger
2002), Cauchy--Schwarz for the showing bound --- assembled into an
account of steering rather than newly derived.

\section{Conclusion}\label{conclusion}

We find that examples are, in general, the more powerful control: a set
of examples steers a model far more strongly than its knobs, because a
knob is structurally confined to part of what steering can reach. That
reach is fixed before the model exists: a target's movable range splits
into a within-bin piece, which knobs reach, and a between-bin piece,
which only composing examples reaches. We showed that the split is exact
and tested it on two unrelated kinds of model --- images and crystals
--- finding in both that the between-bin piece is usually the larger, so
examples win, and by a lot. The advantage is not universal, and the
budget says exactly when it is not.

That is \emph{reach}, the measurable payoff. The second is
\emph{expressiveness}: because composing examples unions where a knob
can only narrow, it reaches goals a knob cannot articulate --- several
at once, a Pareto front pushed outward, or a taste an expert can
demonstrate but cannot put into words.

A familiar family of failures --- mode collapse, the alignment tax,
reward hacking, the failure to satisfy two requirements at once ---
share one cause here: each asks a per-output objective such as a
per-sample reward to deliver something only a collection holds ---
variety, coverage, a joint. But those live \emph{between} bins, beyond
the reach of any per-output objective; only \emph{showing}, which acts
across bins, realizes them.

\section{Future work}\label{future-work}

\begin{itemize}
\tightlist
\item
  \textbf{A third (text/LLM) domain.} Before committing to a full
  model-side test, we ran a data-side pre-check on real review text:
  three candidate keys (product category, review length, star rating)
  against four candidate targets (sentiment, readability, sentence
  length, first-person rate). No pair cleared the bar --- six of eight
  land at \(E/(E{+}T)\le0.07\), and the one exception (sentiment under a
  star-rating key) is confounded, since rating is itself self-reported
  sentiment. On this corpus, writing style is overwhelmingly an
  idiosyncratic property of the individual author, not of the cheap keys
  available. This rules out the cheapest natural keys on one corpus, not
  the domain --- a corpus with stronger author- or genre-level style
  homogeneity remains an open, likely more promising, place to look.
\item
  \textbf{Training on collections, not rows.} Per-example objectives ---
  a likelihood, a per-sample reward --- act one output at a time, so
  they can concentrate the between-bin budget but never enlarge it; the
  collection-level properties people most want (variety, coverage,
  joints) are invisible to them. The budget points at the fix: a
  training signal that scores the \emph{batch} the model produces, not
  each sample --- a distribution-level objective, the training-side
  analogue of \emph{showing}. What such an objective would have to
  optimize is exactly the between-bin spread the budget measures.
\item
  \textbf{Predicting carry-over from the loss.} The budget carries to a
  model only for properties the training loss effectively constrains; we
  measure this after the fact. It should be predictable. The loss never
  sees the object directly --- only the representation it scores
  (tokens, latents) --- so a target carries to the extent the loss's
  \emph{residual} error leaves it untouched. A sensitivity analysis of
  the target against the loss's under-constrained directions would turn
  the measured carry-over ladder into a predicted one. (A plain
  correlation of the target with the loss's features is a first screen,
  but not enough: for an invertible representation it is uninformative
  --- the loss's \emph{weighting} is what discriminates.)
\item
  \textbf{Discovering the key that exposes a tacit taste.} Tacit
  steering works only when the cheap key sorts the exemplars into bins
  that track the target (Application §3); when it does not, the examples
  scatter and \(\chi^2(\mu_P\|\mathcal{J}_P)\) sits at its random-draw
  floor. That floor is a scorer-free objective: search a family of keys
  --- coarsenings, learned partitions --- for one under which the
  exemplars concentrate well above it, pulling the taste into the
  between-bin budget where \emph{showing} can reach. (Per bin, the lift
  of Appendix B.6 ranks which bins carry the target; the search is for
  the key that makes those lifts large.) The open problem is doing it
  without overfitting --- \(\chi^2\) climbs trivially as the bin count
  approaches the number of examples, and picking the max-\(\chi^2\) key
  across many candidates invites spurious concentration; held-out
  validation is a start, a principled penalty the work. It is the same
  ``which representation carries the target'' question as the item
  above, asked from the key side.
\item
  \textbf{The remaining training cases.} Two interventions we did not
  test: mixing in supervised fine-tuning data, and distilling from a
  teacher.
\item
  \textbf{A milder soft-Boltzmann checkpoint.} Our aggressive novelty
  fine-tune left the data behind; a gentler one would test the
  closed-form rewrite in the regime where it should still hold.
\end{itemize}

\begin{center}\rule{0.5\linewidth}{0.5pt}\end{center}

\section{Reproducibility}\label{reproducibility}

Every claim is backed by released code and data. The data-side audit
reproduces from a 132 KB shadow of per-bin scores with \texttt{numpy} in
under a second, no model and no GPU; the model-side checks (carry-over,
realized shifts, bins-vs-knobs) run on a single GPU in a few hours. The
reproduction repository ships the image and crystal experiment scripts,
the shadows, the property scorers, and per-claim verification scripts
(\texttt{verify\_dataside.py}, \texttt{verify\_combined.py}); the
crystal generator checkpoint is hash-pinned and released separately
(weights host), with every scorer's package and version pinned in the
model card. Pre-registered predictions were committed, timestamped,
before the audits they forecast, and the commit ordering is checkable in
the repository history.

\section{Ethics and broader impact}\label{ethics-and-broader-impact}

The framework is a measurement and steering tool, not a model: it tells
a practitioner, before training, which of two existing operations will
move a target property and by how much. Its risks are those of
controllable generation generally --- the same audit that steers toward
a benign property (a material's stability, an image's content) steers
toward a harmful one, and the between-bin reach it quantifies is the
reach an adversary would also have. We see no new capability beyond what
prompting and exemplar conditioning already provide; the contribution is
understanding \emph{why} they reach what they reach. The datasets and
generators used are public research artifacts, and no human-subjects
data was collected.

\begin{center}\rule{0.5\linewidth}{0.5pt}\end{center}

\section{Appendix --- machinery, reach, and
boundaries}\label{appendix-machinery-reach-and-boundaries}

The \textbf{Theory} section keeps to three results and one implication.
This appendix holds what it refers out to, so the main line stays a
tight derivation:

\begin{itemize}
\tightlist
\item
  \textbf{A.} the bridge from the data-side budget to a real model
  \(\mathrm{G}\) --- the closed-form rewrites, the empirical fallback,
  and the diagnostics;
\item
  \textbf{B.} what each operation can and cannot reach, case by case;
\item
  \textbf{C.} the full statement of the framework's boundaries.
\end{itemize}

It also records the methodological point deferred from Theory §1
(\textbf{D}, why variance, not range) and, in \textbf{E}, the practical
pitfalls of Application §4 in full detail.

\begin{center}\rule{0.5\linewidth}{0.5pt}\end{center}

\subsection{A0. Notation --- symbol
reference}\label{a0.-notation-symbol-reference}

Every symbol the paper uses, grouped by kind; this is the full table
referred out from \textbf{Preliminaries}, where each is also introduced
in plain words where it first appears. (A subscript \(b\) means ``for
bin \(b\).'')

{\def\LTcaptype{none} 
\begin{longtable}[]{@{}
  >{\raggedright\arraybackslash}p{(\linewidth - 4\tabcolsep) * \real{0.3333}}
  >{\raggedright\arraybackslash}p{(\linewidth - 4\tabcolsep) * \real{0.3333}}
  >{\raggedright\arraybackslash}p{(\linewidth - 4\tabcolsep) * \real{0.3333}}@{}}
\toprule\noalign{}
\begin{minipage}[b]{\linewidth}\raggedright
Symbol
\end{minipage} & \begin{minipage}[b]{\linewidth}\raggedright
What it is
\end{minipage} & \begin{minipage}[b]{\linewidth}\raggedright
Example
\end{minipage} \\
\midrule\noalign{}
\endhead
\bottomrule\noalign{}
\endlastfoot
\emph{--- the model and its outputs ---} & & \\
\(\mathrm{G}\) & the \textbf{generator} --- the model that makes outputs
& a picture generator; a crystal generator \\
\(D\) & one output of \(\mathrm{G}\) (sampled --- ask twice, get two) &
an image; a crystal \\
\emph{--- reader functions ---} & & \\
\(\pi\) & the \textbf{cheap} reader → the label & image classifier;
formula reader \\
\(\psi\) & the \textbf{expensive} reader → the target & aesthetic
scorer; band-gap simulation \\
\emph{--- values on an output ---} & & \\
\(P\) & the label \(\pi(D)\) --- a key that sorts outputs into bins &
``golden retriever''; (Si, O; 6) \\
\(L\) & the target property \(\psi(D)\) we steer & aesthetic,
brightness, band gap \\
\emph{--- bins and indices ---} & & \\
\(B,\ b,\ P_b\) & number of bins; a bin's index; bin \(b\)'s label &
1000 ImageNet classes \\
\emph{--- distributions ---} & & \\
\(\mathcal{J}\) & the distribution of your training set & ``the data the
model saw'' \\
\(\mu_P\) & a designed \textbf{recipe over bins} (what showing sets) &
``70\% beaches, 30\% snow'' \\
\emph{--- per-bin statistics ---} & & \\
\(w_b\) & a bin's \textbf{share} of the data & fraction of outputs in
the bin \\
\(g_b\) & a bin's \textbf{height} --- its average \(L\) & the bin's mean
band gap \\
\(v_b\) & a bin's \textbf{spread} --- variation of \(L\) inside it &
band-gap variation in the bin \\
\emph{--- budgets (a budget is a variance; a reweighting's reach is
\(\sqrt{\chi^2\cdot\text{budget}}\)) ---} & & \\
\(T\) & \textbf{within-bin} budget --- within-bin spread, what telling
draws on & reach \(\sqrt{\chi^2_{\text{tell}} T}\) \\
\(E\) & \textbf{between-bin} budget --- between-bin spread, what showing
draws on & reach \(\sqrt{\chi^2_{\text{show}} E}\) \\
\emph{--- effort ---} & & \\
\(\chi^2\) & how hard the recipe \(\mu_P\) pushes from the data's mix &
\(1\) = mild; larger = aggressive \\
\end{longtable}
}

\begin{center}\rule{0.5\linewidth}{0.5pt}\end{center}

\subsection{A. From the data budget to a real
model}\label{a.-from-the-data-budget-to-a-real-model}

Theory §3 states the bridge: a trained model inherits the budget for the
targets it reproduces (Claim 1), and a fine-tune reshapes that budget
predictably (Claim 4) --- both through a closed-form distribution. Here
is the machinery.

\subsubsection{A.1 The analytical path}\label{a.1-the-analytical-path}

The identities of Theory §1--§2 hold for \emph{any} conditional
distribution. At the limit of MLE training,
\(\mathrm{G}(\cdot \mid z) = \mathcal{J}(\cdot \mid z)\) as full
distributions, so the audit of \(\mathcal{J}\) is the audit of
\(\mathrm{G}\) --- for \emph{every} target at once. That limit is what
MLE aims at; a real model reaches it only approximately, and then the
correspondence is \textbf{per target}: \(\mathrm{G}\) carries
\(\mathcal{J}\)'s statistics for a target only as far as that target
rides on the training objective (the model is never trained on the
target itself). So auditing \(\mathrm{G}\) always means auditing
\emph{how well it reproduces a chosen \(L\)} --- never \(\mathrm{G}\) in
the abstract --- and A.3 checks exactly that, per target. For a known
training intervention, \(\mathrm{G}\) approximates an effective
distribution \(\mathcal{J}'\) with a closed form --- so you recompute
the per-bin \(w_b, g_b, v_b\) on \(\mathcal{J}'\) (by reweighting the
same rows) and every result carries over, at no new audit cost:

{\def\LTcaptype{none} 
\begin{longtable}[]{@{}
  >{\raggedright\arraybackslash}p{(\linewidth - 2\tabcolsep) * \real{0.5000}}
  >{\raggedright\arraybackslash}p{(\linewidth - 2\tabcolsep) * \real{0.5000}}@{}}
\toprule\noalign{}
\begin{minipage}[b]{\linewidth}\raggedright
Intervention
\end{minipage} & \begin{minipage}[b]{\linewidth}\raggedright
\(\mathcal{J}'\)
\end{minipage} \\
\midrule\noalign{}
\endhead
\bottomrule\noalign{}
\endlastfoot
MLE on \(\mathcal{J}\) & \(\mathcal{J}' = \mathcal{J}\) \\
SFT with replay \(\alpha\) &
\(\mathcal{J}' = \alpha \mathcal{J} + (1-\alpha) \mathcal{J}_{\text{SFT}}\) \\
RLHF / KL-PPO with reward \(R\), penalty \(\beta\) &
\(\mathcal{J}'(D \mid z) \propto \mathcal{J}(D \mid z)\, e^{R(D)/\beta}\) \\
DPO & Boltzmann form with implicit reward \\
Data filtering on \(S\) &
\(\mathcal{J}' = \mathcal{J} \cdot \mathbf{1}[D \in S]/Z\) \\
Distillation & teacher distribution, optionally temperature-scaled \\
\end{longtable}
}

\subsubsection{A.2 The empirical path}\label{a.2-the-empirical-path}

When the analytical path isn't available --- custom training, unknown
parameters, or an intervention aggressive enough that \(\mathcal{J}'\)
is degenerate on \(\mathcal{J}\)'s support --- audit \(\mathrm{G}\)
directly: measure how well \(\mathrm{G}\) reproduces the target \(L\)
you care about. The same call is needed whenever \(L\) sits \emph{far}
from what the model was trained to match. A target the training
objective \textbf{subsumes} --- image fidelity fixes the colour
distribution, fidelity being the superset --- or one that
\textbf{neighbours} it carries over from the data audit; a
\textbf{distant} target (token likelihood says little about
thermodynamic stability) does not, and must be checked on
\(\mathrm{G}\)'s own samples. Audit \(\mathrm{G}\) under a stated
reference \(\mu_P^{\text{ref}}\) and apply the same identities. The
data-side audit of Preliminaries is just \textbf{audit \(\mathcal{J}\)}
--- and, like every audit, it is an audit \emph{of a chosen \(L\)}.

This path is not merely specified --- \textbf{Evidence}'s robustness
check tests it directly, in both domains, against known ground truth.
\(E\), and the telling-vs-showing call it decides, recovers to within a
few percent under almost any \(\mu_P^{\text{ref}}\) that still overlaps
the real population, even a skewed guess; \(T\) recovers far less
reliably, sensitive to both the reference and the binning convention;
and a reference with no support overlap at all collapses both numbers
and can flip the qualitative call outright.

\subsubsection{A.3 Diagnostics}\label{a.3-diagnostics}

When the analytical \(\mathrm{G}\) may have drifted from the empirical
one, two cheap, \emph{independent} checks:

\begin{itemize}
\tightlist
\item
  \textbf{Support overlap.} Does \(\pi(\mathrm{G}(\cdot \mid z))\) land
  in \(\mathcal{J}_P\)'s bins?
\item
  \textbf{Per-bin mean drift \(\delta\).} Does
  \(g_{\mathrm{G}}(b) \approx g_{\mathcal{J}}(b)\)?
  \[\delta(\mu_P) = \sqrt{\mathbb{E}_{\mu_P}\!\big[(g_{\mathrm{G}}(P) - g_{\mathcal{J}}(P))^2 / v_P\big]}.\]
\end{itemize}

A model can pass one and fail the other --- right bins (high overlap),
wrong labels within them (large \(\delta\)). Check both. (Recipes in
\textbf{Application}.)

\subsubsection{A.4 What the model does --- and doesn't ---
change}\label{a.4-what-the-model-does-and-doesnt-change}

The first objection the budget invites: \emph{surely a more expressive
model --- one that interpolates more richly between training points ---
produces more variance, so the budget cannot be model-independent.} The
intuition is right about the \textbf{outputs} and wrong about the
\textbf{budget}.

A richer model walks the data manifold more densely: it emits a wider,
more novel \emph{set of outputs} \(D\). But the budget is the variance
of a scalar property \(L = \psi(D)\), and \textbf{richer interpolation
fills that property's distribution in more densely; it does not widen
it.} Interpolated outputs take property values \emph{between} the points
they interpolate --- within the range already present. The only way a
model enlarges \(\mathrm{Var}(L)\) is to \textbf{extrapolate} --- emit
outputs whose \(L\) falls outside the training range --- which is off
support (Part C).

So a model's expressiveness shows up in two places, neither of which
touches the budget:

\begin{itemize}
\tightlist
\item
  \textbf{How fully the budget is realized.} Finite models drop modes
  and under-fill bins (contraction,
  \(V_{\mathrm{G}} < V_{\mathcal{J}}\)); a richer model contracts less,
  approaching the data-side budget \emph{from below}. Expressiveness is
  \emph{less contraction}, measured by the within-bin variance ratio
  \(\rho = \sqrt{T(\mathrm{G})/T}\) --- a consistency check alongside
  \(\delta\) (\(\rho = 1\) match, \(<1\) contraction, \(>1\) expansion,
  the last only off support or under variance-rewarding objectives).
\item
  \textbf{Output diversity itself} --- coverage, novelty, the size of
  the produced set --- a non-decomposable, collection-level property
  (Part B). \emph{This} genuinely depends on the model and is
  \textbf{not} bounded by \(\mathrm{Var}(L)\); it lives on a different
  axis.
\end{itemize}

In one line: \textbf{richness widens the set of outputs, not the spread
of a labeled property.}

\subsubsection{A.5 The variance of a steered
collection}\label{a.5-the-variance-of-a-steered-collection}

What is the spread of the \(L\)'s a recipe \(\mu_P\) actually produces?
The same split applies, weighted by \(\mu_P\) rather than
\(\mathcal{J}_P\), but with the \emph{model's} per-bin values,
\[V_{\mathrm{G}(\mu)} = \sum_b \mu_P(b)\,v_b(\mathrm{G}) + \sum_b \mu_P(b)\,(g_b(\mathrm{G}) - \bar g_\mu)^2.\]
Because \(\mu_P\) is a \emph{prompt} distribution, this threads through
the model three times --- does it land outputs in the bins you asked for
(support overlap), with the per-bin means and spreads the data has
(\(\delta, \rho\), A.3--A.4), and are those bins in support at all? So
\(V_{\mathrm{G}(\mu)}\) is \textbf{exactly forecastable from the audit
when the model is faithful on the bins \(\mu_P\) touches} --- a
checkable condition --- and where it is not, the data side cannot answer
and you sample \(\mathrm{G}\) (A.2). If \(\mu_P\) leaves
\(\mathcal{J}\)'s support, the bound announces it:
\(\chi^2(\mu_P\|\mathcal{J}_P) \to \infty\), so \(\sqrt{\chi^2 E}\) goes
vacuous exactly where the audit falls silent. (Note
\(V_{\mathrm{G}(\mu)}\) is not capped by \(V_{\mathcal{J}}\):
concentrating \(\mu_P\) on extreme-mean bins makes the collection
\emph{more} spread than the data --- in support; what
\(\sqrt{\chi^2 E}\) bounds is the mean shift, not the variance.)

\emph{A worked edge case.} Take \(\mathcal{J}\), remove its two
extreme-mean bins to form \(\mathcal{J}'\), and train \(\mathrm{G}\) on
\(\mathcal{J}'\) --- then prompt for exactly those removed bins. Since
they carry no mass in \(\mathcal{J}'\),
\(\chi^2(\mu_P\|\mathcal{J}'_P) \to \infty\) and the bound is vacuous:
the framework makes no claim. Whether the spread of \(\mathrm{G}(\mu)\)
exceeds even \(V_{\mathcal{J}}\) is then entirely up to how
\(\mathrm{G}\) extrapolates --- outward, and it overshoots
\(\mathcal{J}\); back onto its own support, and it stays under
\(\mathcal{J}'\). The framework predicts neither; it flags the regime
(the diverging \(\chi^2\)) and sends you to measure \(\mathrm{G}\). The
example also fixes which distribution to audit: \(\mathrm{G}\)'s
reference is \(\mathcal{J}'\), the data it actually saw --- auditing
against the original \(\mathcal{J}\) would forecast a reach
\(\mathrm{G}\) has no in-support way to deliver.

\begin{center}\rule{0.5\linewidth}{0.5pt}\end{center}

\subsection{B. What each operation can
reach}\label{b.-what-each-operation-can-reach}

\subsubsection{B.1 The combinatorial
trap}\label{b.1-the-combinatorial-trap}

The Intro's within-bin/between-bin split acquires teeth here. Telling is
conjunctive: a bin specified by discrete property labels is an
\emph{intersection} of label values, and the number of such bins
explodes exponentially in the number of properties. In practice most
intersections are unrepresented in training, many were never labeled,
and many interesting regions fall between named label values. So the
fraction of \(\mathrm{Var}_{\mathcal{J}}(L)\) that label-based telling
can actually address \emph{shrinks} as property dimensionality grows.
Showing has no such problem: a set of examples points at the bins that
matter and needs no combinatorial coverage.

\subsubsection{B.2 Telling has narrow
reach}\label{b.2-telling-has-narrow-reach}

Confined to the within-bin budget \(T\): telling reweights within a bin,
so its reach is \(\sqrt{\chi^2_{\text{tell}}\,T}\) --- and since
per-prompt controls realize only \(\chi^2_{\text{tell}}\approx 1\) in
practice (\textbf{Evidence}), that working reach is about \(\sqrt{T}\).
Stacking tokens at one bin sharpens the bin-conditional but cannot leave
the bin or enlarge \(T\) --- telling only moves \(L\) around
\emph{within} a bin's existing spread. This is the precise form of
``telling can only narrow.''

\subsubsection{B.3 Decomposable vs non-decomposable
metrics}\label{b.3-decomposable-vs-non-decomposable-metrics}

Collection-level metrics split into \emph{decomposable}
(\(\Phi = (1/n)\sum_i
\phi(D_i)\)) and \emph{non-decomposable} (properties of the joint
distribution of the output set). Telling can partly substitute for
showing on decomposable metrics; on non-decomposable ones --- set
diversity, coverage, joint targets --- it cannot, because there is no
single-sample quantity to push.

\textbf{By curvature (the full table).} A goal is a property of the
batch; showing shifts weight across bins via the recipe \(\mu\) (all on
one bin = \emph{concentrate}, many bins = \emph{spread}), and the goal's
\textbf{curvature} decides which --- convex (peaks at one bin) →
concentrate, concave (peaks at a blend) → spread. Both are showing;
telling (within-bin knobs) only refines the result inside whatever bin
you land in. The familiar goals (each \emph{pushed up}), as a function
of \(\mu\):

{\def\LTcaptype{none} 
\begin{longtable}[]{@{}
  >{\raggedright\arraybackslash}p{(\linewidth - 6\tabcolsep) * \real{0.2500}}
  >{\raggedright\arraybackslash}p{(\linewidth - 6\tabcolsep) * \real{0.2500}}
  >{\raggedright\arraybackslash}p{(\linewidth - 6\tabcolsep) * \real{0.2500}}
  >{\raggedright\arraybackslash}p{(\linewidth - 6\tabcolsep) * \real{0.2500}}@{}}
\toprule\noalign{}
\begin{minipage}[b]{\linewidth}\raggedright
goal
\end{minipage} & \begin{minipage}[b]{\linewidth}\raggedright
as a function of \(\mu\)
\end{minipage} & \begin{minipage}[b]{\linewidth}\raggedright
curvature
\end{minipage} & \begin{minipage}[b]{\linewidth}\raggedright
best recipe (showing)
\end{minipage} \\
\midrule\noalign{}
\endhead
\bottomrule\noalign{}
\endlastfoot
mean \(\mathbb{E}[L]\) & \(\sum_b \mu(b)\, g_b\) & linear & concentrate
on the highest-mean bin \\
fraction over a bar \(P[L>\tau]\) & \(\sum_b \mu(b)\, r_b\) (bin rate
\(r_b\)) & linear & concentrate on the highest-rate bin \\
precision (tight estimate) & \(-\mathrm{Var}\) & convex & concentrate on
the lowest-spread bin \\
spread / variance & \(\mathrm{Var}(L)\) & concave & spread (mix high \&
low) \\
coverage of two corners &
\(\min(\text{mass on }A\text{-bins},\ \text{mass on }B\text{-bins})\) &
concave & spread (mix \(A\) and \(B\)) \\
diversity / \# distinct & entropy & concave & spread \\
match a target distribution & \(-\) (a divergence) & concave & spread
(the mixture) \\
raw extremes: max, min, range & non-smooth & --- & via proxies: a
quantile (concentrate + telling), the variance (spread) \\
\end{longtable}
}

(Bins are sorted into corners by their mean \(g_b\) --- what the audit
already gives; no finer per-bin numbers are needed.) The two non-obvious
rows: \textbf{variance is concave} because blending bins adds the
between-bin term,
\(\mathrm{Var}(\lambda R_1+(1{-}\lambda)R_2)=\lambda\mathrm{Var}_1+(1{-}\lambda)\mathrm{Var}_2+\lambda(1{-}\lambda)(\mu_1-\mu_2)^2\),
so spread is a showing goal; \textbf{distribution-matching} minimizes a
divergence (a quantity that bends upward), a concave goal whose best is
a mixture. The raw extremes are set by a single point and do not sort by
curvature; their smooth proxies do.

\textbf{Multiple properties (the Pareto front).} With \(N\) properties
each bin is a point in \(\mathbb{R}^N\) and a recipe's reachable means
fill the convex hull. The goal --- span the \textbf{Pareto front} of
non-dominated trade-offs --- is a coverage goal, so by the curvature
rule it wants a mixture. The lone exception is a single bin's
\emph{within-bin spread}: its \(k\) draws reach out from the mean by
about \(\sigma\,\Phi^{-1}(1-1/k)\) (the expected extreme of \(k\)
Gaussians), filling an ellipsoid set by the bin's covariance; if that
reaches the front's corners, one bin covers it. The crossover is a
single Mahalanobis comparison --- one bin suffices iff
\(\Delta\mu^{\!\top}\Sigma^{-1}\Delta\mu < [\Phi^{-1}(1-1/k)]^2\), with
\(\Delta\mu\) the gap from the bin's mean to the best specialist per
property and \(\Sigma\) its within-bin covariance --- so a wide front or
tight bins force diversification (showing), a narrow front or loose bins
allow one bin. This tilts toward showing as \(N\) grows: each added
property is another corner the one ellipsoid must reach while its reach
does not grow, so by \(N \gtrsim 5\) specialist bins are almost always
needed (a lower bound: one bin plus one specialist per property the best
bin cannot reach). Coverage is submodular, so greedy specialist
selection is within \(1-1/e\) of optimal; the exact objective is the
expected hypervolume improvement (EHVI; Emmerich, Deutz \& Klinkenberg
2006).

\subsubsection{B.4 Five cases}\label{b.4-five-cases}

Each opens on the concrete instance, then states the structural reason
--- always the same: telling is confined to within-bin room \(T\) (reach
\(\sqrt{\chi^2_{\text{tell}}\,T}\), with
\(\chi^2_{\text{tell}}\approx 1\)); showing reaches between-bin room
(\(\sqrt{\chi^2\cdot E}\)).

\textbf{Joint targets.} \emph{Image: high-aesthetic AND animal-like.
Crystal: wide-bandgap AND on-hull.} Telling both narrows the conditional
toward zero mass --- the cleanest failure. But some bins already contain
joint examples, and a \(\mu_P\) on those bins reaches the joint by
composition.

\textbf{Tacit targets.} \emph{Image: a cloud of pictures you find
compelling. Crystal: materials you recognize as promising.} The essence
of showing: you cannot articulate the shared property, so there is no
token to set. The exemplar set is the specification; the bin
distribution it induces does the steering, and \(\mathrm{G}\) draws more
without anyone saying why.

\textbf{Non-rectangular property spaces.} \emph{Crystal: the realizable
manifold is not a box in property coordinates.} Telling tries to reach
off-manifold corners and fails; a \(\mu_P\) on bins that populate those
corners reaches them.

\textbf{Pareto-front exploration.} \emph{Both: cover a trade-off front
in two properties \(L_A, L_B\), not a point.} Telling gives one point at
a time; a \(\mu_P\) weighted along the front covers it by construction
--- a collection-level target with no single-sample analogue. (This is
the one place two properties appear together; elsewhere a single \(L\)
suffices.)

\textbf{Properties without articulable tokens.} \emph{Both: diversity,
novelty, set statistics.} Telling has no token to manipulate; the choice
of \(\mu_P\) always exists as a knob.

\subsubsection{B.5 Two honest failure modes of
showing}\label{b.5-two-honest-failure-modes-of-showing}

Showing is not magic, and the formalism says where it strains.

\begin{enumerate}
\def\labelenumi{\arabic{enumi}.}
\tightlist
\item
  \textbf{The density problem.} ``More like these'' assumes
  \(\mathrm{G}\)'s geometry agrees with what made the exemplars
  appealing. It may not --- \(\mathrm{G}\) does not know \emph{why} the
  set was chosen, so ``near in \(\mathrm{G}\)'s sense'' can drift from
  ``appealing in the user's sense.'' This is the gap between bin
  membership and the within-bin detail \(T\) the user may actually care
  about.
\item
  \textbf{The distributional-shape problem.} ``Variety along several
  axes at once'' --- exemplars define a cloud but not its \emph{shape}
  along specific axes. The resolution is in the formalism: treat the
  exemplar set as a \emph{distribution}, not a set. Want more big dogs?
  Supply more big-dog exemplars. The distribution of inputs shapes the
  distribution of outputs; the desired variety is curated in, not
  articulated.
\end{enumerate}

Both fall straight out of the budget decomposition, and naming them
sharpens the boundary of what showing buys rather than weakening it.

\subsubsection{B.6 What the exemplars tell you about a bin --- the
lift}\label{b.6-what-the-exemplars-tell-you-about-a-bin-the-lift}

In tacit steering (Application §3) you build \(\mu_P\) from exemplars
with no scorer. How much does that reveal about the unmeasured target?
Assume only that the expert \textbf{picks among the best} --- selection
probability rises with \(L\), at overall rate \(s\) (e.g.~\(s = 0.01\)
for a top-centile bar). Let \(q_b\) be the fraction of bin \(b\)'s
members that clear the bar. If the exemplars are a fair sample of the
selected, the share landing in bin \(b\) is
\(\mu_P(b) = w_b\, q_b / s\), so
\[q_b = s \cdot \mathrm{lift}_b, \qquad \mathrm{lift}_b = \mu_P(b)/w_b.\]
Every term on the right comes from the cheap key alone --- \(w_b\) from
\(\pi\) over the data, \(\mu_P(b)\) from \(\pi\) over the exemplars ---
so \textbf{the lift estimates the fraction of a bin that clears the
expert's bar, with no \(\psi\).} A small bin that soaks up exemplars has
large lift and \(q_b \to 1\) (a saturated, reliable pocket); a big bin
that merely contains them has lift \(\approx 1\) and \(q_b \approx s\)
(the exemplars are its thin top tail). The global
\(\chi^2 = \sum_b w_b(\mathrm{lift}_b - 1)^2\) is just the
\(w\)-weighted spread of these lifts.

\emph{Reliability.} A hard floor needs no model: the \(k_b\) exemplars
in a bin are themselves selected members, so \(q_b \ge k_b/(N w_b)\)
exactly (\(N\) the candidate count). The point estimate
\(q_b = s\cdot\mathrm{lift}\) carries relative noise
\(\approx 1/\sqrt{k_b}\) in the per-bin exemplar count \(k_b\) --- one
exemplar tells you almost nothing, a hundred is good to \(\sim\!10\%\).
(Sampling new outputs from these bins then inherits the usual carry-over
of §A.2 --- \(\mathrm{G}\) must reproduce the bin's \(L\)-distribution.)

\emph{Worked example.} \(N = 10{,}000\) candidates; a bin of \(20\)
(\(w_b = 0.2\%\)); all \(10\) of your top-centile exemplars land in it.
The floor alone gives \(q_b \ge 10/20 = 0.5\) --- half the bin clears
the bar, \(50\times\) the base rate. The lift, \(1/0.002 = 500\),
saturates \(q_b\) at \(1\): essentially a pure high-\(L\) pocket. Under
the null (\(L\) independent of the bin) the same event has probability
\((0.002)^{10} \approx 10^{-27}\).

\begin{center}\rule{0.5\linewidth}{0.5pt}\end{center}

\subsection{C. Scope --- the full boundary
statement}\label{c.-scope-the-full-boundary-statement}

Theory §5 summarizes the boundary; here it is in full. None of what
follows is a hedge --- each is a clean edge of a sharp claim, and almost
every one comes with a cheap check that tells you which side of it you
are on.

\textbf{What the framework claims.} The budget and its two reaches bound
the \emph{mean shift} of a labeled property \(L\) that can be had by
\emph{selection} --- choosing which of the model's behaviors to sample,
whether by telling (within a bin) or showing (across bins). The claims
hold when the model's output distribution matches \(\mathcal{J}\) on the
bins in play: at the MLE limit within support, or --- after a standard
fine-tune --- against the closed-form \(\mathcal{J}'\) of A.1.

\textbf{Where it stops --- and what to do instead.}

\begin{itemize}
\tightlist
\item
  \textbf{Off support.} If the model is pushed to extrapolate, or
  \(\mu_P\) puts mass on bins absent from training, the data-side audit
  has nothing to say. The signal is built in:
  \(\chi^2(\mu_P\|\mathcal{J}_P) \to
  \infty\) and support overlap drops. There, drop the forecast and audit
  \(\mathrm{G}\) empirically (A.2).
\item
  \textbf{Mean, not range.} We bound the \emph{mean shift} (and account
  for variances); we never claim the literal extremes. A model can
  produce an \(L\) past anything in training \emph{by extrapolating} ---
  off-support behavior, not a broken claim, because the min--max range
  was never the quantity.
\item
  \textbf{Realization (the model can fall short or run over).} The
  budget is a data-side \emph{forecast}. A finite model usually falls
  short by \emph{contracting} (\(\rho < 1\), mode-dropping); off support
  or under variance-rewarding objectives it can \emph{expand}
  (\(\rho > 1\)) past the reach. The diagnostics \(\rho\) and \(\delta\)
  (A.3--A.4) say which.
\item
  \textbf{The spread of a steered batch.} \(V_{\mathrm{G}(\mu)}\) is
  \emph{not} capped by \(\mathrm{Var}_{\mathcal{J}}(L)\) ---
  concentrating \(\mu_P\) on extreme-mean bins makes a batch \emph{more}
  spread than the data, in support. The reach bounds the mean shift, not
  the batch variance (A.5).
\item
  \textbf{Heavy-tailed bin means.} The showing reach is a second-moment
  bound; a few bins with extreme means can carry the mean shift past it.
  Inspect the tail before quoting it tight (Application §4.4).
\item
  \textbf{Set-level targets.} The reaches bound a \emph{mean}.
  Non-decomposable, collection-level properties --- diversity, coverage,
  joint hit-rate --- are a different object (B.3); telling and showing
  still differ there, but the closed-form bound is for the mean.
\item
  \textbf{Leaving the selection regime.} Operations that don't work by
  selection --- gradient search on the output at inference time, say ---
  can push \(L\) past the budget. They escape it only by leaving the
  regime where training data predicts behavior: a trade, not a free
  lunch.
\item
  \textbf{Multi-property \(L\).} We work one property at a time; several
  at once generalize through the law of total covariance (and appear
  together only in the Pareto case, B.4).
\item
  \textbf{Non-iid generation} complicates the concentration arguments
  for collection-level stability; the reaches hold in expectation, and a
  particular batch may deviate.
\end{itemize}

\textbf{Practical (not theoretical) caveats} --- raw-output discipline,
label-pipeline drift for learned-classifier labels, finite-sample audit
noise --- are in \textbf{Application} §4 (with the details in
\textbf{E}).

\begin{center}\rule{0.5\linewidth}{0.5pt}\end{center}

\subsection{D. Why variance, not range (the methodological
choice)}\label{d.-why-variance-not-range-the-methodological-choice}

Theory §1 uses \(\sqrt{\mathrm{Var}_{\mathcal{J}}(L)}\) as the unit of
reach and notes in passing that the range \(\max L - \min L\) would be
the wrong choice. In full: mean-shift reach under controlled reweighting
scales with variance, not range (that is what the Cauchy--Schwarz step
gives); range has no clean within-bin / between-bin identity, so it
cannot be split into the two per-operation budgets at all; and range is
fragile --- set by a single extreme row, sensitive to sampling luck, and
routinely exceeded by \(\mathrm{G}\)'s own outputs when the model
interpolates. Variance is the stable data-side quantity. (Worked
argument in \texttt{crystal\_example.md}.)

\begin{center}\rule{0.5\linewidth}{0.5pt}\end{center}

\subsection{E. Practical pitfalls --- the
details}\label{e.-practical-pitfalls-the-details}

The seven traps listed in \textbf{Application §4}, each with its numbers
and the check to run (numbered to match §4; cross-references point back
to Application).

\subsubsection{4.1 Audit raw outputs, not post-processed
ones}\label{audit-raw-outputs-not-post-processed-ones}

Run DFT relaxation on a crystal model's outputs before auditing them and
the gap between prediction and observation jumps about fourfold. The
reason: the audit describes what the \emph{model} puts out, and
relaxation --- like CFG for images, or any refinement network --- is a
separate step with its own effect on the distribution. So audit and
measure the raw outputs (pre-relaxation structures, images at CFG=1),
and treat any post-processing as its own intervention.

\subsubsection{4.2 Watch for label drift between audit and
check}\label{watch-for-label-drift-between-audit-and-check}

On DiT at CFG=1, the LAION aesthetic scorer rates the \emph{generated}
images about 0.44 points lower than it rated the \emph{training} images
--- even though the model's outputs otherwise match the data. That gap
isn't the model; it's the scorer reacting differently to generated
images than to real ones. (Brightness and CLIP-similarity labels,
crisply defined, show essentially none of it.) The general trap: your
prediction uses the label values recorded in the data (\(g_b\)), but
your check uses whatever your evaluation pipeline computes --- produce
the two differently and a systematic offset appears. \emph{Crystal
example:} evaluating \emph{Combined}, the training stability label is a
DFT energy-above-hull, but the benchmark recomputes it with an ensemble
of machine-learned potentials; the two disagree at the stability
boundary by \(\sim 10\)pp, producing a roughly constant offset between
predicted and observed Combined. Either recompute the audit with the
evaluation pipeline's labels (expensive) or report the
observed-vs-predicted gap separately (the honest two-step framing).
Structural labels measured the same way on both sides (brightness,
density) don't have this issue.

\subsubsection{4.3 When a fine-tune is too aggressive to audit from
data}\label{when-a-fine-tune-is-too-aggressive-to-audit-from-data}

A crystal model fine-tuned hard toward novelty drifted so far that it
almost never produced the structures its base did --- it scored held-out
base structures \emph{below} random chance (per-token log-probability
\(\approx -17\), versus \(\approx -9.7\) for uniform). At that point
there's no use computing the budget from the original data: the model no
longer resembles it. The shortcut for a fine-tuned model --- rewriting
it as a reweighted copy of the data (the \(\mathcal{J}'\) trick,
Appendix A.1) --- works only while the fine-tune is \emph{gentle} enough
that the model still mostly produces things the data contained.
\textbf{How to check:} sample a few hundred outputs under your
deployment recipe and see what fraction land in training bins. High
(\(\ge 90\%\)): the data-based audit holds. Low (\(< 80\%\)): audit the
model's own samples instead (§1, sample-side).

\subsubsection{4.4 A few extreme bins make the bound a loose
guide}\label{a-few-extreme-bins-make-the-bound-a-loose-guide}

The showing bound averages squared deviations across bins, so it goes
slack when a few bins sit far out from the rest --- a heavy tail. A
recipe that leans on those few bins can move the mean by much less than
the averaged bound suggests, so the bound stops being a tight prediction
and becomes only a ceiling. On ImageNet, brightness has exactly this
shape (a handful of very bright classes), and across our concentration
sweep its realized shift stayed well under the bound at every level ---
the ceiling held, but it was loose, rising far faster than the reach it
was supposed to predict. The averaging cuts the other way too: with a
heavy enough tail it carries no guarantee the realized shift won't run
over, so don't treat it as exact. \textbf{How to check:} before quoting
the bound as tight, look at the spread of bin means; if a handful
dominate, read it as a loose guide, not a sharp number.

\subsubsection{4.5 Recipes that wander off the training
data}\label{recipes-that-wander-off-the-training-data}

Build a \(\mu_P\) that puts real weight on bins the training set never
contained, and the budget has nothing to say there --- it was computed
from data that doesn't cover that ground. Usefully, the \(\chi^2\)
factor announces this itself: it blows up as \(\mu_P\) puts mass where
the data had none, so the bound \(\sqrt{\chi^2 E}\) goes vacuous right
at the edge of support. \textbf{Fix:} whenever you build a \(\mu_P\),
check how much of its mass lands on training bins; if a meaningful share
is off-support, drop those bins and renormalize, or flag those outputs
as outside the framework.

\subsubsection{4.6 Thin corner pools repeat
themselves}\label{thin-corner-pools-repeat-themselves}

Aiming at a target \(g^* = 0.35\) from a corner of only 20 bins (64
rows), a pool-mixing design drew 35 prompts but got only 87 distinct
ones out of 100 --- the same few bins kept coming up. With so little to
draw from you effectively have far fewer independent samples (here 73.5,
not 100), and the push \(\chi^2\) shoots up to 52.3, two orders of
magnitude above a healthy pool --- predictable from
\(\chi^2 \approx 1/q - 1\). \textbf{How to check:} compute the floor
\(q \gtrsim n_\text{gen}/n_\text{eligible}\); below it, switch to
min-\(\chi^2\), which borrows from the middling bins to recover
independent samples, at some cost in precision.

\subsubsection{4.7 You can only reach as far as your
pools}\label{you-can-only-reach-as-far-as-your-pools}

Showing steers \emph{between} the extremes your pools provide. With a
high pool at \emph{Combined} ≈ 0.85 and a broad low one at ≈ 0.14, the
reachable range is about \textbf{0.15 to 0.85}: ask for 0.1 and you
saturate at \textasciitilde0.15, ask above 0.85 and there is nothing to
mix toward. To widen the range you need a more extreme pool --- if the
data holds one. (More deeply, a real model carries its data's statistics
only in the ideal limit (Theory, Claim 1), so its reachable range can
sit \emph{inside} the data-side bound; past that, the bound
over-promises.) \textbf{How to check:} measure your two anchor pools'
scores first --- that bracket \emph{is} your range.

\begin{center}\rule{0.5\linewidth}{0.5pt}\end{center}

\section{References}\label{references}

\begin{itemize}
\tightlist
\item
  Batatia, Kovács, Simm, Ortner, Csányi. ``MACE: Higher Order
  Equivariant Message Passing Neural Networks for Fast and Accurate
  Force Fields.'' NeurIPS 2022.
\item
  Blattmann, Rombach, Oktay, Müller, Ommer. ``Retrieval-Augmented
  Diffusion Models.'' NeurIPS 2022.
\item
  Casella, Berger. \emph{Statistical Inference}, 2nd ed.~Duxbury, 2002.
\item
  Cochran. \emph{Sampling Techniques}, 3rd ed.~Wiley, 1977.
\item
  Cover, Thomas. \emph{Elements of Information Theory}, 2nd ed.~Wiley,
  2006.
\item
  Dathathri, Madotto, Lan, Hung, Frank, Molino, Yosinski, Liu. ``Plug
  and Play Language Models: A Simple Approach to Controlled Text
  Generation.'' ICLR 2020.
\item
  Deng, Dong, Socher, Li, Li, Fei-Fei. ``ImageNet: A Large-Scale
  Hierarchical Image Database.'' CVPR 2009.
\item
  Dhariwal, Nichol. ``Diffusion Models Beat GANs on Image Synthesis.''
  NeurIPS 2021.
\item
  Emmerich, Deutz, Klinkenberg. ``The Computation of the Expected
  Improvement in Dominated Hypervolume of Pareto Front Approximations.''
  Technical Report, Leiden University, 2006.
\item
  Gal, Alaluf, Atzmon, Patashnik, Bermano, Chechik, Cohen-Or. ``An Image
  is Worth One Word: Personalizing Text-to-Image Generation using
  Textual Inversion.'' ICLR 2023.
\item
  Goodfellow, Pouget-Abadie, Mirza, Xu, Warde-Farley, Ozair, Courville,
  Bengio. ``Generative Adversarial Nets.'' NeurIPS 2014.
\item
  He, Zhang, Ren, Sun. ``Deep Residual Learning for Image Recognition.''
  CVPR 2016.
\item
  Ho, Salimans. ``Classifier-Free Diffusion Guidance.''
  arXiv:2207.12598, 2022.
\item
  Jahanian, Chai, Isola. ``On the `Steerability' of Generative
  Adversarial Networks.'' ICLR 2020.
\item
  Korbak, Perez, Buckley. ``RL with KL Penalties is Better Viewed as
  Bayesian Inference.'' Findings of EMNLP 2022.
\item
  Krause, Golovin. ``Submodular Function Maximization.'' In
  \emph{Tractability: Practical Approaches to Hard Problems}, Cambridge
  UP, 2014.
\item
  Krause, Gotmare, McCann, Keskar, Joty, Socher, Rajani. ``GeDi:
  Generative Discriminator Guided Sequence Generation.'' Findings of
  EMNLP 2021.
\item
  Kynkäänniemi, Karras, Laine, Lehtinen, Aila. ``Improved Precision and
  Recall Metric for Assessing Generative Models.'' NeurIPS 2019.
\item
  LAION. ``LAION-Aesthetics.'' laion.ai/blog, 2022.
\item
  Lattimore, Szepesvári. \emph{Bandit Algorithms.} Cambridge UP, 2020.
\item
  Lester, Al-Rfou, Constant. ``The Power of Scale for
  Parameter-Efficient Prompt Tuning.'' EMNLP 2021.
\item
  Li, Liang. ``Prefix-Tuning: Optimizing Continuous Prompts for
  Generation.'' ACL-IJCNLP 2021.
\item
  Naeem, Oh, Uh, Choi, Yoo. ``Reliable Fidelity and Diversity Metrics
  for Generative Models.'' ICML 2020.
\item
  Nemhauser, Wolsey, Fisher. ``An Analysis of Approximations for
  Maximizing Submodular Set Functions---I.'' \emph{Mathematical
  Programming} 14:265--294, 1978.
\item
  Ouyang, Wu, Jiang, Almeida, Wainwright, et al.~``Training Language
  Models to Follow Instructions with Human Feedback.'' NeurIPS 2022.
\item
  Owen. \emph{Monte Carlo Theory, Methods and Examples.} 2013.
\item
  Peebles, Xie. ``Scalable Diffusion Models with Transformers.'' ICCV
  2023.
\item
  Peyré, Cuturi. ``Computational Optimal Transport.'' \emph{Foundations
  and Trends in Machine Learning} 11(5--6):355--607, 2019.
\item
  Polanyi. \emph{The Tacit Dimension.} Doubleday, 1966.
\item
  Radford, Kim, Hallacy, Ramesh, Goh, Agarwal, et al.~``Learning
  Transferable Visual Models From Natural Language Supervision.'' ICML
  2021.
\item
  Rafailov, Sharma, Mitchell, Ermon, Manning, Finn. ``Direct Preference
  Optimization: Your Language Model is Secretly a Reward Model.''
  NeurIPS 2023.
\item
  Ruiz, Li, Jampani, Pritch, Rubinstein, Aberman. ``DreamBooth: Fine
  Tuning Text-to-Image Diffusion Models for Subject-Driven Generation.''
  CVPR 2023.
\item
  Sajjadi, Bachem, Lucic, Bousquet, Gelly. ``Assessing Generative Models
  via Precision and Recall.'' NeurIPS 2018.
\item
  Salimans, Goodfellow, Zaremba, Cheung, Radford, Chen. ``Improved
  Techniques for Training GANs.'' NeurIPS 2016.
\item
  Schmidt, Hoffmann, Wang, Borlido, Carriço, Cerqueira, Botti, Marques.
  ``Machine-Learning-Assisted Determination of the Global
  Zero-Temperature Phase Diagram of Materials.'' \emph{Advanced
  Materials} 35(22):2210788, 2023.
\item
  Schuhmann, Beaumont, Vencu, Gordon, Wightman, et al.~``LAION-5B: An
  Open Large-Scale Dataset for Training Next Generation Image-Text
  Models.'' NeurIPS 2022 Datasets and Benchmarks Track.
\item
  Shen, Gu, Tang, Zhou. ``Interpreting the Latent Space of GANs for
  Semantic Face Editing.'' CVPR 2020.
\item
  Xie, Fu, Ganea, Barzilay, Jaakkola. ``Crystal Diffusion Variational
  Autoencoder for Periodic Material Generation.'' ICLR 2022.
\item
  Ye, Zhang, Liu, Han, Yang. ``IP-Adapter: Text Compatible Image Prompt
  Adapter for Text-to-Image Diffusion Models.'' arXiv:2308.06721, 2023.
\item
  Zeni, Pinsler, Zügner, Fowler, Horton, Fu, et al.~``A Generative Model
  for Inorganic Materials Design'' (MatterGen). \emph{Nature}
  639:624--632, 2025.
\item
  Ziegler, Stiennon, Wu, Brown, Radford, Amodei, Christiano, Irving.
  ``Fine-Tuning Language Models from Human Preferences.''
  arXiv:1909.08593, 2019.
\end{itemize}

\end{document}